%% file: main.tex
\documentclass{article}
\usepackage[utf8]{inputenc}
\usepackage{main}
\usepackage{microtype}
\usepackage{graphicx}
\usepackage{times}
\usepackage{amsmath}
\usepackage{amssymb}
\usepackage{enumitem}
\usepackage{booktabs}
\usepackage{xcolor}     
\usepackage{natbib}
\definecolor{mydarkblue}{rgb}{0,0.08,0.45}
\usepackage[colorlinks=true,linkcolor=mydarkblue,citecolor=mydarkblue,filecolor=mydarkblue,urlcolor=mydarkblue]{hyperref}
\usepackage{fancyhdr}

\input{pre}

\usepackage{caption}

\usepackage{color}
\usepackage{tabularx}
\usepackage{tabu}
\usepackage{booktabs}
\usepackage{colortbl}
\usepackage{multirow}

\usepackage{graphicx} 
\usepackage{float}
\usepackage{subfigure} 
\usepackage{amssymb}

\usepackage{hyperref}
\usepackage{tablefootnote}
\usepackage{xspace}

\usepackage{float}
\usepackage{amsmath}
\usepackage{bm}
\usepackage{algorithmicx}
\usepackage{algorithm}
\usepackage{algpseudocode}

\usepackage{arydshln}
\usepackage{amssymb}
\usepackage{pifont}
\usepackage{enumitem}

\newenvironment{itemize*}%
 {\leftmargini=10pt\begin{itemize}%
  \setlength{\itemsep}{0pt}%
  \setlength{\parskip}{0pt}%
  }%
 {\end{itemize}}
\newenvironment{enumerate*}%
 {\begin{enumerate}%
  \setlength{\itemsep}{0pt}%
  \setlength{\parskip}{0pt}}%
 {\end{enumerate}}

\usepackage[most]{tcolorbox}
\definecolor{myblue}{rgb}{0.18,0.45,0.73}

\usepackage{pgfplots}
\pgfplotsset{compat=1.18}
\usepackage{xcolor}

\usepackage{booktabs}
\usepackage{siunitx}
\usepackage{color}
\usepackage{tabularx}
\usepackage{tabu}
\usepackage{booktabs}
\usepackage{multirow}
\usepackage{caption}
\usepackage{placeins}
\usepackage{tikz}
\usepackage{forest}
\useforestlibrary{edges}

\usepackage{xcolor}
\usepackage{fancyhdr}
\usepackage{ragged2e}
\usetikzlibrary{shadows, arrows.meta, shapes, positioning}

\usepackage{makecell}
\usepackage{adjustbox}
\usepackage{array}
\setcellgapes{1pt}
\makegapedcells

\usepackage{graphicx} 
\usepackage{float}
\usepackage{subfigure} 
\usepackage{amssymb}

\usepackage{hyperref}
\usepackage{tablefootnote}
\usepackage{xspace}

\usepackage{float}
\usepackage{amsmath}
\usepackage{bm}
\usepackage{algorithmicx}
\usepackage{algorithm}
\usepackage{algpseudocode}

\usepackage{arydshln}

\usepackage{amssymb}
\usepackage{pifont}
\usepackage{enumitem}

\usepackage{microtype}
\usepackage{tcolorbox}

\definecolor{hidden-red}{RGB}{240, 100, 100}
\definecolor{hidden-green}{RGB}{100, 200, 100}
\definecolor{hidden-blue}{RGB}{100, 100, 240}
\definecolor{darkblue}{RGB}{70, 70, 160}
\newcolumntype{Y}{>{\noindent\justifying\ignorespaces\arraybackslash}X}
\tikzstyle{my-box}=[
    rectangle,
    draw=black,
    rounded corners,
    text opacity=1,
    minimum height=1.5em,
    minimum width=5em,
    inner sep=2pt,
    align=left,
    fill opacity=.5,
]
\tikzstyle{leaf5}=[
    my-box,
    fill=darkblue!15,
    text=black,
    font=\scriptsize,
    inner xsep=5pt,
    inner ysep=4pt,
    align=left,
]
\tikzstyle{root-node}=[
    fill=myblue!15,
    text=black,
]

\tikzstyle{gen-node}=[
    my-box,
    fill=hidden-blue!18,
    text=black,
    font=\normalsize,
    inner xsep=5pt,
    inner ysep=4pt,
]

\tikzstyle{after-node}=[
    my-box,
    fill=hidden-green!22,
    text=black,
    font=\normalsize,
    inner xsep=5pt,
    inner ysep=4pt,
]

\tikzstyle{bench-node}=[
    my-box,
    fill=orange!22,
    text=black,
    font=\normalsize,
    inner xsep=5pt,
    inner ysep=4pt,
]
\tikzstyle{gen-leaf}=[leaf5, fill=hidden-blue!18] 
\tikzstyle{after-leaf}=[leaf5, fill=hidden-green!22]
\tikzstyle{bench-leaf}=[leaf5, fill=orange!22]

\usepackage{microtype}

\newcommand{\Yale}{\hspace{.1em}^{\textcolor{YaleBlue}{\boldsymbol{Y}}}}
\newcommand{\TCS}{\hspace{.1em}^{\textcolor{TCSC}{\boldsymbol{T}}}}
\newcommand{\NYU}{\hspace{.1em}^{\textcolor{NYUPurple}{\boldsymbol{N}}}}

\definecolor{YaleBlue}{RGB}{0, 53, 107}
\definecolor{NYUPurple}{RGB}{134, 1, 175}  
\definecolor{TCSC}{RGB}{1, 126, 199}

\begin{document}

\title{Can AI Be a Good Peer Reviewer? A Survey of Peer Review Process, Evaluation, and the Future}

\author{
\textbf{Sihong Wu}$\Yale$ \quad
\textbf{Owen Jiang}$\Yale$ \quad
\textbf{Yilun Zhao}$\Yale$\thanks{Correspondence to: Yilun Zhao (\texttt{yilun.zhao@yale.edu})} \quad
\textbf{Tiansheng Hu}$\NYU$ \quad
\textbf{Yiling Ma}$\Yale$ \\ [3pt]
\textbf{Kaiyan Zhang}$\Yale$ \quad
\textbf{Manasi Patwardhan}$\TCS$ \quad
\textbf{Arman Cohan}$\Yale$ \\ [7pt]
 $\Yale$Yale University \quad $\NYU$New York University \quad $\TCS$TCS Research
}

\maketitle
\thispagestyle{fancy}
\fancyhead{}
\lhead{%
    \includegraphics[height=1.1cm]{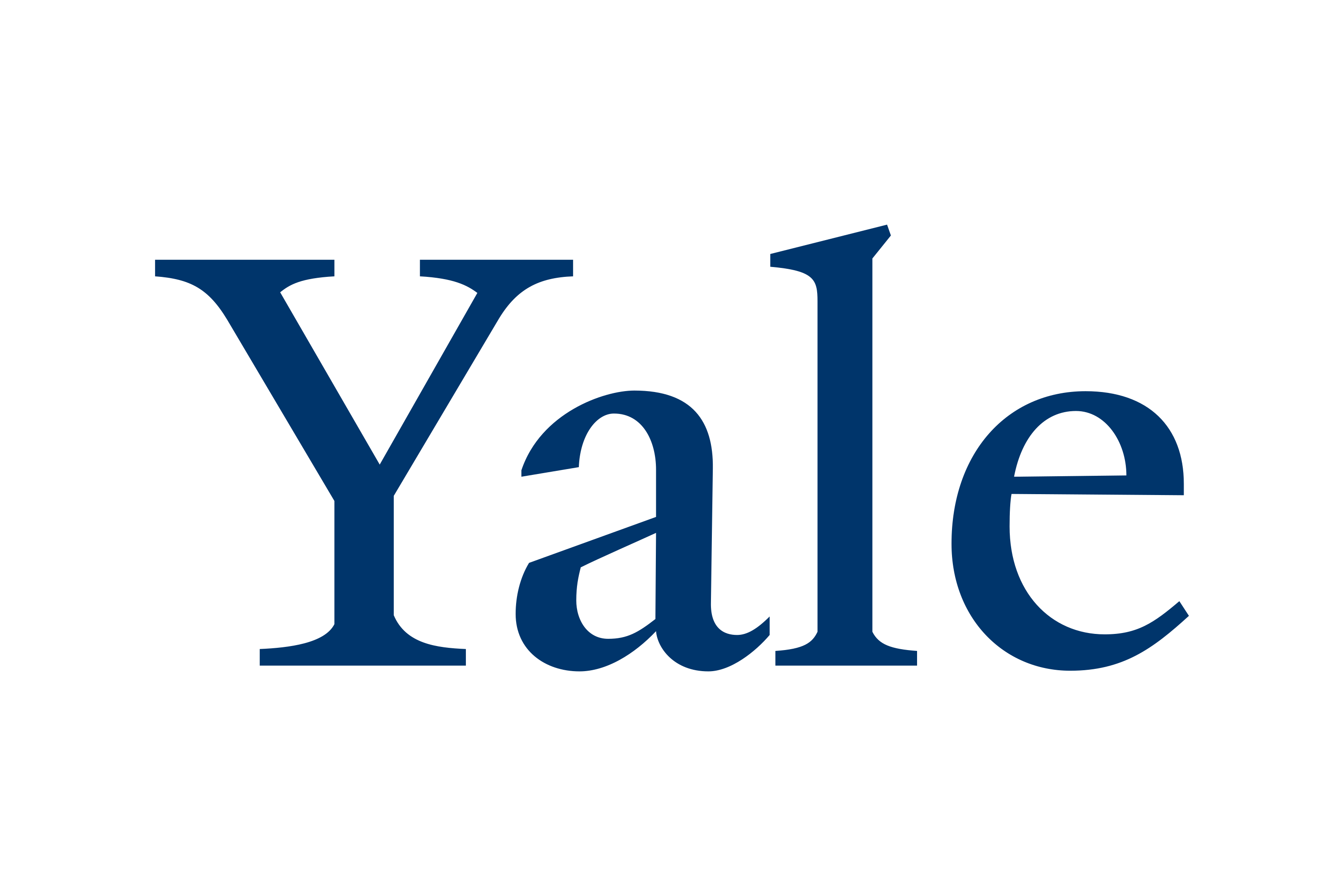}\hspace{0.2cm}%
    \raisebox{0.2cm}{\includegraphics[height=0.7cm]{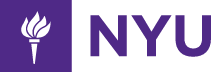}}\hspace{0.2cm}%
    \raisebox{0.05cm}{\includegraphics[height=0.8cm]{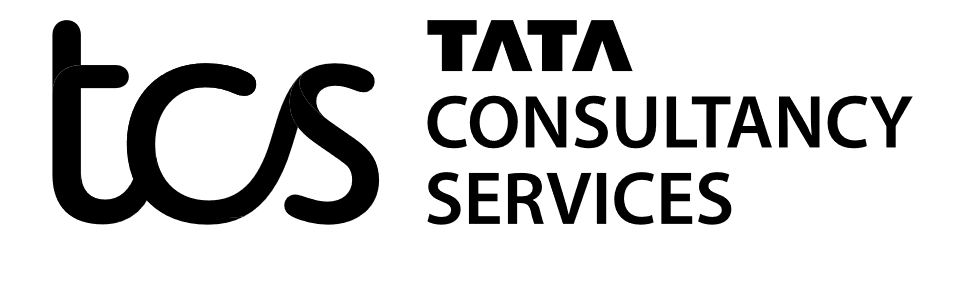}}%
}

\fancyfoot[C]{\thepage}
\renewcommand{\headrulewidth}{0pt}
\setlength{\headheight}{12pt}
\addtolength{\topmargin}{0pt}
\setlength{\headsep}{3mm}

\vspace{-1.0em}

\pagestyle{plain}

\begin{abstract}
\input{main/0-abstract}
\end{abstract}

\input{main/1-introduction}

\input{main/4-methodologies}
\input{main/After_peer_review}
\input{main/3-evaluation}
\input{main/5-discussion_future_directions}

\input{main/6-conclusion}
\input{main/limitations}

\bibliography{custom}

\clearpage
\appendix

\input{appendix/metrics}
\input{appendix/discussion_benchmark}
\input{appendix/discussion_methodologies}
\input{appendix/methodologies_summary}
\input{appendix/data_collection}
\input{appendix/evaluation_summary}
\input{appendix/review_procedure}
\end{document}

%% file: pre.tex
\definecolor{gred}{RGB}{250, 210, 207}
\definecolor{coolblue1}{rgb}{0.91, 0.94, 0.98}
\definecolor{coolblue2}{rgb}{0.76, 0.85, 0.94}
\definecolor{coolblue3}{rgb}{0.54, 0.72, 0.87}
\definecolor{coolblue4}{rgb}{1, 1, 1}

\usepackage[table]{xcolor}  

\usepackage{xspace}
\usepackage{cleveref}
\usepackage[]{multicol, multirow}
\usepackage[most]{tcolorbox}
\usepackage{etoc}

\tcbuselibrary{listingsutf8}
\tcbuselibrary{listingsutf8,breakable}

\newtcolorbox[auto counter]{observation}[1][]{
  colback=black!5!white,
  colframe=black!70!white,
  fonttitle=\bfseries,
  title=Observation~\thetcbcounter,
  enhanced,
  boxrule=0.6pt,
  left=1mm,right=1mm,top=1mm,bottom=1mm,
  #1
}

\newtcolorbox[auto counter]{takeaway}[1][]{
  colback=teal!3!white,
  colframe=teal!55!black,
  fonttitle=\bfseries,
  title=Takeaway~\thetcbcounter,
  enhanced,
  boxrule=0.5pt,
  left=1mm,right=1mm,top=1mm,bottom=1mm,
  #1
}

\newtcolorbox[auto counter]{practicalguidance}[1][]{
  colback=cyan!3!white,
  colframe=cyan!60!black,
  fonttitle=\bfseries,
  title=Practical Guidance~\thetcbcounter,
  enhanced,
  boxrule=0.5pt,
  left=1mm,right=1mm,top=1mm,bottom=1mm,
  #1
}

\newtcolorbox[auto counter]{discussion}[1][]{
  colback=violet!4!white,
  colframe=violet!60!black,
  fonttitle=\bfseries,
  title=Discussion~\thetcbcounter,
  enhanced,
  boxrule=0.5pt,
  left=1mm,right=1mm,top=1mm,bottom=1mm,
  #1
}



%% file: main/0-abstract.tex
Peer review is a multi-stage process involving reviews, rebuttals, meta-reviews, final decisions, and subsequent manuscript revisions. Recent advances in large language models (LLMs) have motivated methods that assist or automate different stages of this pipeline. In this survey, we synthesize techniques for (i) peer review generation, including fine-tuning strategies, agent-based systems, RL-based methods, and emerging paradigms to enhance generation; (ii) after-review tasks including rebuttals, meta-review and revision aligned to reviews; and (iii) evaluation methods spanning human-centered, reference-based, LLM-based and aspect-oriented. We catalog datasets, compare modeling choices, and discuss limitations, ethical concerns, and future directions. The survey aims to provide practical guidance for building, evaluating, and integrating LLM systems across the full peer review workflow. A collection of papers is available at the following repository: 
\url{https://github.com/formula12/Awesome-Peer-Review}.

%% file: main/1-introduction.tex
\section{Introduction}

Peer review is the cornerstone of scientific publishing. The rapid advancement of AI4Research has fundamentally transformed the landscape of scholarly communication, spawning a surge of interest in automating and enhancing the peer review process~\cite{chen2025ai4researchsurveyartificialintelligence}. 
The academic peer review pipeline includes submission, review, rebuttal, discussion, meta-review, and final decision-making. Each stage involves the generation of critical textual artifacts. For instance, research has explored the creation of initial referee reports \cite{yuan2021automate,nlpee,peerreviewas} and author rebuttals that respond to reviewer feedback \cite{cheng-etal-2020-ape,purkayastha-etal-2023-exploring}. Further along the pipeline, work has focused on consolidating multiple reviews into a unified meta-review \cite{li-etal-2023-summarizing,zeng2023checklist,kumar2024pipeline} and providing structured arguments to support the final publication decision \cite{Sukpanichnant2024}.

Early approaches in this domain were confined to sub-tasks, such as predicting paper acceptance from metadata and abstract features or regressing review scores from paper content~\cite{peerread, Checco2021AI}. The advent of LLMs marked a profound paradigm shift~\cite{vaswani2017attention, radford2019language, bommasani2021opportunities, qwen, touvron2023llama,openai2024gpt4technicalreport,zhao2026survey}. 
Around 2023, models began to process full manuscripts and produce credible end-to-end reviews~\cite{gpt4slightly, Hosseini2023Fighting}. Current systems go beyond single-prompt generation to adopt multi-agent designs that emulate panel workflows~\cite{agentreview, durante2024agentaisurveyinghorizons} and reinforcement learning~\cite{ziegler2020finetuninglanguagemodelshuman, christiano2023deepreinforcementlearninghuman}
to align outputs with nuanced human preferences. These developments coexist with a growing recognition that evaluation must keep pace~\cite{zhang2025reviewingscientificpaperscritical}.

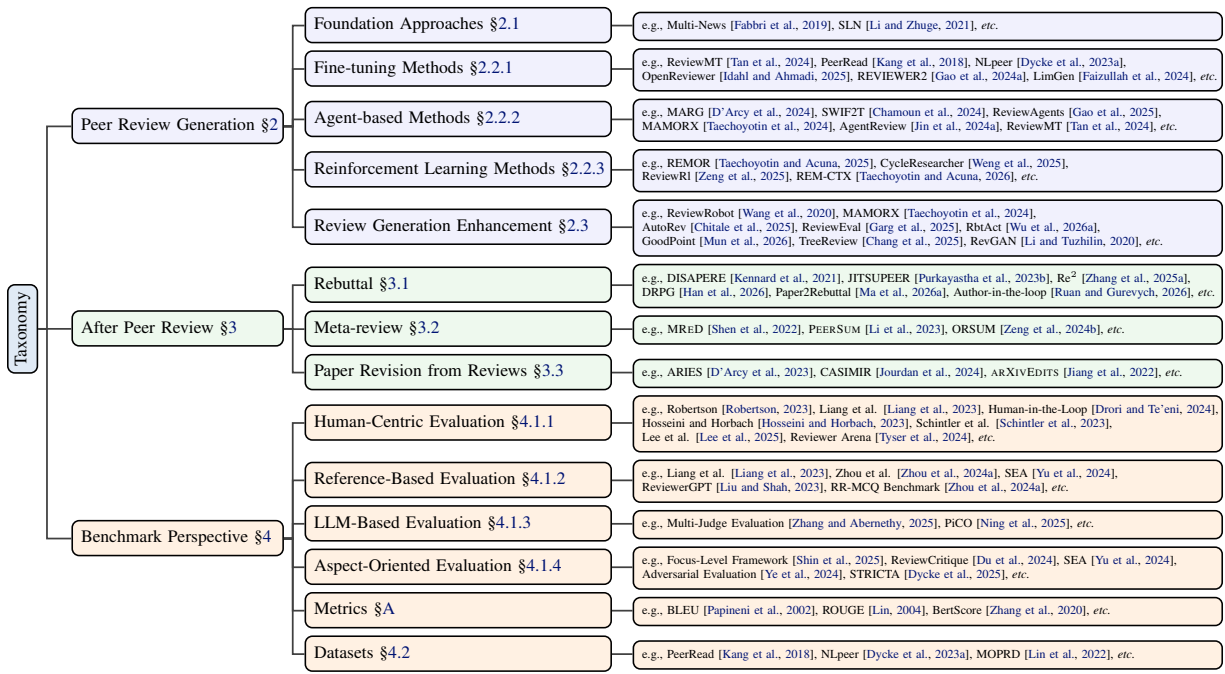
\begin{figure*}[t]
\centering
\input{figures/tree}
\vspace{-2mm}
\caption{Taxonomy of AI for Peer Review Process and Evaluation: Key Areas and Example Systems.}
\label{fig:taxonomy}
\end{figure*}

While other valuable surveys exist, they either cover the AI4Research landscape too broadly, with peer review as only a minor component~\cite{chen2025ai4researchsurveyartificialintelligence}, or due to the field's rapid progress, do not fully capture the latest wave of agentic and RL-based methodologies~\cite{Zhuang_2025}. Crucially, few existing works provide a systematic analysis of the critical landscape of evaluation for peer review generation. Our survey fills this gap by providing a holistic overview of peer review process,\footnote{This survey is intended to support research on AI-assisted peer review workflows, not to replace human reviewers.} with a dual focus on cutting-edge generation methodologies and a taxonomy of their evaluation. The review procedure is in Appendix~\ref{app:reviewprocedure}.


This survey is structured as follows: Section~\ref{sec:methodologies} details the evolution of peer review generation methodologies. Section~\ref{sec:after_peer_review} introduces methods of after-review tasks including rebuttal, meta-review and paper revision from reviews. Section~\ref{sec:benchmark_perspective} offers a systematic taxonomy of evaluation methods, metrics, and datasets. Section~\ref{sec:discussion} discusses key challenges and outlines future research directions. 


%% file: figures/tree.tex
\begin{adjustbox}{max width=\textwidth, center}
\begin{forest}
for tree={
  grow=east,
  rectangle,
  rounded corners,
  draw=black,
  base=left,
  font=\normalsize,
  edge+={darkgray, line width=1pt},
  s sep=4pt,
  inner sep=4pt,
  line width=0.9pt,
  forked edges,
  parent anchor=mid east,
  child anchor=mid west,
  anchor=mid west,
},
where level=0{
  calign=child,
  calign child=2,
  child anchor=mid west,
  anchor=center,
}{},
where level=1{
  minimum width=10em,
  text width=11em,
  align=center,
  parent anchor=mid east,
  child anchor=mid west,
  anchor=mid west,
}{},
where level=2{
  minimum width=14em,
  text width=16.4em,
  align=left,
  parent anchor=mid east,
  child anchor=mid west,
  anchor=mid west,
}{},
where level=3{
  minimum width=36em,
  text width=32.5em,
  align=left,
  font=\normalsize,
  tier=parent,
  parent anchor=mid east,
  child anchor=mid west,
  anchor=mid west,
}{},
[{\rotatebox{90}{Taxonomy}}, root-node
  [Benchmark Perspective \S\ref{sec:benchmark_perspective}, bench-node
  [Datasets \S\ref{sec:datasets}, bench-node
    [{e.g., PeerRead \cite{peerread}, NLpeer \cite{nlpee}, MOPRD \cite{lin2022moprd}, \textit{etc.}}, bench-leaf]
  ]
  [Metrics \S\ref{app:metrics}, bench-node
    [{e.g., BLEU \cite{bleu}, ROUGE \cite{rouge}, BertScore \cite{bertscore}, \textit{etc.}}, bench-leaf]
  ]
  [Aspect-Oriented Evaluation \S\ref{sec:aspect_oriented}, bench-node
    [{e.g., Focus-Level Framework \cite{shin2025mindblindspotsfocuslevel}, ReviewCritique \cite{du2024critique}, SEA \cite{yu2024automatedpeerreviewingpaper},\\ Adversarial Evaluation \cite{ye2024yetrevealingrisksutilizing}, STRICTA \cite{dycke-etal-2025-stricta}, \textit{etc.}}, bench-leaf]
  ]
  [LLM-Based Evaluation \S\ref{sec:llm_as_judge}, bench-node
    [{e.g., Multi-Judge Evaluation \cite{zhang2025reviewingscientificpaperscritical}, PiCO \cite{ning2025picopeerreviewllms}, \textit{etc.}}, bench-leaf]
  ]
  [Reference-Based Evaluation \S\ref{sec:reference_based}, bench-node
    [{e.g., Liang et al. \cite{liang2023largelanguagemodelsprovide}, Zhou et al. \cite{zhou-etal-2024-llm}, SEA \cite{yu2024automatedpeerreviewingpaper},\\ ReviewerGPT \cite{liu2023reviewergptexploratorystudyusing}, RR-MCQ Benchmark \cite{zhou-etal-2024-llm}, \textit{etc.}}, bench-leaf]
  ]
  [Human-Centric Evaluation \S\ref{sec:human_centric}, bench-node
    [{e.g., Robertson \cite{gpt4slightly}, Liang et al. \cite{liang2023largelanguagemodelsprovide}, Human-in-the-Loop \cite{Drori2024Human},\\ Hosseini and Horbach \cite{Hosseini2023Fighting}, Schintler et al. \cite{schintler2023criticalexaminationethicsaimediated},\\ Lee et al. \cite{Lee2025Role}, Reviewer Arena \cite{tyser2024aidrivenreviewsystemsevaluating}, \textit{etc.}}, bench-leaf]
  ]
]
  [After Peer Review \S\ref{sec:after_peer_review}, after-node
  [Paper Revision from Reviews \S\ref{sec:revision}, after-node
    [{e.g., ARIES \cite{darcy2023aries}, CASIMIR \cite{jourdan2024casimircorpusscientificarticles}, \textsc{arXivEdits} \cite{jiang2022arxiveditsunderstandinghumanrevision}, \textit{etc.}}, after-leaf]
  ]
  [Meta-review \S\ref{sec:metareview}, after-node
    [{e.g., \textsc{MReD} \cite{shen-etal-2022-mred}, \textsc{PeerSum} \cite{li-etal-2023-summarizing}, ORSUM \cite{zeng2024scientificopinionsummarizationpaper}, \textit{etc.}}, after-leaf]
  ]
  [Rebuttal \S\ref{sec:rebuttal}, after-node
    [{e.g., DISAPERE \cite{kennard2021disapere}, JITSUPEER \cite{purkayastha2023exploringjiujitsuargumentationwriting}, Re$^{2}$ \cite{zhang2025re2},\\ DRPG~\cite{han2026drpgdecomposeretrieveplan}, Paper2Rebuttal~\cite{ma2026paper2rebuttalmultiagentframeworktransparent}, Author-in-the-loop~\cite{ruan2026authorintheloopresponsegenerationevaluation}, \textit{etc.}}, after-leaf]
  ]
]
  [Peer Review Generation \S\ref{sec:methodologies}, gen-node
  [Review Generation Enhancement \S\ref{sec:enhancement}, gen-node
    [{e.g., ReviewRobot \cite{Wang_2020}, MAMORX \cite{mamorx}, \\ AutoRev \cite{chitale2025autorevautomaticpeerreview},  ReviewEval \cite{garg2024revieweval}, RbtAct~\cite{wu2026rbtactrebuttalsupervisionactionable}, \\ GoodPoint~\cite{mun2026goodpointlearningconstructivescientific}, TreeReview \cite{chang2025treereviewdynamictreequestions}, RevGAN \cite{li2020controllablepersonalizedreviewgeneration}, \textit{etc.}}, gen-leaf]
  ]
  [Reinforcement Learning Methods \S\ref{sec:rl}, gen-node
    [{e.g., REMOR \cite{remor}, CycleResearcher \cite{cycleresearcher},\\ ReviewRl \cite{zeng2025reviewrlautomatedscientificreview}, REM-CTX~\cite{taechoyotin2026remctxautomatedpeerreview}, \textit{etc.}}, gen-leaf]
  ]
  [Agent-based Methods \S\ref{sec:agent}, gen-node
    [{e.g., MARG \cite{marg}, SWIF2T \cite{chamoun-etal-2024-automated}, ReviewAgents \cite{reviewagents},\\ MAMORX \cite{mamorx}, AgentReview \cite{agentreview}, ReviewMT \cite{peerreviewas}, \textit{etc.}}, gen-leaf]
  ]
  [Fine-tuning Methods \S\ref{sec:fine-tuning}, gen-node
    [{e.g., ReviewMT \cite{peerreviewas}, PeerRead \cite{peerread}, NLpeer \cite{nlpee},\\ OpenReviewer \cite{openreview}, REVIEWER2 \cite{reviewer2}, LimGen \cite{limgen}, \textit{etc.}}, gen-leaf]
  ]
  [Foundation Approaches \S\ref{sec:foundation}, gen-node
    [{e.g., Multi-News \cite{multinews}, SLN \cite{8736808}, \textit{etc.}}, gen-leaf]
  ]
]
]
\end{forest}
\end{adjustbox}

%% file: main/4-methodologies.tex
\section{Peer Review Generation}
\label{sec:methodologies}
\begin{figure}[!t]
        \centering
	\includegraphics[width=0.75\linewidth]{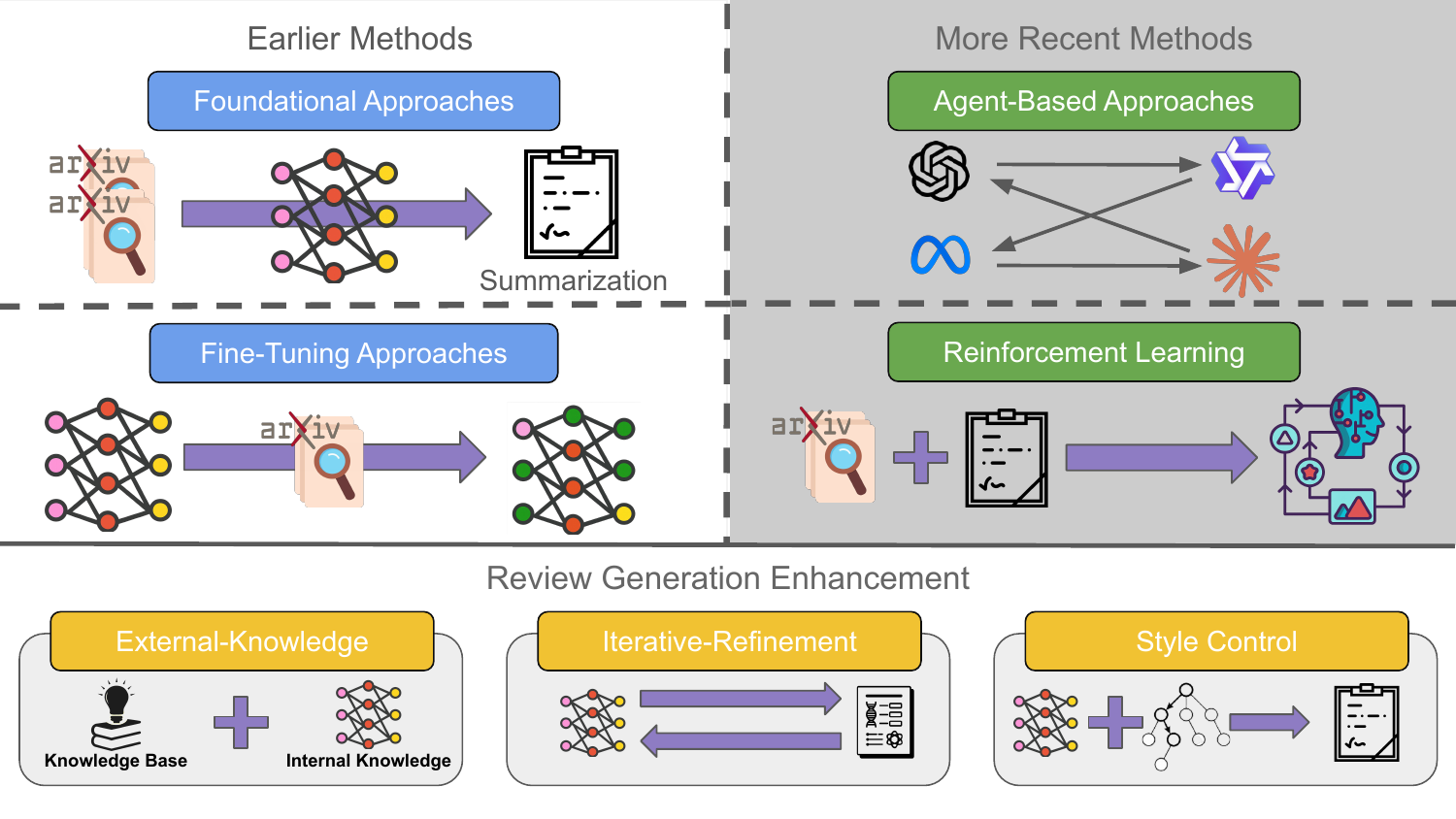}
	\caption{The methods of peer review generation: (1) Foundation approaches; (2) Fine-tuning methods; (3) Agent-based methods; (4) Reinforcement learning methods; and (5) Review Generation Enhancement.}
	\label{fig:methodologies}
\end{figure}

We categorize peer review generation methodologies into five main paradigms—foundation approaches, fine-tuning methods, agent-based methods, reinforcement learning methods and generation enhancement, as illustrated in 
Figure~\ref{fig:methodologies}. 

\input{tables/methodologies_summary}

\subsection{Foundation Approaches}
\label{sec:foundation}
Prior to the widespread adoption of LLMs, the automated generation of complete, high-quality peer reviews was difficult~\cite{Drozdz2024PeerReview}. Early attempts to generate reviews often addressed a different problem, such as the multi-document summarization of existing reviews~\cite{peersum,multinews}, or relied on citation networks rather than generating novel critique directly from a source manuscript~\cite{8736808}. Consequently, research in the pre-2023 era predominantly focused on a suite of more manageable sub-tasks. These foundational efforts deconstructed the peer review process into analytical~\cite{teufel-etal-1999-annotation}, predictive~\cite{peerread}, and narrowly generative components~\cite{dycke-etal-2023-overview}. While limited in their ability to produce complete reviews, these researches established critical datasets for subsequent breakthroughs.

\subsection{LLM-based Approaches}
We next review recent advancements in peer review generation using LLMs, including fine-tuning strategies, multi-agent systems, and reinforcement learning methods for optimizing review quality.

\subsubsection{Fine-tuning Methods}
\label{sec:fine-tuning}
To overcome the limitations of zero-shot prompting~\cite{Zero-shot,sivarajkumar2023empiricalevaluationpromptingstrategies}, such as the generation of overly positive and generic feedback and the inability to produce reviews that conform to the required format~\cite{peerreviewas,reviewer2}, research has rapidly pivoted towards fine-tuning LLMs on domain-specific data~\cite{hu2021loralowrankadaptationlarge}. Some of these datasets are directly derived from foundational datasets~\cite{peerread,nlpee}, while others are constructed by leveraging APIs and specialized tools to curate custom review datasets (we summarize methods in ~\autoref{app:data_collection})~\cite{dycke2022yesyesyesproactivedatacollection}.

The contemporary approach involves specializing LLMs to capture the distinct style and critical nature of peer review. To address the challenge of generating reviews in the required format, ~\citet{peerreviewas} has fine-tuned LLMs on the ReviewMT dataset. Experimental results demonstrate that such supervised fine-tuning substantially improves the review and decision hit rates compared to zero-shot prompting. 
OpenReviewer~\cite{openreview} fine-tunes a Llama-8B model~\cite{llama3modelcard} on a curated dataset of 79,000 expert reviews. Its high performance stemmed from curating reviews with diverse critique patterns such as methodological flaws, reproducibility issues and balanced scores. We also organize different LLM Backbone Strategies in~\autoref{tab:backbone_comparison} in Appendix.

Other frameworks decompose the generation task itself to improve specificity and control. REVIEWER2~\cite{reviewer2} utilizes a two-stage pipeline where one fine-tuned model first generates a set of relevant aspect prompts for a given paper, and a second model generates review text conditioned on these specific aspects. This decomposition helps 
lead to more detailed reviews that better cover the range of topics. The versatility of fine-tuning is further demonstrated by its application to specific sub-tasks, as seen in LimGen~\cite{limgen}, which fine-tunes models specifically for a task requires inferring potential weaknesses not explicitly stated by authors.

\subsubsection{Agent-based Methods}
\label{sec:agent}
Recognizing that peer review is a complex cognitive process, recent work has shifted towards agent-based systems. This paradigm decomposes the review process, assigning distinct roles and sub-tasks to multiple collaborating LLMs. These approaches can be broadly categorized by their primary objective: \emph{Task Decomposition} and \emph{Process Simulation}.

\paragraph{Task Decomposition.}
A significant body of work focuses on task decomposition to generate higher-quality reviews. ~\citet{marg} propose MARG, a framework utilizing a leader-worker architecture where specialized expert agents are tasked with critiquing specific aspects of a paper, such as its experiments, clarity, and impact. This division of labor helps produce more targeted, helpful feedback. Similarly, SWIF²T, introduced by \citet{chamoun-etal-2024-automated}, employs a four-component pipeline of Planner, Investigator, Reviewer, and Controller agents to generate focused and actionable feedback on specific weaknesses identified within a manuscript. Building on this concept of emulating human workflows, ~\citet{reviewagents} develop ReviewAgents, a framework designed to mirror the structured reasoning of human experts through summarization, analysis, and conclusion stages. DeepReview~\cite{deepreview} introduces a multi-stage framework that decomposes the review into three stages: novelty verification, multi-dimension review and reliability verification, aims to mitigate issues of limited domain expertise and hallucinated reasoning in LLM-based reviewers. Pushing the boundaries of analysis beyond text, ~\citet{mamorx} present MAMORX, a multi-agent system that integrates attention to text, figures, and citations together with external knowledge sources.
DIAGPaper~\cite{zou2026diagpaperdiagnosingvalidspecific} further decomposes weakness identification into criterion-grounded reviewing, rebuttal-based validation, and severity prioritization, improving the validity and specificity of detected weaknesses.

\paragraph{Process Simulation.}
A distinct but complementary line of research uses multi-agent systems to simulate and study the peer review process itself. ~\citet{agentreview} introduce AgentReview, an extensible simulation testbed populated by LLM agents representing authors, reviewers, and area chairs. In a similar vein, ~\citet{peerreviewas} reformulate peer review as a multi-turn, long-context dialogue among Reviewer, Author, and Decision Maker agents. This approach, supported by the large-scale ReviewMT dataset, explicitly models the iterative rebuttal and discussion cycles that are central to real-world academic review but are missed by generation models. Collectively, these multi-agent methodologies represent a significant leap in higher-quality reviews with dynamic simulation for deeper process understanding.

\subsubsection{Reinforcement Learning Methods}
\label{sec:rl}
Beyond supervised fine-tuning, which primarily teaches models stylistic and structural conventions, reinforcement learning (RL) has emerged as a crucial paradigm for optimizing generated content~\cite{bai2022traininghelpfulharmlessassistant, swamy2025roadsleadlikelihoodvalue}. This approach allows models to learn from feedback signals that represent complex goals, such as producing insightful and helpful scientific critiques. The application of RL in this domain shows a clear trajectory of optimizing the entire scientific process~\cite{rao2020rlcycleganreinforcementlearningaware,novikov2025alphaevolvecodingagentscientific}.

A direct application of this principle is seen in Remor~\cite{remor}, which employs multi-objective reinforcement learning to address the common failure mode of AI-generated reviews being shallow~\cite{wei2025aiimperativescalinghighquality}. The framework's core innovation is its Human-aligned Peer Review Reward, a composite function that quantifies multiple facets of review quality, including criticism, relevance, and actionable suggestions. By optimizing this multi-objective reward using Group Relative Policy Optimization (GRPO)~\cite{deepseekmath}, Remor learns to generate reviews that are better aligned with nuanced human preferences.

Expanding this paradigm, other frameworks use the generated review itself as a reward signal to refine the primary artifact. CycleResearcher~\cite{cycleresearcher} exemplifies this with a dual-agent system where a CycleReviewer model, trained to mimic human evaluation scores, provides a reward signal for a CycleResearcher model that generates manuscripts~\cite{meng2024simposimplepreferenceoptimization}. This establishes a closed "Research-Review-Refinement" loop~\cite{tang2025airesearcherautonomousscientificinnovation}.
ReviewRL~\cite{zeng2025reviewrlautomatedscientificreview} extends this line by combining retrieval-augmented context construction with composite-reward RL to jointly improve review quality and rating consistency. Some RL work also incorporates auxiliary context beyond the manuscript text itself, where REM-CTX~\cite{taechoyotin2026remctxautomatedpeerreview} uses correspondence-aware rewards to align generated reviews with figures and external scholarly signals.


\subsection{Review Generation Enhancement}
\label{sec:enhancement}

\paragraph{External Knowledge-Enhanced Generation.}
A significant challenge in peer review generation is ensuring that comments are not only coherent but also factually grounded to prevent hallucination problems~\cite{shuster2021retrievalaugmentationreduceshallucination,kovács2025lettucedetecthallucinationdetectionframework}. Models that generate reviews based solely on the text of a submitted paper may produce generic critiques, lacking the necessary context of the broader scientific landscape. To address this, a key line of research has focused on enhancing generation systems with external knowledge. These approaches parallel the wider adoption of Retrieval-Augmented Generation (RAG)~\cite{lewis2021retrievalaugmentedgenerationknowledgeintensivenlp,gao2024retrievalaugmentedgenerationlargelanguage}.
Early, pre-LLM work in this area pioneered the use of structured knowledge synthesis to generate evidence-backed reviews~\cite{Nauta_2023}. ReviewRobot~\cite{Wang_2020} operationalizes this by constructing three distinct knowledge graphs (KGs): one from the target paper, one from its cited works, and a background KG from a large collection of domain literature. With the advent of LLMs, methodologies have shifted towards more direct integration of external knowledge. MAMORX~\cite{mamorx} exemplifies this modern RAG paradigm by employing a specialized agent to query external scholarly databases to assess a paper's novelty. ~\citet{afzal2025notnovelenoughenriching} retrieves and re-ranks related literature, performing contribution-wise comparisons to produce novelty judgments.

\paragraph{Iterative Refinement.}This paradigm replaces the single-pass generation approach with multi-step processes where an initial output is progressively improved~\cite{kamoi-etal-2024-llms}. The refinement is driven by feedback generated either by the model itself or through interaction with other modules or agents~\cite{gou2024criticlargelanguagemodels, han2024smalllanguagemodelselfcorrect}.
~\citet{marg} operationalizes this with an explicit, final refinement stage. After aspect-specific agents generate initial comments, a separate multi-agent group is convened to assess each comment for validity and clarity, collaboratively deciding whether to revise or prune it. ~\citet{garg2024revieweval} proposes a self-refinement loop and  an external improvement loop that iteratively optimizes its intermediate outputs and the final reviews. Taking a different approach, ~\citet{peerreviewas} inherently creates an iterative refinement cycle where the author's rebuttal serves as feedback. ~\citet{remor} also employ an RL-based refinement strategy, but focus on generating the review itself. RbtAct~\cite{wu2026rbtactrebuttalsupervisionactionable} similarly leverages rebuttals as implicit supervision and preference signals to train generators that produce more actionable review feedback. GoodPoint~\cite{mun2026goodpointlearningconstructivescientific} also uses author responses to define constructive feedback as comments that are both valid and actionable. 
ActReview~\cite{actreview} formulates actionable peer review as a review-to-revision process with rubric-guided RL to generate actionable feedback.

\paragraph{Structure and Style Control.}
Structure refers to the logical organization, format, and argumentative framework of a review. Generating a review is a complex task that requires beyond insight into technical. TreeReview~\cite{chang2025treereviewdynamictreequestions} models peer review as a hierarchical question-answering process. The process begins with a top-down decomposition stage, where a high-level review task is recursively broken down into a tree of more fine-grained sub-questions. AutoRev~\cite{chitale2025autorevautomaticpeerreview} leverages graph-based structural modeling of the paper itself, capturing hierarchical relationships between passages.
The RevGAN model~\cite{li2020controllablepersonalizedreviewgeneration}, developed for generating controllable and personalized product reviews, provides a clear and powerful architecture for fine-grained stylistic control.


%% file: tables/methodologies_summary.tex
\begin{table*}[!htbp]
\centering
\setlength{\tabcolsep}{3pt}
\resizebox{\textwidth}{!}{%
\begin{tabular}{@{}lllll@{}}
\toprule
\textbf{Method \& Framework} & \textbf{Venue/Year} & \textbf{Core Paradigm} & \textbf{Dataset(s)} & \textbf{Key Contributions \& Features} \\
\midrule
\multicolumn{5}{l}{\textit{\textbf{Foundational Datasets}}} \\
\midrule
PeerRead~\citet{peerread} & NAACL 2018 & Data Collection & PeerRead & Large-scale papers+reviews dataset. \\
NLpeer~\citet{nlpee} & ACL 2023 & Data Collection & NLpeer & Unified, expanded peer-review resource. \\
MOPRD~\citet{lin2022moprd} & NCA 2023 & Data Collection & MOPRD & Multidisciplinary open peer-review set. \\
\midrule
\multicolumn{5}{l}{\textit{\textbf{LLM-based Generation Approaches}}} \\
\midrule
\multicolumn{5}{l}{\textbf{\textit{A. Fine-tuning Methods}}} \\
OpenReviewer~\citet{openreview} & NAACL Demo 2025 & Fine-tuning & 79k expert reviews & Llama-8B fine-tuned on curated expert reviews. \\
REVIEWER2~\citet{reviewer2} & Preprint 2024 & Fine-tuning & 27k papers, 99k reviews & Two stage: prompt generation and review. \\
LimGen~\citet{limgen} & ECML-PKDD 2024 & Fine-tuning & LimGen & SLG for limitation suggestions. \\
\midrule
\multicolumn{5}{l}{\textbf{\textit{B. Agent-based Systems}}} \\
MARG~\citet{marg} & Preprint 2024 & Agent-based (Dec.) & – & Leader–worker agents; iterative refinement. \\
ReviewAgents~\citet{reviewagents} & Preprint 2025 & Agent-based (Dec.) & Review-CoT & Summarize, analyze and conclude with RAG. \\
DeepReview~\citet{deepreview} & ACL 2025 & Agent-based (Dec.) & DeepReview-13K & Three stages: novelty, multi-dimension, reliability. \\
MAMORX~\citet{mamorx} & NeurIPS 2024 WS & Agent-based (Dec.) & – & Multimodal; function-calling RAG for novelty. \\
SWIF²T~\citet{chamoun-etal-2024-automated} & ACL Find. 2024 & Agent-based (Dec.) & 300 peer reviews & Four agents: plan/investigate/review/control. \\
DIAGPaper~\citet{zou2026diagpaperdiagnosingvalidspecific} & Preprint 2026 & Agent-based (Dec.) & AAAR, ReviewCritique & Multi-agent weakness diagnosis \\
AgentReview~\citet{agentreview} & EMNLP 2024 & Agent-based (Sim.) & – & Simulated peer review; authority-bias study. \\
ReviewMT~\citet{peerreviewas} & Preprint 2024 & Agent-based (Sim.) & ReviewMT & SFT $>$ zero-shot; rebuttal as dialogue. \\
\midrule
\multicolumn{5}{l}{\textbf{\textit{C. Reinforcement Learning (RL)}}} \\
Remor~\citet{remor} & Preprint 2025 & RL (GRPO) & PeerRT & Multi-objective RL with HPRR reward. \\
CycleResearcher~\citet{cycleresearcher} & ICLR 2025 & RL (SimPO) & Review-5k, Research-14k & Dual-agent research–review–refine loop. \\
ReviewRL~\citet{zeng2025reviewrlautomatedscientificreview} & EMNLP 2025 & RL (Rule-based) & ICLR 2025 papers & Composite-reward RL for grounded reviews. \\
REM-CTX~\citet{taechoyotin2026remctxautomatedpeerreview} & Preprint 2026 & RL (GRPO) & PeerRTEx, FCR/NCRDat & Auxiliary-context RL for grounded peer reviews. \\
\midrule
\multicolumn{5}{l}{\textit{\textbf{Review Generation Enhancement}}} \\
\midrule
ReviewRobot~\citet{Wang_2020} & INLG 2020 & RAG (KG-based) & – & Pre-LLM; synthesizes three knowledge graphs. \\
Novelty~\citet{afzal2025notnovelenoughenriching} & EACL 2026 & RAG (Novelty) & – & Contribution-wise novelty comparisons. \\
ReviewEval~\citet{garg2024revieweval} & EMNLP Find. 2025 & Iterative Refinement & FullCorpus-120 & Self-refinement + external improvement loops. \\
RbtAct~\citet{wu2026rbtactrebuttalsupervisionactionable} & ACL Find. 2026 & Iterative Refinement & RMR-75K & Rebuttal-derived optimization for actionability. \\
GoodPoint~\citet{mun2026goodpointlearningconstructivescientific} & Preprint 2026 & Iterative Refinement & GoodPoint-ICLR & Author-response supervision. \\
ActReview~\citet{actreview} & Preprint 2026 & Iterative Refinement & ActReview-40K, ActReview-Bench & Review-to-revision; rubric-guided RL. \\
TreeReview~\citet{chang2025treereviewdynamictreequestions} & EMNLP 2025 & Structure Control & Venue-derived benchmark & Dynamic question tree for analysis. \\
AutoRev~\citet{chitale2025autorevautomaticpeerreview} & Preprint 2025 & Structure Control & – & Document graph to target key passages. \\
RevGAN~\citet{li2020controllablepersonalizedreviewgeneration} & EMNLP 2019 & Style Control (GAN) & – & Controllable, personalized reviews (pre-LLM). \\
\bottomrule
\end{tabular}
}
\caption{A summary of methodologies in peer review generation. The table also highlights key contributions. ``(Dec.)'' denotes Task Decomposition and ``(Sim.)'' denotes Process Simulation.}
\label{tab:methodologies_summary}
\end{table*}

%% file: main/After_peer_review.tex
\section{After Peer Review}
\label{sec:after_peer_review}
Beyond generating reviews, several tasks model the post-review stage including rebuttal generation, meta-review generation and paper revision.
\subsection{Rebuttal}
\label{sec:rebuttal}
Rebuttal generation aims to produce author responses that directly and faithfully address reviewer critiques~\cite{gao-etal-2019-rebuttal,Huang_2023}. Early resources focused on linking reviews to rebuttals and labeling discourse functions, enabling supervised conditioning of responses on critique aspects and actions~\cite{kennard2021disapere} and argument-pair extraction between review claims and rebuttal~\cite{cheng-etal-2020-ape}.
The first work to explicitly formulate rebuttal generation is~\cite{purkayastha2023exploringjiujitsuargumentationwriting}, who introduce JITSUPEER and cast the task as attitude root or theme-guided generation of canonical rebuttals conditioned on rebuttal actions.
Recent datasets move from single-turn to multi-turn dialog settings by involving rebuttal: ReviewMT reframes peer review as long-context, role-based interaction~\cite{peerreviewas}, and Re$^{2}$~\cite{zhang2025re2} aggregates review data into structured multi-turn rebuttal discussions.
More recently, the focus is shifting from simple sentence generation to strategic, evidence-based dialog planning.
DRPG~\cite{han2026drpgdecomposeretrieveplan} formalizes this shift through a four-stage pipeline of concern decomposition, evidence retrieval, perspective planning, and response generation.
Paper2Rebuttal~\cite{ma2026paper2rebuttalmultiagentframeworktransparent} further reframes rebuttal assistance as a verify-then-write evidence organization problem with explicit checkpoints for grounding and consistency.
Recent author-in-the-loop work~\cite{ruan2026authorintheloopresponsegenerationevaluation} also shows that effective response generation benefits from explicit author signals, controllable planning, and evaluation-guided refinement.
Even minimal author guidance can improve factual correctness and targeted refutation over direct generation baselines~\cite{khatri2026defendautomatedrebuttalspeer}.

\subsection{Meta-review}
\label{sec:metareview}
Meta-review generation synthesizes multiple reviewer opinions into a summary for area chairs. Early systems framed the task as extract-then-write with predicting the accept/reject and conditioning generation on it~\cite{Bhatia,Kumar}. Public datasets have enabled further study. \textsc{MReD} adds sentence-level functional labels to meta-reviews for structure-controllable generation~\cite{shen-etal-2022-mred}; \textsc{PeerSum} models the hierarchical conversational structure of reviews–rebuttals and introduces \textsc{RAMMER} for structure-aware aggregation~\cite{li-etal-2023-summarizing}. ORSUM scales coverage across venues while proposing checklist-guided, multi-stage introspection for opinion consolidation and evaluation~\cite{zeng2024scientificopinionsummarizationpaper}. Methods increasingly leverage argument structure and rebuttal content to surface consensus vs.\ controversy~\cite{wu2022prrca}, and decompose generation via facet-level judgement extraction and sentiment consolidation to improve decision quality~\cite{li-etal-2024-sentiment}.
~\citet{purkayastha2026decisionmakingdeliberationmetareviewingdocumentgrounded} also reframes meta-reviewing as a document-grounded dialogue problem, emphasizing deliberation and decision support beyond summarization alone.

\subsection{Paper Revision from Reviews}
\label{sec:revision}
A further step is editing the manuscript itself based on peer feedback. ARIES~\cite{darcy2023aries} introduces the task and a dataset aligning review comments to concrete paper edits, enabling edit generation. CASIMIR~\cite{jourdan2024casimircorpusscientificarticles} compiles and analyzes papers and their reviews to understand the intent behind edits, which helps plan and evaluate future revisions. \textsc{arXivEdits}~\cite{jiang2022arxiveditsunderstandinghumanrevision} provides gold sentence alignments across versions and fine-grained span-level edit intents, complementing review-linked corpora. 

%% file: main/3-evaluation.tex
\section{Benchmark Perspective}
\label{sec:benchmark_perspective}
\begin{figure}[!t]
  \centering
  \includegraphics[width=0.75\linewidth]{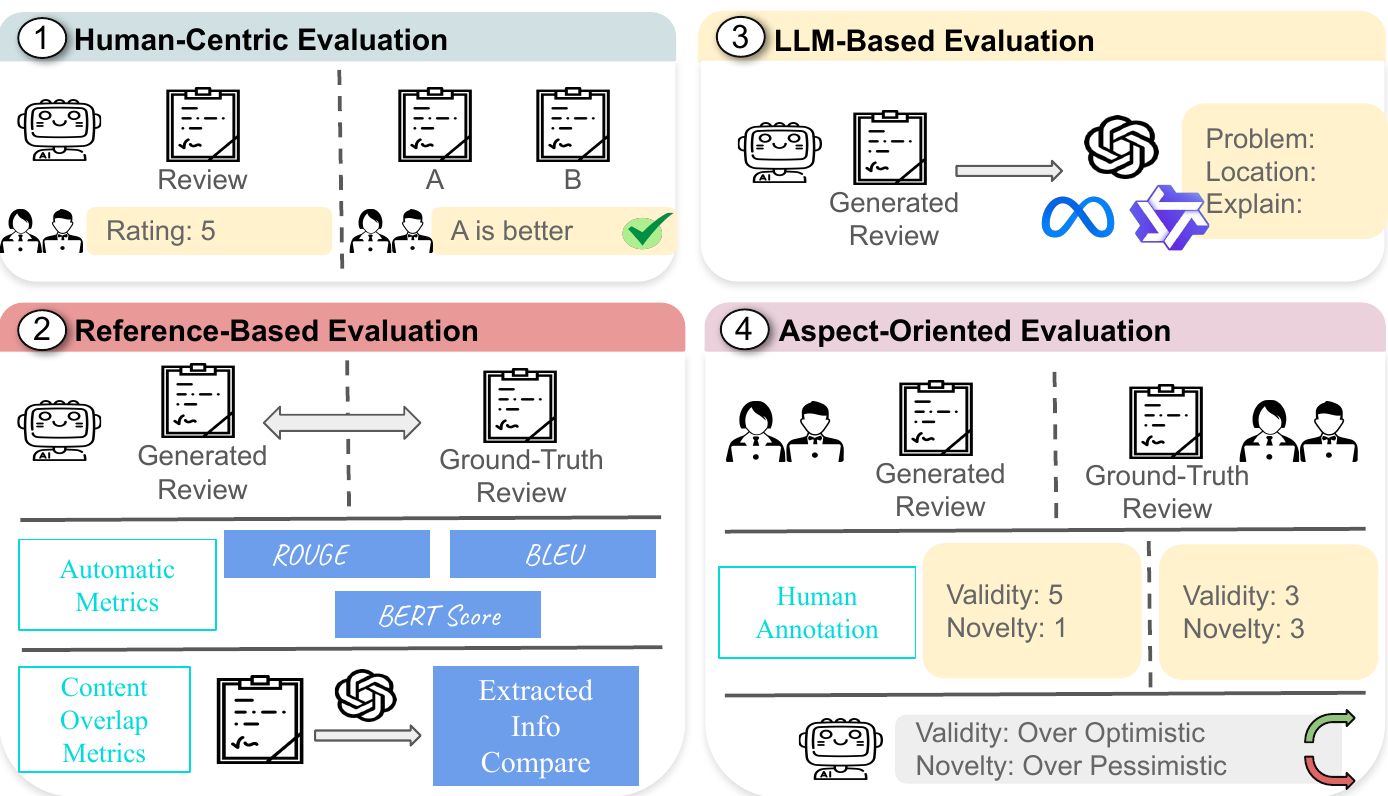}
  \caption{The main evaluation methods that we discuss are: (1) Human-centric evaluation; (2) Reference-based automated evaluation; (3) LLM-based evaluation; and (4) Aspect-oriented evaluation.}
  \label{fig:benchmark}
\end{figure}

Evaluating generated reviews remains challenging due to the subjective nature of review quality. Ensuring reliability is essential for maintaining the integrity of academic publishing~\cite{zhou-etal-2024-llm}.
\subsection{Evaluation Methods}
The validity, trustworthiness, and ultimate utility of any automated review generation system hinge entirely on the rigor of its evaluation. A poorly evaluated system risks amplifying biases, providing superficial feedback, or undermining the very scientific integrity it aims to support~\cite{zhang2025reviewingscientificpaperscritical, tyser2024aidrivenreviewsystemsevaluating, lee2013biaspeerreview, bender2021dangers}. Consequently, the methods used to evaluate these systems are as important as the systems themselves. This review provides a systematic taxonomy of the evaluation methodologies, 
in~\autoref{fig:benchmark}
, that have emerged in this rapidly developing field, with their pros and cons, which is also shown in~\autoref{tab:evaluation_proscons} and~\autoref{app:evaluation_summary}.

\subsubsection{Human-Centric Evaluation}
\label{sec:human_centric}
The most straightforward method for evaluating a generated peer review is to ask a human expert to assess its quality. 
~\citet{gpt4slightly} conducted the first pilot study comparing GPT-4 to human reviewers: ten participants rated the helpfulness of each review, finding identical mean scores for GPT-4 and human reviews, although GPT-4’s responses showed higher variance. 
Expanding on this approach, ~\citet{liang2023largelanguagemodelsprovide} conducted a large-scale user study involving 308 researchers who received LLM-generated feedback on their own papers, finding that users perceived the feedback as valuable. ~\citet{Drori2024Human} structured GPT-4 reviews against standard review criteria and concluded that AI assistance can meaningfully reduce reviewer workload but is “not completely” reliable for all cases. However, qualitative critiques warn of risks: ~\citet{Hosseini2023Fighting} reviews the potential benefits and risks of using LLMs in academic peer review, highlighting both efficiency gains and concerns about bias and reproducibility, and ~\citet{schintler2023criticalexaminationethicsaimediated} emphasize ethical concerns that unchecked AI could compromise review integrity. In a domain-specific review, ~\citet{Lee2025Role} highlight that LLMs excel at language tasks (screening, summarization, language editing) but struggle with assessing scientific validity, recommending they be used cautiously as aids under clear guidelines.
Recognizing the difficulty of assigning absolute numerical scores, a parallel approach has gained prominence: leveraging relative human preference. This principle has given rise to ``Arena'' platforms~\cite{chiang2024chatbotarenaopenplatform, zhao2025sciarenaopenevaluationplatform}. ~\citet{tyser2024aidrivenreviewsystemsevaluating} employs an arena-based methodology called Reviewer Arena, leveraging pairwise human (and LLM-predicted) preference comparisons to evaluate the quality of LLM-generated academic reviews.

\input{tables/evaluationproscons}

\subsubsection{Reference-Based Evaluation}
\label{sec:reference_based}
To overcome the scalability limitations of human-centric evaluation, researchers have long relied on automated metrics.
The evaluation assumes that a higher-quality generated review will exhibit greater lexical and semantic overlap with its human counterparts. This approach was a first step, leveraging metrics like BLEU, ROUGE, and BERTScore~\cite{bleu,rouge,rouge2,bertscore}. We provide a detailed summary of the metrics used for peer review generation evaluation in ~\autoref{app:metrics}. 
\citet{liang2023largelanguagemodelsprovide} propose a two-stage evaluation: first extracting key comments from reviews with GPT-4, then semantically matching them to compute a content overlap ``hit rate'', offering a more nuanced assessment than document-level metrics. ~\citet{zhou-etal-2024-llm} employed a suite of metrics including ROUGE and BERTScore to compare generated reviews against human references. The SEA framework proposed by ~\citet{yu2024automatedpeerreviewingpaper} also utilizes these metrics and highlights that ROUGE is indicative of the key information.
These metrics provide a scalable, automated way to assess surface-level quality but are limited in their ability to evaluate deeper aspects like correctness or constructiveness~\cite{asano-etal-2017-reference}.
Other work compares LLM outputs to known answers. \citet{liu2023reviewergptexploratorystudyusing} evaluate LLMs on targeted tasks such as error detection and checklist verification~\cite{Buyukkaramikli2019TECHVER}. ~\citet{zhou-etal-2024-llm} created an RR-MCQ benchmark and tested LLMs on multiple-choice questions about papers. Both studies find while LLMs achieve reasonable accuracy, they still struggle with complex review tasks and critical feedback.

\subsubsection{LLM-Based Evaluation}
\label{sec:llm_as_judge}
The LLM-Based paradigm has emerged as a scalable alternative to human-centric and reference-based evaluation~\cite{gu2025surveyllmasajudge}. ~\citet{zhang2025reviewingscientificpaperscritical} used a multi-judge pipeline where two distinct LLM judges must agree for a case to count as a ``hit'', reducing single-model bias.  
Recent work also explores unsupervised evaluation without gold labels~\cite{lin2023llmevalunifiedmultidimensionalautomatic, allamanis2024unsupervisedevaluationcodellms, ning2025picopeerreviewllms}.

\subsubsection{Aspect-Oriented Evaluation}
\label{sec:aspect_oriented}
Moving beyond generic similarity, the aspect-oriented paradigm deconstructs review quality into specific, fine-grained dimensions such as correctness, clarity, substance, and impact~\cite{lu2025identifyingaspectspeerreviews}. Some work evaluates reviews by examining specific facets or error types. ~\citet{shin2025mindblindspotsfocuslevel} develop a focus-level framework: they annotate review content by targets and aspects, then compare where LLM reviews concentrate versus human reviews. ~\citet{du2024critique} released ReviewCritique, a dataset annotated with 23 fine-grained error types at the sentence level, such as ``Unstated Statement'', ``Missing Reference'', testing whether an LLM-generated review flags the same issues as humans. STRICTA~\cite{dycke-etal-2025-stricta} decomposes peer review into a graph of aspect-level reasoning steps, annotated by experts, thereby offering interpretable evaluation beyond black-box scoring. A complementary line evaluates review utility from the author’s perspective through comment-level dimensions~\cite{sadallah2025goodbadconstructiveautomatically}.
Adversarial testing can be seen as a specific form of aspect-oriented evaluation focused on robustness and vulnerability~\cite{zhang2025adversarialtestingllmsinsights,liao2025redteamcuarealisticadversarialtesting}. \citet{ye2024yetrevealingrisksutilizing} found that LLMs are vulnerable to explicit manipulation through review injection attacks and to implicit manipulation by overemphasizing minor limitations.
\citet{yu2024automatedpeerreviewingpaper} introduced a novel metric called the mismatch score as part of their SEA framework, predicting the degree of inconsistency between a given paper and a review.


\subsection{Datasets}
\label{sec:datasets}
We collected datasets for peer review across two broad eras: pre-2023 and post-2023, which is shown in~\autoref{tab:datasets}. Pre-2023 resources such as \textsc{PeerRead}~\cite{peerread} laid the groundwork for review tasks. These early datasets primarily enabled score prediction, acceptance classification, and basic review generation tasks.
Since 2023, a wave of new datasets has significantly expanded the scope and depth. Compared to earlier corpora, these newer datasets are more diverse, task-specific, and comprehensive.

Despite this progress, we discuss several key challenges remain in benchmark perspective in Appendix~\ref{app:benchmarkdiscussion}, along with recommendations for improving dataset quality.



%% file: tables/evaluationproscons.tex
\begin{table*}[!ht]
\centering
\footnotesize
\begin{tabularx}{\textwidth}{>{\raggedright\arraybackslash}p{2.5cm} X X}
\toprule
\textbf{Methodology} & \textbf{Pros} & \textbf{Cons} \\
\midrule

\textbf{Human-Centric} & 
\textbullet{} \textbf{Most Direct Assessment:} Best captures nuanced qualities such as helpfulness, insightfulness, and practical utility. \newline
\textbullet{} \textbf{Strong Reliability in Pairwise Settings:} Relative preference judgments are often more consistent than absolute ratings. &
\textbullet{} \textbf{Limited Scalability:} Expensive and time-consuming, especially when expert reviewers are required. \newline
\textbullet{} \textbf{Subjectivity and Bias:} Prone to inter-annotator disagreement, personal bias, and rubric inconsistency. \\

\midrule

\textbf{Reference-Based} & 
\textbullet{} \textbf{Scalable and Efficient:} Metrics such as ROUGE and BERTScore are fast and easy to apply at scale. \newline
\textbullet{} \textbf{Reproducible:} Produces deterministic scores that are easy to compare across systems. \newline
\textbullet{} \textbf{Useful for Targeted Subtasks:} Can support focused evaluation such as MCQ answering or error detection. &
\textbullet{} \textbf{Shallow Assessment:} Often captures lexical or semantic overlap rather than scientific validity, reasoning quality, or constructiveness. \newline
\textbullet{} \textbf{Reference Dependency:} Relies heavily on the quality and coverage of human-written references, and may penalize valid but differently phrased feedback. \\

\midrule

\textbf{LLM-Based} &
\textbullet{} \textbf{Scalable and Nuanced:} Combines automation with the ability to assess higher-level qualities using rubric-based prompting. \newline
\textbullet{} \textbf{Reduced Reference Dependence:} Can evaluate outputs without requiring gold reviews for every instance. \newline
\textbullet{} \textbf{Flexible:} Evaluation criteria can be adapted easily through natural language instructions. &
\textbullet{} \textbf{LLM Judge Biases:} Susceptible to position bias, verbosity bias, and style preference. \newline
\textbullet{} \textbf{Reliability Concerns:} Sensitive to judge model choice, prompt design, and calibration, and may still fail on complex scientific reasoning. \newline
\textbullet{} \textbf{Cost:} API-based evaluation can be expensive at scale. \\

\midrule

\textbf{Aspect-Oriented} & 
\textbullet{} \textbf{Fine-Grained Diagnosis:} Reveals strengths and weaknesses along specific dimensions such as novelty, clarity, or evidence use. \newline
\textbullet{} \textbf{Supports Targeted Improvement:} Helps identify concrete failure modes for model development. \newline
\textbullet{} \textbf{Specialized Testing:} Enables focused evaluation such as robustness or deficiency-specific analysis. &
\textbullet{} \textbf{High Annotation Overhead:} Often requires task-specific schemas and manually annotated data. \newline
\textbullet{} \textbf{Harder to Aggregate:} Produces multidimensional profiles rather than a single summary score, which can complicate direct model comparison. \\

\bottomrule
\end{tabularx}
\caption{Pros and cons of evaluation methods for peer review generation. Each paradigm reflects a different trade-off between scalability, cost, and evaluative depth.}
\label{tab:evaluation_proscons}
\end{table*}

%% file: main/5-discussion_future_directions.tex
\section{Discussion and Future Directions}
\label{sec:discussion}

\paragraph{\ding{172} Novelty Evaluation.}
While LLMs can fluently mimic reviewer tone, they remain weak at judging true novelty. Early systems like \textsc{OpenReviewer}~\cite{openreview} and \textsc{Reviewer2}~\cite{reviewer2} generate coherent critiques but fail to distinguish incremental from groundbreaking work without broader scientific context. Recent efforts explicitly tackle novelty evaluation in peer review: SchNovel creates benchmark for novelty judgments~\cite{lin2024evaluatingenhancinglargelanguage}, and structured pipelines achieve high agreement with human reviewers in distinguishing genuine contributions~\cite{afzal2025notnovelenoughenriching}. Other directions integrate literature graphs for contribution scoring~\cite{rubaiat2025mappingevolutionresearchcontributions}. NovBench~\cite{wu2026novbenchevaluatinglargelanguage} also isolates textual novelty assessment as a dedicated evaluation problem, further highlighting the difficulty current LLMs face in producing reliable novelty judgments. 

\paragraph{\ding{173} Automated Evaluation.}
Review quality remains hard to quantify. While models like \textsc{MARG}~\cite{marg} and \textsc{SWIF$^2$T}~\cite{chamoun-etal-2024-automated} decompose review generation into targeted components, evaluation is often limited to coarse metrics like BLEU. New datasets such as \textsc{ReviewCritique}~\cite{du2024critique} and \textsc{SubstanReview}~\cite{guo2023automatic} offer fine-grained labels that allow content-aware judge. Future research could combine sentence-level assessments with reviewer consistency and informativeness benchmarks.


\paragraph{\ding{174} Beyond NLP and AI Domains.}
Most peer review datasets to date are sourced from NLP and ML conferences (e.g., PeerRead~\cite{peerread}, NLPEER~\cite{nlpee}), limiting generalization. Multidisciplinary resources such as \textsc{MOPRD}~\cite{lin2022moprd} are a welcome step, but broader coverage is necessary to evaluate whether current models overfit to domain-specific language.

\paragraph{\ding{175} Beyond Review Generation.}
While peer review generation and evaluation have received considerable attention~\cite{deepreview}, the automation of subsequent steps in the review process remains underexplored. Tasks such as rebuttal generation~\cite{gao-etal-2019-rebuttal,purkayastha2023exploringjiujitsuargumentationwriting}, meta-review drafting~\cite{li-etal-2023-summarizing}, and paper revision from reviews~\cite{darcy2023aries} involve richer discourse structures and complex reasoning about critique-response dynamics. Compared to review generation, these tasks demand deeper alignment with argumentative discourse and collaborative intent, yet existing benchmarks and methodologies are sparse. Addressing these underdeveloped areas opens new challenges for dataset construction, evaluation, and the design of models that can faithfully support the full peer review process.

\paragraph{\ding{176} Multimodal Review Tasks.}
Current review models operate on textual inputs, yet real submissions often include figures, tables, or code. Multimodal systems such as \textsc{MAMORX}~\cite{mamorx} extend the frontier by analyzing visual and tabular components, offering richer critiques. This line of work should be expanded with datasets that pair visual artifacts and expert annotations. We further discuss this task in Appendix~\ref{app:multimodalinput}.

\paragraph{\ding{177} Ethical and Transparent Deployment.}
A primary concern is the potential for these models to introduce and amplify biases. For instance, studies have shown that LLMs can exhibit affiliation bias, favoring authors from highly-ranked institutions in single-blinded settings~\cite{wedel2024affiliation}. Furthermore, a corpus-level study estimates that 6.5-16.9\% of review text at top AI venues was likely substantially modified by LLMs (beyond simple grammar fixes), with usage spiking near deadlines and among low-confidence reviews~\cite{liang2024monitoringaimodifiedcontentscale}.
It is essential for the research community to establish clear guidelines for the ethical use of LLMs in peer review. This includes a call for transparency, where the use of an LLM in any part of the review process is explicitly disclosed. 

%% file: main/6-conclusion.tex
\section{Conclusion}
The landscape of peer review process has undergone a transformative shift, driven by advances in large language models and latest paradigms. This survey has provided a comprehensive synthesis of state-of-the-art methods, spanning review generation as well as after-review tasks including rebuttal, meta-review, and revision. We also present a systematic taxonomy of evaluation methods, metrics and datasets. Despite remarkable improvements in generation methods, significant challenges remain, including robust content-level evaluation and extensive capabilities. As the field advances, future research must prioritize transparent and ethical deployment of automated reviews.


%% file: main/limitations.tex
\section*{Limitations}
While this survey strives to provide a comprehensive and up-to-date overview of the peer review landscape, several limitations remain. First, the rapid pace of progress in large language models, agent-based systems, and evaluation protocols means that new methodologies and benchmarks may emerge soon after publication, potentially outdating some of our analyses. Second, the majority of available datasets and evaluation studies are concentrated in NLP and machine learning domains, limiting the generalizability of our findings to other scientific fields. Finally, our survey is based on publicly accessible literature and resources; proprietary systems or unpublished industrial advances may not be adequately represented. Despite these limitations, we believe this survey remains the most comprehensive and fine-grained taxonomy of peer review process and its evaluation to date.

\section*{Acknowledgments}
This work was supported in part by the Google Research Scholar Program and by a research grant from TCS Research.

%% file: appendix/metrics.tex
\section{Evaluation Metrics}
\label{app:metrics}
Evaluating the quality of generated peer reviews is a multifaceted and non-trivial task. The field has developed a diverse array of evaluation metrics, which have evolved from standard text-similarity measures to more sophisticated, multi-dimensional frameworks that aim to capture the functional utility of a review. These metrics can be broadly categorized into five facets: reference-based text similarity, set overlap, aspect- and focus-level analysis, task-based proxies, and unsupervised or reward-based methods.

\subsection{Reference-based Text Similarity}
Initial approaches to evaluating generated reviews, particularly those framing the task as a form of summarization, relied on standard metrics that measure the similarity between a generated review and a ground-truth human review. These metrics provide a scalable, automated way to assess surface-level quality.

\vspace{0.5em}
\textbf{BLEU (Bilingual Evaluation Understudy):} 
BLEU is a widely used metric for evaluating the quality of generated text by comparing it to one or more reference texts. It computes the degree of n-gram overlap, prioritizing precision (the proportion of n-grams in the candidate that appear in the reference) and incorporates a brevity penalty to discourage overly short outputs. The BLEU score is calculated as:
\[
\mathrm{BLEU} = \mathrm{BP} \cdot \exp\left(\sum_{n=1}^N w_n \log p_n \right)
\]
where $p_n$ denotes the modified n-gram precision for n-grams of size $n$, $w_n$ are the weights (usually uniform), and $\mathrm{BP}$ is the brevity penalty, defined as $1$ if $c > r$ or $\exp(1-r/c)$ if $c \leq r$, with $c$ and $r$ being the lengths of the candidate and reference texts, respectively. BLEU scores range from 0 to 1, with higher values indicating greater similarity to the reference.

\vspace{0.5em}
\textbf{BERTScore:}
BERTScore evaluates the similarity between generated and reference texts using contextual embeddings from pre-trained language models such as BERT. Instead of relying on exact n-gram matches, it computes token-level similarity via cosine similarity in the embedding space, capturing semantic similarity. The score is calculated by aligning each token in the candidate with the most similar token in the reference and averaging the resulting similarities. BERTScore reports precision, recall, and F1 scores, offering a more nuanced assessment of meaning preservation than traditional lexical overlap metrics.

\vspace{0.5em}
\textbf{ROUGE (Recall-Oriented Understudy for Gisting Evaluation):}
ROUGE-L evaluates the correspondence between a generated sequence and a reference sequence by identifying their Longest Common Subsequence (LCS). This metric emphasizes the preservation of word order without requiring strict consecutive matches.

Recall is calculated as the fraction of the LCS length over the total length of the reference sentence:
\[
R = \frac{\mathrm{LCS}(\text{Generated},\, \text{Reference})}{\lvert\text{Reference}\rvert}
\]

Precision is similarly defined, using the length of the generated sequence as the denominator:
\[
P = \frac{\mathrm{LCS}(\text{Generated},\, \text{Reference})}{\lvert\text{Generated}\rvert}
\]

To balance both precision and recall, the $F_1$ score is computed as follows:
\[
F_1 = \frac{(1+\beta^2)\, P\, R}{\beta^2 P + R}
\]
where $\beta$ allows for adjustable weighting between recall and precision. ROUGE-L scores reflects the degree of sequence similarity: a score closer to 1 indicates a higher overlap between the candidate and reference sequences.
 ROUGE scores are typically reported as recall, precision, and F1.

\vspace{0.5em}
\textbf{METEOR (Metric for Evaluation of Translation with Explicit ORdering):}
METEOR evaluates machine-generated text by aligning it with reference texts at the word level, accounting for exact matches, stem matches, synonym matches, and paraphrase matches. The metric calculates unigram precision and recall, combines them into an F-score, and applies a fragmentation penalty to account for differences in word order. The METEOR score is given by:
\[
\mathrm{METEOR} = F_{\text{mean}} \cdot (1 - \mathrm{Pen})
\]
where $F_{\text{mean}}$ is a weighted harmonic mean of unigram precision and recall, and $\mathrm{Pen}$ is a penalty based on the number of chunks (contiguous matched subsequences). METEOR is designed to better correlate with human judgments than simple n-gram-based metrics.

\subsection{Set Overlap Metrics}
Recognizing that a review is a collection of distinct ideas or comments, some evaluation frameworks have moved beyond holistic text similarity to measure the overlap of these discrete points. This approach treats the generated and reference reviews as sets of comments and assesses their intersection.

\vspace{0.5em}
\textbf{Recall:}
Recall measures the proportion of relevant items in the reference set that are successfully retrieved by the model. For review generation, it reflects the fraction of human-written comments that are also captured by the generated review. The formula is:  
\[
\mathrm{Recall} = \frac{|\text{Matched}_{\text{Gen} \to \text{Ref}}|}{|\text{Reference}|}
\]
where $|\text{Matched}_{\text{Gen} \to \text{Ref}}|$ is the number of reference comments matched by generated comments, and $|\text{Reference}|$ is the total number of human-written reference comments. Recall ranges from 0 to 1, with higher values indicating more comprehensive coverage of reference content.

\vspace{0.5em}
\textbf{Precision:}
Precision evaluates the proportion of generated items that are relevant, i.e., the fraction of generated comments that can be aligned with human reference comments. The formula is:  
\[
\mathrm{Precision} = \frac{|\text{Matched}_{\text{Gen} \to \text{Ref}}|}{|\text{Generated}|}
\]
where $|\text{Matched}_{\text{Gen} \to \text{Ref}}|$ is the number of generated comments matched to the reference, and $|\text{Generated}|$ is the total number of generated comments. Higher precision indicates that most generated comments are meaningful and consistent with human feedback.

\vspace{0.5em}
\textbf{Jaccard Index:}
The Jaccard index measures the similarity between the generated and reference sets by dividing the size of their intersection by the size of their union. In the review context, it quantifies how many unique comments are shared between the two sets. The formula is:  
\[
\mathrm{Jaccard} = \frac{|\text{Matched}_{\text{Gen} \to \text{Ref}}|}{|\text{Reference} \cup \text{Generated}|}
\]
Here, $|\text{Matched}_{\text{Gen} \to \text{Ref}}|$ is the number of matched comments, and $|\text{Reference} \cup \text{Generated}|$ is the total number of unique comments across both sets. The Jaccard index ranges from 0 to 1, with higher values indicating greater overlap.

\vspace{0.5em}
\textbf{Comment Matching (LLM-based):} 
To determine whether a generated comment matches a reference comment, a large language model (e.g., GPT-4) may be employed to assess semantic relatedness and relative specificity. Only pairs with high relatedness and sufficient specificity are counted as matches, ensuring that low-utility or generic comments do not inflate metric scores. This LLM-based matching provides a more nuanced, semantically grounded evaluation than traditional surface-level overlap.

\subsection{Aspect and Focus Level Metrics}
A significant advancement in evaluation has been the shift towards more diagnostic, fine-grained analysis. Instead of producing a single quality score, these methods dissect reviews to understand what they are about and how they critique the paper.

\vspace{0.5em}
\textbf{F1 Score:} 
The F1 score is a widely-used metric that balances precision and recall for classification tasks. In the context of aspect-based coding schemes for peer review generation, it evaluates the accuracy of a model in assigning the correct (Target, Aspect) label to each review point, compared to expert annotations. The F1 score is defined as the harmonic mean of precision ($P$) and recall ($R$):
\[
F_1 = \frac{2PR}{P + R}
\]
where precision is the proportion of correctly predicted labels among all predicted labels, and recall is the proportion of correctly predicted labels among all true labels. The F1 score ranges from 0 to 1, with higher values indicating better classification performance.

\vspace{0.5em}
\textbf{Kullback-Leibler (KL) Divergence:} 
KL Divergence is a distributional metric that quantifies the difference between two probability distributions. In peer review evaluation, it is used to compare the focus distribution of LLM-generated reviews with that of human-written reviews—i.e., how attention is allocated across various targets and aspects. The KL Divergence from the human distribution $P$ to the model distribution $Q$ is given by:
\[
D_{\mathrm{KL}}(P \parallel Q) = \sum_{i} P(i) \log \frac{P(i)}{Q(i)}
\]
where $P(i)$ and $Q(i)$ are the probabilities assigned to target-aspect pair $i$ by the human and model distributions, respectively. Lower values of KL Divergence indicate greater alignment between model and human focus.

\subsection{Task-Based Proxy Metrics}
Another evaluation strategy is to measure a model's performance on tasks that are proxies for the abilities required to write a good review. Instead of assessing the generated text directly, these metrics evaluate the model's capacity for judgment and deep comprehension.

Score Prediction: This task assesses whether a model can predict the final outcome of the peer review process. This can involve a binary classification task of predicting paper acceptance or a regression task of predicting the numerical scores for specific criteria like 'originality' or 'impact'. Performance is measured using standard metrics like accuracy for classification or Mean Absolute Error (MAE) for regression.

Multiple-Choice Question Answering: To probe deeper comprehension, specialized datasets have been created. For example, the RR-MCQ dataset contains multiple-choice questions derived from real review-rebuttal interactions. Performance is measured by accuracy, with results indicating that while current models perform better than random.

\textbf{Accuracy:} 
Accuracy measures the proportion of correctly predicted instances among all instances. In peer review generation, it is used both for classification tasks (e.g., predicting paper acceptance) and for multiple-choice question answering. Its formula is:
\[
\mathrm{Accuracy} = \frac{\text{Number of Correct Predictions}}{\text{Total Number of Predictions}}
\]
Accuracy values range from 0 to 1 (or 0\% to 100\%), with higher values indicating better predictive performance.

\vspace{0.5em}
\textbf{Mean Absolute Error (MAE):} 
MAE evaluates the average magnitude of errors in a set of predictions, without considering their direction. In peer review evaluation, it is typically used for regression tasks, such as predicting numerical review scores (e.g., for originality or impact). The formula is:
\[
\mathrm{MAE} = \frac{1}{n} \sum_{i=1}^n \left| y_i - \hat{y}_i \right|
\]
where $y_i$ is the true score, $\hat{y}_i$ is the predicted score, and $n$ is the total number of items. Lower MAE values indicate more accurate predictions.

\subsection{Unsupervised and Reward-Based Metrics}
The need for scalable evaluation without a direct human reference for every generated review has driven the development of unsupervised and reward-based metrics.

\vspace{0.5em}
\textbf{Spearman’s Rank Correlation Coefficient ($\rho$):} 
Spearman’s $\rho$ measures the strength and direction of the monotonic relationship between two ranked variables. In peer review evaluation, it is used to compare the model-based ranking of reviews or systems with human-preference leaderboards. The formula is:
\[
\rho = 1 - \frac{6 \sum_{i=1}^{n} d_i^2}{n(n^2 - 1)}
\]
where $d_i$ is the difference between the ranks assigned by the two systems for item $i$, and $n$ is the number of items. $\rho$ ranges from $-1$ (perfect inverse correlation) to $1$ (perfect agreement), with $0$ indicating no correlation.

\vspace{0.5em}
\textbf{Kendall’s Rank Correlation Coefficient ($\tau$):} 
Kendall’s $\tau$ assesses the ordinal association between two ranked lists by counting the number of concordant and discordant pairs. Its formula is:
\[
\tau = \frac{(C - D)}{\frac{1}{2} n(n-1)}
\]
where $C$ is the number of concordant pairs, $D$ is the number of discordant pairs, and $n$ is the number of items. Values of $\tau$ range from $-1$ (complete disagreement) to $1$ (complete agreement).

\vspace{0.5em}
\textbf{Human-aligned Peer Review Reward (HPRR):} 
HPRR is a composite, multi-objective reward function designed for reinforcement learning in peer review generation. It quantifies review quality as a weighted sum of several sentence-level attributes, such as Criticism ($Cr$), Examples ($Ex$), Suggestions and Solutions ($SuSo$), and overall relevance (often measured by METEOR). The formula can be abstracted as:
\[
\mathrm{HPRR} = w_1 \cdot Cr + w_2 \cdot Ex + w_3 \cdot SuSo + w_4 \cdot \mathrm{M} + \ldots
\]
where $w_i$ are learned weights reflecting human preferences, M is METEOR. HPRR enables automated optimization toward human-aligned review usefulness and quality.

%% file: appendix/discussion_benchmark.tex
\section{Further Discussion of Benchmark Perspective}
\label{app:benchmarkdiscussion}
\subsection{Open Challenge of Evaluation}
\paragraph{Evaluating Deep Scientific Reasoning.}
Current methods, even the most advanced, are still limited in their ability to assess the deep, domain-specific scientific reasoning that is the hallmark of a true expert review. Identifying logical fallacies, assessing the validity of a complex experimental design, or judging the true intellectual contribution of a novel method remains profoundly difficult to automate. Probing for critical errors in papers with known flaws is a promising first step, but significant work is needed to develop evaluations that can reliably measure these higher-order reasoning capabilities.

\paragraph{Towards Holistic Evaluation Frameworks.}
The future of evaluation does not lie in finding a single "best" metric. Instead, it requires the development of integrated, holistic frameworks that combine the strengths of multiple paradigms. A robust evaluation protocol for a new review generation system might involve using an unsupervised method like PiCO~\cite{ning2025picopeerreviewllms} for rapid, large-scale comparison against other models, employing a suite of aspect-oriented metrics (e.g., Focus-Level, SubstanScore) for deep diagnostic analysis of its specific flaws, and conducting targeted, comparative human evaluation on a small but representative sample of its outputs for final validation and bias checking.

\subsection{Challenges for using the latest datasets (post-2023)}
\label{app:dataset-challenges}
This part of the appendix discusses the practical challenges associated with using post-2023 peer review datasets and proposes mitigation strategies. Our goal is to help researchers make informed choices and inspire improvements in future dataset construction and refinement.
\paragraph{Challenges.}
Despite significant advancements in dataset scale, diversity, and annotation richness, several limitations persist across recent datasets for peer review generation. Many corpora remain optimized for auxiliary tasks such as score prediction or classification, rather than full-text review generation, which results in weak alignment between individual review sentences and specific paper content. Dialogue-based datasets often suffer from inconsistency in speaker roles, incomplete conversational threads, or ambiguous author responses, making them difficult to use in multi-agent training setups. Annotation scarcity is another common obstacle: datasets offering fine-grained critique labels are often small, limiting their utility for supervised learning. Additionally, review-level scores and aspect annotations are rarely paired with sentence-level metadata, making it hard to evaluate or optimize local review quality. Domain coverage also remains narrow, with most datasets focusing exclusively on NLP and ML venues—raising concerns about generalizability.

\vspace{0.5em}
\paragraph{Mitigation.} To address these issues, future dataset efforts should emphasize structural coherence and alignment. Establishing mappings between review sentences and the specific sections they critique would enable content-aware generation and interpretation. Dialogue datasets can benefit from more rigorous role standardization and filtering pipelines that remove incomplete or incoherent threads. Where annotation budgets are limited, semi-supervised expansion techniques—such as label propagation, weak supervision, or LLM-based pseudo-annotation—could help scale fine-grained labels. Additionally, pairing review-level scores with sentence-level utility judgments would enable more precise evaluation and training of critique generation. Finally, future datasets should aim to extend beyond the NLP/ML domain and incorporate reviews from diverse disciplines to better support general-purpose review generation models and cross-domain evaluation.

%% file: appendix/discussion_methodologies.tex
\section{Further Discussion of Methodologies}
\subsection{Open Challenge}
\paragraph{Mitigating Bias in Automated Review.}
A critical and still under-explored challenge in automated peer review is the profound risk of these systems inheriting, amplifying the human biases present in their training data. At the same time, the very technologies for Controllable Text Generation (CTG) that could exacerbate this problem also offer a potential path toward building fairer and more objective reviewers. To effectively mitigate bias, we must first be able to measure it. This necessitates the parallel development of "bias benchmarks" designed specifically for the scientific domain.

\paragraph{Alignment Between Review Points and Final Decision.}
Recent studies expose a persistent disconnect between aspect scores and final decisions~\cite{Sukpanichnant2024}. While models can predict both accurately in isolation~\cite{muangkammuen2022,yuan2021automate}, bridging them remains difficult. Checklist-guided generation~\cite{zeng2023checklist} and decision-aware meta-reviewing~\cite{kumar2024pipeline} offer partial solutions by explicitly linking critique structure to recommendations. Future work should explore causal modeling of score-to-decision transitions, perhaps with richer datasets like \textsc{MOPRD}~\cite{lin2022moprd} or \textsc{ReviewEval}~\cite{garg2024revieweval}.

\subsection{Future Direction}
\paragraph{Reimagining Peer Review as a Process.}
Many generation models treat review writing as a one-shot task. Agent-based frameworks—such as \textsc{ReviewAgents}~\cite{reviewagents}, \textsc{AgentReview}~\cite{agentreview}, and \textsc{PeerReviewAs}~\cite{peerreviewas}, instead simulate multi-turn workflows involving multiple roles. These systems better capture the iterative nature of review, rebuttal, and decision, and create opportunities for testing social factors (e.g., bias, anchoring, consensus drift) under controlled conditions. Their development offers both new modeling challenges and a tool for studying the peer review process itself.

\paragraph{Reinforcement Learning and Agentic Frameworks for Deeper Reasoning.}
To overcome the tendency of simple fine-tuned models to produce shallow, generic, and overly positive reviews, the field is rapidly adopting more complex learning paradigms. Multi-objective reinforcement learning and multi-agent or multi-stage frameworks are emerging as powerful approaches. These methods decompose the monolithic task of "writing a review" into more manageable, specialized, and logically sequenced sub-problems, thereby fostering deeper and more structured reasoning.

\paragraph{Peer Review as Improver.}
Is peer review's primary role gatekeeping acting as a filter, or should it embrace the role of an improver who provides actionable feedback to elevate the quality of the manuscript? Our survey of the literature reveals that current systems predominantly lean towards the gatekeeping paradigm. However, we believe that the future of automated scientific discovery~\cite{novikov2025alphaevolvecodingagentscientific}, hinges on a decisive shift towards the improvement-oriented paradigm. We hypothesize that improvement-oriented training in which the generated review directly informs the revision and is rewarded when it yields measurable improvements to the manuscript will better support automated discovery workflows. Recent frameworks link review signals to optimization loops such as CycleResearcher’s RLHF loop~\cite{cycleresearcher} for research-review-refinement, suggesting a pathway from actionable review to iterative scientific progress.

\paragraph{CfP-aware Venue Fit and Policy Checking.} Call for Papers (CfP) documents already encode conference scope, topical focus, and LLM-usage or desk-reject policies for venues such as NeurIPS and ARR. Existing LLM-based peer-review systems like SEA~\cite{yu2024automatedpeerreviewingpaper}, OpenReviewer~\cite{openreview} and ReviewEval~\cite{garg2024revieweval} align prompts or metrics with reviewer guidelines, but they do not treat CfPs themselves as retrieval sources. Future work could integrate CfP passages into RAG pipelines as machine-readable constraints to support venue-fit assessment, automatic flagging of policy violations, and assisted desk-reject decisions.

\paragraph{Remaining Non‑Generative Components.}
Predictive tasks such as aspect‑score regression~\cite{muangkammuen2022}, final‑decision forecasting~\cite{Sukpanichnant2024}, and desk‑triage systems, i.e., automatic screening or “reject‑option” classifiers~\cite{hendrickx2021reject}, remain critical, but their tooling is now intertwined with generation (e.g., conditioning meta‑review generation on predicted scores).

\paragraph{Interactive Human-in-the-Loop Workflows.}
Rather than aiming for full automation, future systems will likely focus on creating more seamless and powerful human-AI collaboration tools. This could involve real-time, interactive interfaces where a human reviewer can guide, correct, and query the AI assistant, leveraging its ability to rapidly retrieve literature, analyze data, and draft text while retaining ultimate human oversight and judgment.

\subsection{Multimodal Input}
\label{app:multimodalinput}
A comprehensive and high-fidelity automated peer review system must possess the capacity to understand and reason about the entirety of a scientific manuscript, including its non-textual components~\cite{kumar-etal-2024-longform}. The central challenge in this domain extends beyond the technical task of processing pixels or parsing table cells; it lies in bridging a profound semantic gap. Therefore, the most formidable aspect of multimodal integration is not the technical encoding of visual data, but rather teaching a model to perform this complex interpretive leap from perception to conceptual understanding. This suggests that generic vision-language pre-training, while useful, is likely insufficient. Progress will depend on fine-tuning models on specialized tasks that explicitly reward this form of semantic interpretation~\cite{wang2025sciverevaluatingfoundationmodels}.

A leading paradigm in this area is the Hybrid Modality Approach, exemplified by the Multi-Modal Automated Academic Papers Interpretation System (MMAPIS)~\cite{jiang2024bridgingresearchreadersmultimodal}. For text and mathematical formula extraction, it leverages a transformer-based model like Nougat, which is adept at converting PDF pages into a structured format like Markdown. This process preserves not only the plaintext but also the intricate mathematical notations that are indispensable for understanding technical content. Concurrently, for visual elements, the system employs a separate tool, PDFFigures 2.0. This tool reasons about the document's layout, such as the empty regions and whitespace separating blocks of text, to identify, extract, and caption figures and tables~\cite{im-etal-2021-self}.

\subsection{Comparison Between Methodologies}
\label{app:compare}
\paragraph{Agent-Based Methods vs. Reinforcement Learning.}
Agent-based modeling and Reinforcement Learning represent two distinct philosophies for generating peer reviews. Agent-based approaches, such as AgentReview \cite{agentreview}, excel at simulating the complex, interactive social dynamics of the entire peer review ecosystem. However, the primary goal of these systems is often to study emergent behaviors rather than to optimize the quality of any single generated review. This focus on process simulation. In contrast, Reinforcement Learning directly targets the optimization of the review artifact. Frameworks like REMOR~\cite{remor} frame review generation as a policy that is optimized to maximize a multi-objective reward function, which is carefully designed to encapsulate human preferences for what constitutes a helpful review. This allows RL-based methods to effectively steer generation away from the shallow, overly positive feedback that vanilla LLMs often produce.

\paragraph{Iterative Refinement vs. Single-Pass Generation.}
Iterative refinement is a powerful technique that mimics the human cognitive process of drafting, critiquing, and revising~\cite{xu2025chaindraftthinkingfaster,kamoi-etal-2024-llms}. This approach is particularly effective for complex, multi-faceted tasks like peer review, as it allows the model to address objectives that are often missed in a single pass and can unlock the full potential of LLMS. The primary drawback is a significant increase in latency and computational cost, as each feedback-and-refine cycle constitutes an additional, expensive call.

Single-pass generation, the standard autoregressive approach, is the most computationally efficient method. However, it is inherently limited by its inability to perform global planning or correct early mistakes.

\paragraph{Retrieval-Augmented vs. Self-Contained Generation.}
Retrieval-Augmented Generation (RAG)~\cite{lewis2021retrievalaugmentedgenerationknowledgeintensivenlp} is critical for grounding peer reviews in verifiable, external knowledge, a necessity for tasks like assessing a paper's novelty or checking for factual inaccuracies. However, RAG introduces significant architectural complexity and a new critical point of failure: the retriever. The quality of the final generation is fundamentally bottlenecked by the quality of the retrieved context; irrelevant or low-quality retrieval leads to poor, unhelpful reviews.

Self-contained generation, which relies solely on the LLM's parametric knowledge, is architecturally simpler and faster at inference. Its primary limitation is that its knowledge is static, potentially outdated, and prone to factual invention (hallucination), making it fundamentally unsuited for rigorously evaluating a submission's contribution to a rapidly evolving field. Therefore, while self-contained generation may suffice for assessing aspects like clarity or presentation, any deep evaluation of a paper's technical substance and novelty necessitates grounding in external knowledge.

%% file: appendix/methodologies_summary.tex
\section{Methodologies Summary}
\label{app:methodologies_summary}
We summarizes the main methodologies for peer review generation in~\autoref{tab:methodologies_summary}. We also organizes different LLM backbones used in methods in~\autoref{tab:backbone_comparison}.

\input{tables/LLM_backbones}

%% file: tables/LLM_backbones.tex
\begin{table*}[!ht]
\centering
\resizebox{\textwidth}{!}{%
\begin{tabular}{@{}lllp{5.5cm}p{1.5cm}@{}}
\toprule
\textbf{System} & \textbf{Backbone LLM} & \textbf{Modification Type} & \textbf{Purpose / Role in the Framework} & \textbf{Open-source?} \\
\midrule

\textbf{CycleResearcher} — \citet{cycleresearcher} & 
\begin{tabular}[c]{@{}l@{}}Mistral-Nemo-12B~\citep{mistral7b}\\ Qwen2.5-Instruct-72B~\citep{qwen2.5}\\ Mistral-Large-2 (123B)~\citep{mistrallarge}\end{tabular} & 
\begin{tabular}[c]{@{}l@{}}SFT + RL (SimPO)\\ + Reward model via LoRA\end{tabular} & 
\begin{tabular}[c]{@{}l@{}}CycleResearcher generates papers \\ using RLHF.\\ CycleReviewer serves as the \\ reward model.\end{tabular} & 
Yes \\

\midrule

\textbf{DeepReview} — \citet{deepreview} & 
GPT-4 (API)~\citep{openai2024gpt4technicalreport} & 
Zero-shot (no finetuning) & 
\begin{tabular}[c]{@{}l@{}}Thinker agent produces reasoning \\traces.\\ Writer agent generates the\\ structured review.\end{tabular} & 
No \\

\midrule

\textbf{OpenReviewer} — \citet{openreview} & 
Llama-3.1-8B-Instruct~\citep{llama3modelcard} & 
Full finetuning (Axolotl) & 
Markdown-based review generation with template-specific prompts and PDF-to-text preprocessing. & 
Yes \\

\midrule

\textbf{REMOR} — \citet{remor} & 
DeepSeek-R1-Distill-Qwen-7B~\citep{deepseekr1} & 
LoRA + RL (GRPO) & 
\begin{tabular}[c]{@{}l@{}}Backbone for both supervised and \\ RL stages.\\ Trained into REMOR-U \\ and REMOR-H variants.\end{tabular} & 
Yes \\

\bottomrule
\end{tabular}%
}
\caption{Comparison of LLM Backbone Strategies in Recent Peer Review Generation Systems. We summarize the core base models used, how they are customized, their functional roles, and whether they are publicly available.}
\label{tab:backbone_comparison}
\end{table*}

%% file: appendix/data_collection.tex
\section{Data Collection}
\label{app:data_collection}
We summarize in~\autoref{tab: Information gathering} the typical ways to gather papers together with their reviews. These range from directly using the OpenReview API to other APIs. In practice, researchers often combine multiple sources to ensure both coverage and diversity across venues.

Meanwhile, once data is collected, heterogeneous formats (PDFs, HTML pages, JSON exports, or XML metadata) need to be standardized. \autoref{tab:conversion_tools} lists common tools and pipelines for data type conversion.

Together, these resources provide a practical toolkit for curating review corpora that are both machine-readable and comparable across venues.


\input{tables/information_gathering_methods}
\input{tables/Data_conversion_method}

%% file: tables/information_gathering_methods.tex
\begin{table*}[ht]
\centering
\footnotesize
\setlength{\tabcolsep}{5pt}
\label{tab:collection_methods}
\begin{tabularx}{\textwidth}{l
  >{\raggedright\arraybackslash}p{0.28\textwidth}
  l
  >{\raggedright\arraybackslash}p{0.30\textwidth}}
\toprule
\textbf{Channel / Platform} & \textbf{Representative Studies} & \textbf{Disciplines} & \textbf{Collection Methods \& APIs (abbr.)} \\
\midrule
OpenReview &
\citet{peerread,nlpee,cycleresearcher} &
ML / NLP &
\textbullet\;OR-API \\ 
&&&\textbullet\;GQL \\ 
&&&\textbullet\;HTML-WC \\[4pt]

Softconf / START &
\citet{peerread,nlpee} &
NLP &
\textbullet\;OPT-IN \\ 
&&&\textbullet\;SQL-DMP \\[4pt]

F1000Research &
\citet{nlpee} &
Life Sci &
\textbullet\;CSV-EXP \\ 
&&&\textbullet\;HTML-SCR \\ 
&&&\textbullet\;DOI-TCK \\[4pt]

PeerJ  &
\citet{lin2022moprd} &
Bio / Chem / CS &
\textbullet\;REST-CRL \\ 
&&&\textbullet\;HTML-CRL \\[4pt]

Nature Comms PR File &
\citet{peerreviewas} &
Multidomain &
\textbullet\;HTTP-DL \\
\bottomrule
\end{tabularx}

\vspace{4pt}
\footnotesize
\caption{Channels and APIs for Harvesting Paper–Review Corpora. Abbreviation key:\;
OR-API = \textit{openreview-py} API,\;
GQL = GraphQL query,\;
HTML-WC = web crawler,\;
OPT-IN = author/reviewer opt-in dump,\;
SQL-DMP = SQL/CSV dump,\;
CSV-EXP = official CSV export,\;
HTML-SCR = HTML scraping,\;
DOI-TCK = DOI tracking,\;
REST-CRL = REST crawler,\;
HTTP-DL = bulk HTTP download.}
\label{tab: Information gathering}
\end{table*}

%% file: tables/Data_conversion_method.tex
\begin{table*}[ht]
\centering
\setlength{\tabcolsep}{4pt}          
\renewcommand{\arraystretch}{1.03}   
\footnotesize
\begin{tabularx}{\textwidth}{l l l X}
\toprule
\textbf{Tool / Stage} & \textbf{Used in Studies} & \textbf{Input $\;\rightarrow\;$ Output} & \textbf{Main Purpose \& Highlights} \\
\midrule
GROBID \cite{GROBID} & \citet{lin2022moprd} & PDF $\;\rightarrow\;$ XML & Accurate scholarly parse; keeps sections/cites; batch-ready. \\

Science Parse \cite{ScienceParse} & \citet{peerread} & PDF $\;\rightarrow\;$ JSON & Fast metadata + rough text; web-scale. \\

Marker \cite{Marker} & \citet{peerreviewas} & PDF $\;\rightarrow\;$ MD & Long-context MD; preserves headings and formulas. \\

MagicDoc \cite{2024magic-doc} & \citet{cycleresearcher} & PDF $\;\rightarrow\;$ MD & Bulk Softconf PDFs to MD; code-block safe. \\

MinerU \cite{mineru} & \citet{deepreview} & PDF $\;\rightarrow\;$ MD / JSON & High-precision; layout/reading-order faithful. \\

Semantic Scholar \cite{Kinney2023TheSS} & \citet{nlpee} & DOI $\;\rightarrow\;$ BibJSON & Reference enrichment; citation, year, impact. \\
\bottomrule
\end{tabularx}
\caption{Tools Frequently Used to Convert / Enrich Paper–Review Data}
\label{tab:conversion_tools}
\end{table*}

%% file: appendix/evaluation_summary.tex
\section{Evaluation Summary}
\label{app:evaluation_summary}
We summarizes the evaluation methods for peer review generation in~\autoref{tab:evaluation_taxonomy}. We also compares their pros\&cons in~\autoref{tab:evaluation_proscons}. The datasets we organizes are shown in~\autoref{tab:datasets}.
\input{tables/evaluation_summary}

\input{tables/dataset}

%% file: tables/evaluation_summary.tex
\begin{table*}[!ht]
\centering
\resizebox{\textwidth}{!}{%
\begin{tabular}{@{}lllp{5cm}p{5cm}@{}}
\toprule
\textbf{Category} & \textbf{Paper} & \textbf{Core Method} & \textbf{Metrics / Framework Details} & \textbf{Key Finding / Contribution} \\
\midrule

\multirow{8}{*}{\begin{tabular}[c]{@{}l@{}}\textbf{Human-Centric}\\ \textit{(Direct Rating)}\end{tabular}}
& \citet{gpt4slightly} & Pilot Study & 10 users rated helpfulness (1-5 scale). & GPT-4 reviews had same mean helpfulness as humans (3/5) but higher variance. \\
\cmidrule(l){2-5}
& \citet{liang2023largelanguagemodelsprovide} & Large-Scale User Study & 308 researchers rated LLM feedback on their own papers. & High perceived value: 57.4\% found it helpful; 82.4\% found it more beneficial than some human reviews. \\
\cmidrule(l){2-5}
& \citet{Drori2024Human} & Structured Evaluation & GPT-4 reviews structured against standard review criteria. & AI assistance can reduce reviewer workload but is not completely reliable for all cases. \\
\cmidrule(l){2-5}
& \begin{tabular}[c]{@{}l@{}}\citet{Hosseini2023Fighting},\\ \citet{schintler2023criticalexaminationethicsaimediated},\\ \citet{Lee2025Role}\end{tabular} & Qualitative Critique & Review of benefits, risks, and ethical concerns. & Highlighted efficiency gains but warned of bias, compromised integrity, and weakness in assessing scientific validity. \\
\midrule
\multirow{1}{*}{\begin{tabular}[c]{@{}l@{}}\textbf{Human-Centric}\\ \textit{(Pairwise Comparison)}\end{tabular}}
& \citet{tyser2024aidrivenreviewsystemsevaluating} & Arena-Based Evaluation & \textbf{Reviewer Arena}: Pairwise human (and LLM) preference comparisons. & Proposes a scalable method for quality assessment based on relative judgments rather than absolute scores. \\
\midrule

\multirow{4}{*}{\begin{tabular}[c]{@{}l@{}}\textbf{Reference-Based}\\ \textit{(Overlap Metrics)}\end{tabular}}
& \citet{liang2023largelanguagemodelsprovide} & Semantic Matching & Two-stage process: GPT-4 extracts key comments, then semantic matching calculates a "hit rate". & Offers a nuanced content overlap metric beyond simple lexical similarity. \\
\cmidrule(l){2-5}
& \citet{zhou-etal-2024-llm} & Standard Metrics & ROUGE, BERTScore against concatenated human reference reviews. & Established a baseline methodology for using standard NLP metrics in this domain. \\
\cmidrule(l){2-5}
& \citet{yu2024automatedpeerreviewingpaper} & Standard Metrics (SEA) & BLEU, ROUGE, BERTScore against concatenated human references. & Noted that ROUGE-Recall is particularly indicative of review comprehensiveness. \\
\midrule
\multirow{2}{*}{\begin{tabular}[c]{@{}l@{}}\textbf{Reference-Based}\\ \textit{(Task-Based)}\end{tabular}}
& \citet{liu2023reviewergptexploratorystudyusing} & Targeted Tasks & Seeded error detection, checklist verification, abstract comparison. & LLMs show mixed results: successful at some tasks (checklist) but poor at others (error detection, comparative judgment). \\
\cmidrule(l){2-5}
& \citet{zhou-etal-2024-llm} & MCQ Benchmark & \textbf{RR-MCQ}: A multiple-choice question dataset from ICLR review-rebuttal exchanges. & LLMs achieve passable accuracy ($>$60\%) but struggle with long papers and providing deep critical feedback. \\
\midrule

\multirow{1}{*}{\textbf{LLM-Based}}
& \citet{zhang2025reviewingscientificpaperscritical} & Multi-Judge Pipeline & Requires an affirmative vote from two distinct LLM judges to confirm a "hit" (finding a critical error). & Aims to reduce single-model bias and increase the reliability of automated evaluation. \\
\cmidrule(l){2-5}
& \citet{ning2025picopeerreviewllms} & Peer-Evaluation & \textbf{PiCO Framework}: LLMs both generate reviews and provide pairwise preferences on peers' outputs. Ranks models via consistency optimization. & Pioneers fully unsupervised evaluation without human labels or gold references. \\
\midrule

\multirow{4}{*}{\textbf{Aspect-Oriented}}
& \citet{shin2025mindblindspotsfocuslevel} & Focus-Level Framework & Annotates review content by targets (e.g., Method) and aspects (e.g., Novelty) to compare LLM vs. human focus. & Revealed LLM "blind spots": over-emphasis on technical validity, under-emphasis on novelty. \\
\cmidrule(l){2-5}
& \citet{du2024critique} & Fine-Grained Annotation & \textbf{ReviewCritique}: Dataset with 23 fine-grained "deficiency" types annotated at the sentence level. & Enables testing of whether models can identify specific types of errors found by humans. \\
\cmidrule(l){2-5}
& \citet{ye2024yetrevealingrisksutilizing} & Adversarial Testing & "Review injection attacks" and testing susceptibility to authors' proactive disclosure of minor limitations. & Demonstrates LLMs' vulnerability to both explicit and implicit manipulation. \\
\cmidrule(l){2-5}
& \citet{yu2024automatedpeerreviewingpaper} & Inconsistency Metric & \textbf{Mismatch Score}: Predicts the degree of inconsistency between a paper and a given review. & Introduces a novel aspect-specific metric focused on review-paper faithfulness. \\
\cmidrule(l){2-5}
& \citet{sadallah2025goodbadconstructiveautomatically} & Comment-Level Utility & \textbf{RevUtil}: Four author-oriented dimensions for weakness comments. & Proposes an aspect-based utility framework and finds LLM reviews generally underperform human reviews. \\


\bottomrule
\end{tabular}%
}
\caption{Taxonomy of Evaluation Methodologies for Automated Peer Review Generation. This table summarizes the primary approaches, key papers, specific metrics or frameworks, and notable findings discussed in the literature.}
\label{tab:evaluation_taxonomy}
\end{table*}

%% file: tables/dataset.tex
\begin{table*}[t]
\centering
\scriptsize
\setlength{\tabcolsep}{2.2pt}
\renewcommand{\arraystretch}{0.92}
\begin{tabularx}{\textwidth}{l l p{2.0cm} p{2.75cm} p{1.9cm} p{3.4cm} p{2.60cm}}
\toprule
\textbf{Era} & \textbf{Phase} & \textbf{Name} & \textbf{Main Use} & \textbf{Scale} & \textbf{Source} & \textbf{Expert Annotation} \\
\midrule

\multirow{9}{*}{\rotatebox[origin=c]{90}{Pre-2023}}
& Reviewing
& \textsc{PeerRead} \citep{peerread}
& Review-score prediction, acceptance prediction, review generation
& 14.7k papers; 10.7k reviews
& ICLR 2017--2019 OpenReview; ACL 2017 author-consented drafts/reviews; NeurIPS 2013--2017 drafts matched to accepted arXiv papers
& -- \\
\addlinespace[0.35ex]

& 
& \textsc{DISAPERE} \citep{kennard2021disapere}
& Discourse analysis of reviewer--author discussion
& 506 threads
& ICLR 2019--2020 OpenReview discussion threads
& Sentence-level discourse relations in review--rebuttal discussions \\
\addlinespace[0.35ex]

&
& \textsc{PEERAssist} \citep{bharti2021peerassist}
& Acceptance prediction, review-score prediction
& 4,467 papers; 13.4k reviews
& ICLR 2017--2020 OpenReview submissions and reviews
& -- \\
\addlinespace[0.35ex]

&
& \textsc{ASAP-Review} \citep{yuan2022asapreview}
& Aspect-aware review generation, aspect tagging
& 8,877 papers; 28.1k reviews
& ICLR 2017--2020 and NeurIPS 2016--2019 OpenReview data
& 1k-review subset annotated with 15 aspect labels \\
\addlinespace[0.35ex]

&
& \textsc{ReAct} \citep{choudhary2022react}
& Actionability classification, comment-type tagging
& 1.25k labeled; 52k unlabeled comments
& ICLR 2018 OpenReview comments
& Crowdsourced actionability and 7 comment types \\
\addlinespace[0.35ex]

& Meta-review
& \textsc{MReD} \citep{shen2021mred}
& Structure-controlled meta-review generation
& 7,089 meta-reviews; 45k sentences
& ICLR 2018--2021 OpenReview meta-reviews
& Every sentence labeled with 9 discourse tags; adjudicated \\
\addlinespace[0.35ex]

&
& \textsc{PRRCA} \citep{wu2022prrca}
& Meta-review generation from reviews and rebuttals
& 6,138 submissions
& ICLR 2017--2022 OpenReview threads with reviews, rebuttals, decisions, and meta-reviews
& -- \\
\addlinespace[0.35ex]

& Multi-stage
& \textsc{ICLR-DB} \citep{zhang2022fairness}
& Fairness analysis, decision/review prediction, review generation
& 10.3k submissions; 36.5k reviews; 68.7k responses
& ICLR 2017--2022 OpenReview threads, enriched with author profiles and LLM-derived features
& -- \\
\addlinespace[0.35ex]

&
& \textsc{MOPRD} \citep{lin2022moprd}
& Meta-review generation, decision prediction, rebuttal generation
& 6,578 papers
& PeerJ peer-review threads with reviews, rebuttals, meta-reviews, and decisions
& Manually aligned review-process records \\
\midrule

\multirow{13}{*}{\rotatebox[origin=c]{90}{Since 2023}}
& Meta-review
& \textsc{PeerSum} \citep{li-etal-2023-summarizing}
& Meta-review generation from discussions
& 14,993 triples
& ICLR 2020--2022 OpenReview threads with reviews, discussions, and meta-reviews
& -- \\
\addlinespace[0.35ex]

& \multirow[t]{7}{*}{Reviewing}
& \textsc{NLpeer} \citep{nlpee}
& Score prediction, pragmatic labeling, guided skimming
& 5,672 papers; 11k+ reviews
& ICLR and NeurIPS 2017--2022 OpenReview submissions and reviews
& Sentence-level pragmatic tags on an F1000 subset \\
\addlinespace[0.35ex]

&
& \textsc{SubstanReview} \citep{guo2023automatic}
& Claim substantiation detection and scoring
& 550 reviews
& ARR 2021--2022 reviews
& Expert span-level claim--evidence annotation \\
\addlinespace[0.35ex]

&
& \textsc{ReviewCritique} \citep{du2024critique}
& Deficiency detection in human and LLM reviews
& 100 papers
& 100 NLP papers with human and LLM reviews
& Experts annotate 23 fine-grained deficiency types \\
\addlinespace[0.35ex]

&
& \textsc{ReviewEval} \citep{garg2024revieweval}
& Evaluation of human and LLM reviews
& 120 papers
& 120 NeurIPS, ICLR, and UAI submissions with gold and candidate reviews
& -- \\
\addlinespace[0.35ex]

&
& \textsc{Review-5k} \citep{cycleresearcher}
& Aspect-level score prediction, review evaluation
& 4,189 train + 782 test; 16k+ comments
& ICLR 2023 OpenReview reviews and comments
& -- \\
\addlinespace[0.35ex]

&
& \textsc{Reviewer2} \citep{gao2024reviewer2}
& Aspect-prompted review generation, diversity benchmarking
& 27k papers; 99k reviews
& Reviews from 6 major ML/NLP venues, 2017--2022
& LLM-generated aspect prompts \\
\addlinespace[0.35ex]

&
& \textsc{DeepReview-13K} \citep{zhu2025deepreview}
& Deep-thinking review generation
& 13k structured reviews
& OpenReview papers and reviews, 2017--2023
& Reasoning-step annotations for supervision \\
\addlinespace[0.35ex]

& Revision
& \textsc{ARIES} \citep{darcy2023aries}
& Comment--edit alignment, revision generation
& 3.9k comment--edit pairs; 196 test cases
& ICLR, NeurIPS, and ACL OpenReview papers, 2018--2022
& Expert-annotated gold test set \\
\addlinespace[0.35ex]

& Rebuttal-included
& \textsc{AgentReview} \citep{jin2024agentreview}
& Peer-review simulation, bias/dynamics analysis, synthetic review generation
& $\sim$500 submissions; 10k reviews; 53.8k artifacts
& LLM-agent simulations over ICLR 2018--2021 papers
& -- \\
\addlinespace[0.35ex]

&
& \textsc{RR-MCQ} \citep{zhou2024rrmcq}
& Review--revision QA, reviewer reliability benchmark
& 197 MCQs
& ICLR 2023 review--rebuttal threads
& Manually written and labeled MCQs \\
\addlinespace[0.35ex]

&
& \textsc{ReviewMT} \citep{peerreviewas}
& Dialogue-style review simulation
& 26k+ papers; 92k+ reviews
& OpenReview reviews reorganized as multi-turn dialogues
& Speaker roles labeled \\
\addlinespace[0.35ex]

&
& \textsc{Re$^{2}$} \citep{zhang2025re2}
& Full-stage review and rebuttal modeling, decision prediction, dialogue generation
& 19.9k submissions; 70.7k reviews; 53.8k rebuttals
& OpenReview data from 24 conferences and 21 workshops, 2017--2025
& Consistency-filtered multi-turn rebuttal threads \\

&
& \textsc{RMR-75K} \citep{wu2026rbtactrebuttalsupervisionactionable}
& Review--rebuttal alignment, actionable feedback generation
& 75.5k mappings; 4,825 papers
& ICLR 2024, reviews, and rebuttal threads
& Segment-level review--rebuttal mapping with perspective and impact labels \\

\bottomrule
\end{tabularx}
\caption{Public peer-review datasets, grouped by era and phase.}
\label{tab:datasets}
\end{table*}

%% file: appendix/review_procedure.tex
\section{Literature Review Procedure}
\label{app:reviewprocedure}
To ensure transparency and rigor, we document here the procedure by which our literature review was conducted. 

\paragraph{Databases.} We searched major sources including ACL Anthology, OpenReview (ICLR, NeurIPS), arXiv, Semantic Scholar, DBLP, and Google Scholar. 

\paragraph{Search Strategy.} We applied keyword combinations such as ``peer review generation,'' ``meta-review,'' and ``rebuttal generation'' within the time range 2018--2025. In addition, we adopted a snowballing strategy by tracing the references and citations of seminal works (e.g., PeerRead~\cite{peerread}) and recent contributions (e.g., DeepReview~\cite{deepreview}). 

\paragraph{Inclusion Criteria (IC).} 
\begin{itemize}[leftmargin=*]
  \item IC1: The paper must directly address AI-based peer review generation, its evaluation, or closely related subtasks (e.g., meta-review, rebuttal generation).  
  \item IC2: The paper must be publicly available as a journal article, conference paper, or preprint.  
  \item IC3: The paper must be written in English.  
  \item IC4: Focus primarily on work published after 2018, especially post-LLM developments.  
\end{itemize}

\paragraph{Exclusion Criteria (EC).} 
\begin{itemize}[leftmargin=*]
  \item EC1: Studies that only predict or analyze peer review scores without generating review text (e.g.,~\cite{thelwall2024evaluatingpredictivecapacitychatgpt, Laureano2025Predicting}).  
  \item EC2: Abstracts, short articles, and non-academic blog posts.  
  \item EC3: Papers without accessible full text.  
\end{itemize}

\paragraph{Screening and Statistics.} 
Our initial screening retrieved approximately 500 articles. After deduplication, around 400 articles remained. Applying inclusion/exclusion criteria yielded 194 papers, of which 138 directly focus on peer review generation and its evaluation. 

\paragraph{Methodological Rigor.} 
Our protocol is informed by established guidelines for systematic reviews, such as the Kitchenham and Charters~\cite{kitchenham2007guidelines} SLR framework and the PRISMA 2020 statement~\cite{Pagen71}. These emphasize transparent reporting of search strings, last search date, de-duplication, per-stage counts, and inclusion/exclusion flows. By following these standards, we ensure that our literature review process is rigorous, reproducible, and aligned with recognized best practices.

%% file: custom.bib
@misc{chen2025ai4researchsurveyartificialintelligence,
      title={AI4Research: A Survey of Artificial Intelligence for Scientific Research}, 
      author={Qiguang Chen and Mingda Yang and Libo Qin and Jinhao Liu and Zheng Yan and Jiannan Guan and Dengyun Peng and Yiyan Ji and Hanjing Li and Mengkang Hu and Yimeng Zhang and Yihao Liang and Yuhang Zhou and Jiaqi Wang and Zhi Chen and Wanxiang Che},
      year={2025},
      eprint={2507.01903},
      archivePrefix={arXiv},
      primaryClass={cs.CL},
      url={https://arxiv.org/abs/2507.01903}, 
}

@misc{openai2024gpt4technicalreport,
      title={GPT-4 Technical Report}, 
      author={OpenAI and Josh Achiam and Steven Adler and Sandhini Agarwal and Lama Ahmad and Ilge Akkaya and Florencia Leoni Aleman and Diogo Almeida and Janko Altenschmidt and Sam Altman and Shyamal Anadkat and Red Avila and Igor Babuschkin and Suchir Balaji and Valerie Balcom and Paul Baltescu and Haiming Bao and Mohammad Bavarian and Jeff Belgum and Irwan Bello and Jake Berdine and Gabriel Bernadett-Shapiro and Christopher Berner and Lenny Bogdonoff and Oleg Boiko and Madelaine Boyd and Anna-Luisa Brakman and Greg Brockman and Tim Brooks and Miles Brundage and Kevin Button and Trevor Cai and Rosie Campbell and Andrew Cann and Brittany Carey and Chelsea Carlson and Rory Carmichael and Brooke Chan and Che Chang and Fotis Chantzis and Derek Chen and Sully Chen and Ruby Chen and Jason Chen and Mark Chen and Ben Chess and Chester Cho and Casey Chu and Hyung Won Chung and Dave Cummings and Jeremiah Currier and Yunxing Dai and Cory Decareaux and Thomas Degry and Noah Deutsch and Damien Deville and Arka Dhar and David Dohan and Steve Dowling and Sheila Dunning and Adrien Ecoffet and Atty Eleti and Tyna Eloundou and David Farhi and Liam Fedus and Niko Felix and Simón Posada Fishman and Juston Forte and Isabella Fulford and Leo Gao and Elie Georges and Christian Gibson and Vik Goel and Tarun Gogineni and Gabriel Goh and Rapha Gontijo-Lopes and Jonathan Gordon and Morgan Grafstein and Scott Gray and Ryan Greene and Joshua Gross and Shixiang Shane Gu and Yufei Guo and Chris Hallacy and Jesse Han and Jeff Harris and Yuchen He and Mike Heaton and Johannes Heidecke and Chris Hesse and Alan Hickey and Wade Hickey and Peter Hoeschele and Brandon Houghton and Kenny Hsu and Shengli Hu and Xin Hu and Joost Huizinga and Shantanu Jain and Shawn Jain and Joanne Jang and Angela Jiang and Roger Jiang and Haozhun Jin and Denny Jin and Shino Jomoto and Billie Jonn and Heewoo Jun and Tomer Kaftan and Łukasz Kaiser and Ali Kamali and Ingmar Kanitscheider and Nitish Shirish Keskar and Tabarak Khan and Logan Kilpatrick and Jong Wook Kim and Christina Kim and Yongjik Kim and Jan Hendrik Kirchner and Jamie Kiros and Matt Knight and Daniel Kokotajlo and Łukasz Kondraciuk and Andrew Kondrich and Aris Konstantinidis and Kyle Kosic and Gretchen Krueger and Vishal Kuo and Michael Lampe and Ikai Lan and Teddy Lee and Jan Leike and Jade Leung and Daniel Levy and Chak Ming Li and Rachel Lim and Molly Lin and Stephanie Lin and Mateusz Litwin and Theresa Lopez and Ryan Lowe and Patricia Lue and Anna Makanju and Kim Malfacini and Sam Manning and Todor Markov and Yaniv Markovski and Bianca Martin and Katie Mayer and Andrew Mayne and Bob McGrew and Scott Mayer McKinney and Christine McLeavey and Paul McMillan and Jake McNeil and David Medina and Aalok Mehta and Jacob Menick and Luke Metz and Andrey Mishchenko and Pamela Mishkin and Vinnie Monaco and Evan Morikawa and Daniel Mossing and Tong Mu and Mira Murati and Oleg Murk and David Mély and Ashvin Nair and Reiichiro Nakano and Rajeev Nayak and Arvind Neelakantan and Richard Ngo and Hyeonwoo Noh and Long Ouyang and Cullen O'Keefe and Jakub Pachocki and Alex Paino and Joe Palermo and Ashley Pantuliano and Giambattista Parascandolo and Joel Parish and Emy Parparita and Alex Passos and Mikhail Pavlov and Andrew Peng and Adam Perelman and Filipe de Avila Belbute Peres and Michael Petrov and Henrique Ponde de Oliveira Pinto and Michael and Pokorny and Michelle Pokrass and Vitchyr H. Pong and Tolly Powell and Alethea Power and Boris Power and Elizabeth Proehl and Raul Puri and Alec Radford and Jack Rae and Aditya Ramesh and Cameron Raymond and Francis Real and Kendra Rimbach and Carl Ross and Bob Rotsted and Henri Roussez and Nick Ryder and Mario Saltarelli and Ted Sanders and Shibani Santurkar and Girish Sastry and Heather Schmidt and David Schnurr and John Schulman and Daniel Selsam and Kyla Sheppard and Toki Sherbakov and Jessica Shieh and Sarah Shoker and Pranav Shyam and Szymon Sidor and Eric Sigler and Maddie Simens and Jordan Sitkin and Katarina Slama and Ian Sohl and Benjamin Sokolowsky and Yang Song and Natalie Staudacher and Felipe Petroski Such and Natalie Summers and Ilya Sutskever and Jie Tang and Nikolas Tezak and Madeleine B. Thompson and Phil Tillet and Amin Tootoonchian and Elizabeth Tseng and Preston Tuggle and Nick Turley and Jerry Tworek and Juan Felipe Cerón Uribe and Andrea Vallone and Arun Vijayvergiya and Chelsea Voss and Carroll Wainwright and Justin Jay Wang and Alvin Wang and Ben Wang and Jonathan Ward and Jason Wei and CJ Weinmann and Akila Welihinda and Peter Welinder and Jiayi Weng and Lilian Weng and Matt Wiethoff and Dave Willner and Clemens Winter and Samuel Wolrich and Hannah Wong and Lauren Workman and Sherwin Wu and Jeff Wu and Michael Wu and Kai Xiao and Tao Xu and Sarah Yoo and Kevin Yu and Qiming Yuan and Wojciech Zaremba and Rowan Zellers and Chong Zhang and Marvin Zhang and Shengjia Zhao and Tianhao Zheng and Juntang Zhuang and William Zhuk and Barret Zoph},
      year={2024},
      eprint={2303.08774},
      archivePrefix={arXiv},
      primaryClass={cs.CL},
      url={https://arxiv.org/abs/2303.08774}, 
}

@article{Checco2021AI,
  author    = {Checco, A. and Bracciale, L. and Loreti, P. and others},
  title     = {AI-assisted peer review},
  journal   = {Humanities and Social Sciences Communications},
  year      = {2021},
  volume    = {8},
  pages     = {25},
  doi       = {10.1057/s41599-020-00703-8},
  url       = {https://doi.org/10.1057/s41599-020-00703-8}
}

@misc{durante2024agentaisurveyinghorizons,
      title={Agent AI: Surveying the Horizons of Multimodal Interaction}, 
      author={Zane Durante and Qiuyuan Huang and Naoki Wake and Ran Gong and Jae Sung Park and Bidipta Sarkar and Rohan Taori and Yusuke Noda and Demetri Terzopoulos and Yejin Choi and Katsushi Ikeuchi and Hoi Vo and Li Fei-Fei and Jianfeng Gao},
      year={2024},
      eprint={2401.03568},
      archivePrefix={arXiv},
      primaryClass={cs.AI},
      url={https://arxiv.org/abs/2401.03568}, 
}

@article{Zhuang_2025,
   title={Large language models for automated scholarly paper review: A survey},
   volume={124},
   ISSN={1566-2535},
   url={http://dx.doi.org/10.1016/j.inffus.2025.103332},
   DOI={10.1016/j.inffus.2025.103332},
   journal={Information Fusion},
   publisher={Elsevier BV},
   author={Zhuang, Zhenzhen and Chen, Jiandong and Xu, Hongfeng and Jiang, Yuwen and Lin, Jialiang},
   year={2025},
   month=dec, pages={103332} }

@misc{peerreviewas,
      title={Peer Review as A Multi-Turn and Long-Context Dialogue with Role-Based Interactions}, 
      author={Cheng Tan and Dongxin Lyu and Siyuan Li and Zhangyang Gao and Jingxuan Wei and Siqi Ma and Zicheng Liu and Stan Z. Li},
      year={2024},
      eprint={2406.05688},
      archivePrefix={arXiv},
      primaryClass={cs.CL},
      url={https://arxiv.org/abs/2406.05688}, 
}

@misc{openreview,
      title={OpenReviewer: A Specialized Large Language Model for Generating Critical Scientific Paper Reviews}, 
      author={Maximilian Idahl and Zahra Ahmadi},
      year={2025},
      eprint={2412.11948},
      archivePrefix={arXiv},
      primaryClass={cs.AI},
      url={https://arxiv.org/abs/2412.11948}, 
}

@misc{reviewer2,
      title={Reviewer2: Optimizing Review Generation Through Prompt Generation}, 
      author={Zhaolin Gao and Kianté Brantley and Thorsten Joachims},
      year={2024},
      eprint={2402.10886},
      archivePrefix={arXiv},
      primaryClass={cs.CL},
      url={https://arxiv.org/abs/2402.10886}, 
}

@inproceedings{peerread,
    title = "A Dataset of Peer Reviews ({P}eer{R}ead): Collection, Insights and {NLP} Applications",
    author = "Kang, Dongyeop  and
      Ammar, Waleed  and
      Dalvi, Bhavana  and
      van Zuylen, Madeleine  and
      Kohlmeier, Sebastian  and
      Hovy, Eduard  and
      Schwartz, Roy",
    editor = "Walker, Marilyn  and
      Ji, Heng  and
      Stent, Amanda",
    booktitle = "Proceedings of the 2018 Conference of the North {A}merican Chapter of the Association for Computational Linguistics: Human Language Technologies, Volume 1 (Long Papers)",
    month = jun,
    year = "2018",
    address = "New Orleans, Louisiana",
    publisher = "Association for Computational Linguistics",
    url = "https://aclanthology.org/N18-1149/",
    doi = "10.18653/v1/N18-1149",
    pages = "1647--1661"
}

@misc{nlpee,
      title={NLPeer: A Unified Resource for the Computational Study of Peer Review}, 
      author={Nils Dycke and Ilia Kuznetsov and Iryna Gurevych},
      year={2023},
      eprint={2211.06651},
      archivePrefix={arXiv},
      primaryClass={cs.CL},
      url={https://arxiv.org/abs/2211.06651}, 
}

@misc{limgen,
      title={LimGen: Probing the LLMs for Generating Suggestive Limitations of Research Papers}, 
      author={Abdur Rahman Bin Md Faizullah and Ashok Urlana and Rahul Mishra},
      year={2024},
      eprint={2403.15529},
      archivePrefix={arXiv},
      primaryClass={cs.CL},
      url={https://arxiv.org/abs/2403.15529}, 
}

@misc{marg,
      title={MARG: Multi-Agent Review Generation for Scientific Papers}, 
      author={Mike D'Arcy and Tom Hope and Larry Birnbaum and Doug Downey},
      year={2024},
      eprint={2401.04259},
      archivePrefix={arXiv},
      primaryClass={cs.CL},
      url={https://arxiv.org/abs/2401.04259}, 
}

@inproceedings{
mamorx,
title={{MAMORX}: Multi-agent Multi-Modal Scientific Review Generation with External Knowledge},
author={Pawin Taechoyotin and Guanchao Wang and Tong Zeng and Bradley Sides and Daniel Acuna},
booktitle={Neurips 2024 Workshop Foundation Models for Science: Progress, Opportunities, and Challenges},
year={2024},
url={https://openreview.net/forum?id=frvkE8rCfX}
}

@misc{deepreview,
      title={DeepReview: Improving LLM-based Paper Review with Human-like Deep Thinking Process}, 
      author={Minjun Zhu and Yixuan Weng and Linyi Yang and Yue Zhang},
      year={2025},
      eprint={2503.08569},
      archivePrefix={arXiv},
      primaryClass={cs.CL},
      url={https://arxiv.org/abs/2503.08569}, 
}

@misc{agentreview,
      title={AgentReview: Exploring Peer Review Dynamics with LLM Agents}, 
      author={Yiqiao Jin and Qinlin Zhao and Yiyang Wang and Hao Chen and Kaijie Zhu and Yijia Xiao and Jindong Wang},
      year={2024},
      eprint={2406.12708},
      archivePrefix={arXiv},
      primaryClass={cs.CL},
      url={https://arxiv.org/abs/2406.12708}, 
}

@misc{reviewagents,
      title={ReviewAgents: Bridging the Gap Between Human and AI-Generated Paper Reviews}, 
      author={Xian Gao and Jiacheng Ruan and Zongyun Zhang and Jingsheng Gao and Ting Liu and Yuzhuo Fu},
      year={2025},
      eprint={2503.08506},
      archivePrefix={arXiv},
      primaryClass={cs.CL},
      url={https://arxiv.org/abs/2503.08506}, 
}

@misc{remor,
      title={REMOR: Automated Peer Review Generation with LLM Reasoning and Multi-Objective Reinforcement Learning}, 
      author={Pawin Taechoyotin and Daniel Acuna},
      year={2025},
      eprint={2505.11718},
      archivePrefix={arXiv},
      primaryClass={cs.AI},
      url={https://arxiv.org/abs/2505.11718}, 
}

@misc{deepseekmath,
      title={DeepSeekMath: Pushing the Limits of Mathematical Reasoning in Open Language Models}, 
      author={Zhihong Shao and Peiyi Wang and Qihao Zhu and Runxin Xu and Junxiao Song and Xiao Bi and Haowei Zhang and Mingchuan Zhang and Y. K. Li and Y. Wu and Daya Guo},
      year={2024},
      eprint={2402.03300},
      archivePrefix={arXiv},
      primaryClass={cs.CL},
      url={https://arxiv.org/abs/2402.03300}, 
}

@misc{cycleresearcher,
      title={CycleResearcher: Improving Automated Research via Automated Review}, 
      author={Yixuan Weng and Minjun Zhu and Guangsheng Bao and Hongbo Zhang and Jindong Wang and Yue Zhang and Linyi Yang},
      year={2025},
      eprint={2411.00816},
      archivePrefix={arXiv},
      primaryClass={cs.CL},
      url={https://arxiv.org/abs/2411.00816}, 
}

@misc{multinews,
      title={Multi-News: a Large-Scale Multi-Document Summarization Dataset and Abstractive Hierarchical Model}, 
      author={Alexander R. Fabbri and Irene Li and Tianwei She and Suyi Li and Dragomir R. Radev},
      year={2019},
      eprint={1906.01749},
      archivePrefix={arXiv},
      primaryClass={cs.CL},
      url={https://arxiv.org/abs/1906.01749}, 
}

@ARTICLE{8736808,
  author={Li, Wei and Zhuge, Hai},
  journal={IEEE Transactions on Knowledge and Data Engineering}, 
  title={Abstractive Multi-Document Summarization Based on Semantic Link Network}, 
  year={2021},
  volume={33},
  number={1},
  pages={43-54},
  doi={10.1109/TKDE.2019.2922957}}

@inproceedings{chamoun-etal-2024-automated,
    title = "Automated Focused Feedback Generation for Scientific Writing Assistance",
    author = "Chamoun, Eric  and
      Schlichtkrull, Michael  and
      Vlachos, Andreas",
    editor = "Ku, Lun-Wei  and
      Martins, Andre  and
      Srikumar, Vivek",
    booktitle = "Findings of the Association for Computational Linguistics: ACL 2024",
    month = aug,
    year = "2024",
    address = "Bangkok, Thailand",
    publisher = "Association for Computational Linguistics",
    url = "https://aclanthology.org/2024.findings-acl.580/",
    doi = "10.18653/v1/2024.findings-acl.580",
    pages = "9742--9763"
}

@article{hendrickx2021reject,
  author  = {Hendrickx, Kilian and Perini, Lorenzo and Van der Plas, Dries and Meert, Wannes and Davis, Jesse},
  title   = {Machine Learning with a Reject Option: A Survey},
  journal = {IEEE Transactions on Knowledge and Data Engineering},
  year    = {2021},
  volume  = {33},
  number  = {1},
  pages   = {43--54},
  doi     = {https://doi.org/10.48550/arXiv.2107.11277}
}

@inproceedings{muangkammuen2022,
  author    = {Muangkammuen, Panitan and Fukumoto, Fumiyo and Li, Jiyi and Suzuki, Yoshimi},
  title     = {Exploiting Labeled and Unlabeled Data via Transformer Fine-tuning for Peer-Review Score Prediction},
  booktitle = {Findings of the Association for Computational Linguistics: EMNLP 2022},
  year      = {2022},
  pages     = {2233--2240},
  url       = {https://aclanthology.org/2022.findings-emnlp.164}
}

@inproceedings{yuan2021automate,
  author    = {Weizhe Yuan and Pengfei Liu and Graham Neubig},
  title     = {Can We Automate Scientific Reviewing?},
  booktitle = {Proceedings of the 2021 Conference on Empirical Methods in Natural Language Processing (EMNLP)},
  year      = {2021},
  pages     = {6269--6284},
  url       = {https://arxiv.org/abs/2102.00176}
}

@inproceedings{cheng-etal-2020-ape,
  title = "{APE}: Argument Pair Extraction from Peer Review and Rebuttal via Multi-task Learning",
  author = {Cheng, Liying and Bing, Lidong and Yu, Qian and Lu, Wei and Si, Luo},
  booktitle = {Proceedings of the 2020 Conference on Empirical Methods in Natural Language Processing (EMNLP)},
  year = {2020},
  pages = {7000--7011},
  url = {https://aclanthology.org/2020.emnlp-main.569},
  doi = {10.18653/v1/2020.emnlp-main.569}
}

@inproceedings{purkayastha-etal-2023-exploring,
  author    = {Sukannya Purkayastha and Anne Lauscher and Iryna Gurevych},
  title     = {Exploring Jiu‑Jitsu Argumentation for Writing Peer Review Rebuttals},
  booktitle = {Proceedings of the 2023 Conference on Empirical Methods in Natural Language Processing},
  year      = {2023},
  pages     = {14479--14495},
  doi       = {10.18653/v1/2023.emnlp-main.894},
  url       = {https://aclanthology.org/2023.emnlp-main.894}
}

@inproceedings{li-etal-2023-summarizing,
  author    = {Li, Miao and Hovy, Eduard and Lau, Jey Han},
  title     = {Summarizing Multiple Documents with Conversational Structure for Meta‑Review Generation},
  booktitle = {Findings of the Association for Computational Linguistics: EMNLP 2023},
  year      = {2023},
  pages     = {7089--7112},
  doi       = {10.18653/v1/2023.findings-emnlp.472}
}

@article{kumar2024pipeline,
author = {Kumar, Asheesh and Ghosal, Tirthankar and Bhattacharjee, Saprativa and Ekbal, Asif},
title = {Towards automated meta-review generation via an NLP/ML pipeline in different stages of the scholarly peer review process},
year = {2023},
issue_date = {Sep 2024},
publisher = {Springer-Verlag},
address = {Berlin, Heidelberg},
volume = {25},
number = {3},
issn = {1432-5012},
url = {https://doi.org/10.1007/s00799-023-00359-0},
doi = {10.1007/s00799-023-00359-0},
journal = {Int. J. Digit. Libr.},
month = apr,
pages = {493–504},
numpages = {12}
}

@misc{zeng2023checklist,
      title={Scientific Opinion Summarization: Paper Meta-review Generation Dataset, Methods, and Evaluation}, 
      author={Qi Zeng and Mankeerat Sidhu and Ansel Blume and Hou Pong Chan and Lu Wang and Heng Ji},
      year={2024},
      eprint={2305.14647},
      archivePrefix={arXiv},
      primaryClass={cs.CL},
      url={https://arxiv.org/abs/2305.14647}, 
}

@misc{Sukpanichnant2024,
  author = {Sukpanichnant, Purin and Rapberger, Anna and Toni, Francesca},
  title  = {PeerArg: Argumentative Peer Review with LLMs},
  year   = {2024},
  howpublished = {arXiv preprint arXiv:2409.16813},
  url    = {https://arxiv.org/abs/2409.16813}
}

@misc{lin2022moprd,
  author       = {Lin, Jialiang and Song, Jiaxin and Zhou, Zhangping and Chen, Yidong and Shi, Xiaodong},
  title        = {MOPRD: A Multidisciplinary Open Peer Review Dataset},
  howpublished = {arXiv preprint arXiv:2212.04972},
  year         = {2022},
  url          = {https://arxiv.org/abs/2212.04972}
}

@inproceedings{guo2023automatic,
  author    = {Guo, Yanzhu and Shang, Guokan and Rennard, Virgile and Vazirgiannis, Michalis and Clavel, Chlo{\'e}},
  title     = {Automatic Analysis of Substantiation in Scientific Peer Reviews},
  booktitle = {Findings of the Association for Computational Linguistics: EMNLP 2023},
  year      = {2023},
  pages     = {10198--10216},
  doi       = {10.18653/v1/2023.findings-emnlp.684},
  url       = {https://aclanthology.org/2023.findings-emnlp.684/}
}

@inproceedings{du2024critique,
    title = "{LLM}s Assist {NLP} Researchers: Critique Paper (Meta-)Reviewing",
    author = "Du, Jiangshu  and
      Wang, Yibo  and
      Zhao, Wenting  and
      Deng, Zhongfen  and
      Liu, Shuaiqi  and
      Lou, Renze  and
      Zou, Henry Peng  and
      Narayanan Venkit, Pranav  and
      Zhang, Nan  and
      Srinath, Mukund  and
      Zhang, Haoran Ranran  and
      Gupta, Vipul  and
      Li, Yinghui  and
      Li, Tao  and
      Wang, Fei  and
      Liu, Qin  and
      Liu, Tianlin  and
      Gao, Pengzhi  and
      Xia, Congying  and
      Xing, Chen  and
      Jiayang, Cheng  and
      Wang, Zhaowei  and
      Su, Ying  and
      Shah, Raj Sanjay  and
      Guo, Ruohao  and
      Gu, Jing  and
      Li, Haoran  and
      Wei, Kangda  and
      Wang, Zihao  and
      Cheng, Lu  and
      Ranathunga, Surangika  and
      Fang, Meng  and
      Fu, Jie  and
      Liu, Fei  and
      Huang, Ruihong  and
      Blanco, Eduardo  and
      Cao, Yixin  and
      Zhang, Rui  and
      Yu, Philip S.  and
      Yin, Wenpeng",
    editor = "Al-Onaizan, Yaser  and
      Bansal, Mohit  and
      Chen, Yun-Nung",
    booktitle = "Proceedings of the 2024 Conference on Empirical Methods in Natural Language Processing",
    month = nov,
    year = "2024",
    address = "Miami, Florida, USA",
    publisher = "Association for Computational Linguistics",
    url = "https://aclanthology.org/2024.emnlp-main.292/",
    doi = "10.18653/v1/2024.emnlp-main.292",
    pages = "5081--5099"
}

@misc{gpt4slightly,
      title={GPT4 is Slightly Helpful for Peer-Review Assistance: A Pilot Study}, 
      author={Zachary Robertson},
      year={2023},
      eprint={2307.05492},
      archivePrefix={arXiv},
      primaryClass={cs.HC},
      url={https://arxiv.org/abs/2307.05492}, 
}

@misc{liang2023largelanguagemodelsprovide,
      title={Can large language models provide useful feedback on research papers? A large-scale empirical analysis}, 
      author={Weixin Liang and Yuhui Zhang and Hancheng Cao and Binglu Wang and Daisy Ding and Xinyu Yang and Kailas Vodrahalli and Siyu He and Daniel Smith and Yian Yin and Daniel McFarland and James Zou},
      year={2023},
      eprint={2310.01783},
      archivePrefix={arXiv},
      primaryClass={cs.LG},
      url={https://arxiv.org/abs/2310.01783}, 
}

@article{Drori2024Human,
  author    = {Iddo Drori and Dov Te'eni},
  title     = {Human-in-the-Loop AI Reviewing: Feasibility, Opportunities, and Risks},
  journal   = {Journal of the Association for Information Systems},
  year      = {2024},
  volume    = {25},
  number    = {1},
  pages     = {98--109},
  doi       = {10.17705/1jais.00867},
  url       = {https://aisel.aisnet.org/jais/vol25/iss1/7}
}

@article{Hosseini2023Fighting,
  author    = {Hosseini, M. and Horbach, S. P. J. M.},
  title     = {Fighting reviewer fatigue or amplifying bias? Considerations and recommendations for use of ChatGPT and other large language models in scholarly peer review},
  journal   = {Research Integrity and Peer Review},
  year      = {2023},
  volume    = {8},
  number    = {1},
  pages     = {4},
  doi       = {10.1186/s41073-023-00133-5},
  url       = {https://doi.org/10.1186/s41073-023-00133-5}
}

@misc{schintler2023criticalexaminationethicsaimediated,
      title={A Critical Examination of the Ethics of AI-Mediated Peer Review}, 
      author={Laurie A. Schintler and Connie L. McNeely and James Witte},
      year={2023},
      eprint={2309.12356},
      archivePrefix={arXiv},
      primaryClass={cs.CY},
      url={https://arxiv.org/abs/2309.12356}, 
}

@article{Lee2025Role,
  author    = {Lee, J and Lee, J and Yoo, J. J.},
  title     = {The role of large language models in the peer-review process: opportunities and challenges for medical journal reviewers and editors},
  journal   = {Journal of Educational Evaluation for Health Professions},
  year      = {2025},
  volume    = {22},
  pages     = {4},
  doi       = {10.3352/jeehp.2025.22.4},
  pmid      = {40122672},
  pmcid     = {PMC11952698},
  note      = {Epub 2025 Jan 16}
}

@inproceedings{zhou-etal-2024-llm,
    title = "Is {LLM} a Reliable Reviewer? A Comprehensive Evaluation of {LLM} on Automatic Paper Reviewing Tasks",
    author = "Zhou, Ruiyang  and
      Chen, Lu  and
      Yu, Kai",
    editor = "Calzolari, Nicoletta  and
      Kan, Min-Yen  and
      Hoste, Veronique  and
      Lenci, Alessandro  and
      Sakti, Sakriani  and
      Xue, Nianwen",
    booktitle = "Proceedings of the 2024 Joint International Conference on Computational Linguistics, Language Resources and Evaluation (LREC-COLING 2024)",
    month = may,
    year = "2024",
    address = "Torino, Italia",
    publisher = "ELRA and ICCL",
    url = "https://aclanthology.org/2024.lrec-main.816/",
    pages = "9340--9351"
}

@misc{yu2024automatedpeerreviewingpaper,
      title={Automated Peer Reviewing in Paper SEA: Standardization, Evaluation, and Analysis}, 
      author={Jianxiang Yu and Zichen Ding and Jiaqi Tan and Kangyang Luo and Zhenmin Weng and Chenghua Gong and Long Zeng and Renjing Cui and Chengcheng Han and Qiushi Sun and Zhiyong Wu and Yunshi Lan and Xiang Li},
      year={2024},
      eprint={2407.12857},
      archivePrefix={arXiv},
      primaryClass={cs.CL},
      url={https://arxiv.org/abs/2407.12857}, 
}

@misc{liu2023reviewergptexploratorystudyusing,
      title={ReviewerGPT? An Exploratory Study on Using Large Language Models for Paper Reviewing}, 
      author={Ryan Liu and Nihar B. Shah},
      year={2023},
      eprint={2306.00622},
      archivePrefix={arXiv},
      primaryClass={cs.CL},
      url={https://arxiv.org/abs/2306.00622}, 
}

@misc{gu2025surveyllmasajudge,
      title={A Survey on LLM-as-a-Judge}, 
      author={Jiawei Gu and Xuhui Jiang and Zhichao Shi and Hexiang Tan and Xuehao Zhai and Chengjin Xu and Wei Li and Yinghan Shen and Shengjie Ma and Honghao Liu and Saizhuo Wang and Kun Zhang and Yuanzhuo Wang and Wen Gao and Lionel Ni and Jian Guo},
      year={2025},
      eprint={2411.15594},
      archivePrefix={arXiv},
      primaryClass={cs.CL},
      url={https://arxiv.org/abs/2411.15594}, 
}

@misc{shin2025mindblindspotsfocuslevel,
      title={Mind the Blind Spots: A Focus-Level Evaluation Framework for LLM Reviews}, 
      author={Hyungyu Shin and Jingyu Tang and Yoonjoo Lee and Nayoung Kim and Hyunseung Lim and Ji Yong Cho and Hwajung Hong and Moontae Lee and Juho Kim},
      year={2025},
      eprint={2502.17086},
      archivePrefix={arXiv},
      primaryClass={cs.CL},
      url={https://arxiv.org/abs/2502.17086}, 
}

@misc{ning2025picopeerreviewllms,
      title={PiCO: Peer Review in LLMs based on the Consistency Optimization}, 
      author={Kun-Peng Ning and Shuo Yang and Yu-Yang Liu and Jia-Yu Yao and Zhen-Hui Liu and Yong-Hong Tian and Yibing Song and Li Yuan},
      year={2025},
      eprint={2402.01830},
      archivePrefix={arXiv},
      primaryClass={cs.CL},
      url={https://arxiv.org/abs/2402.01830}, 
}

@misc{ye2024yetrevealingrisksutilizing,
      title={Are We There Yet? Revealing the Risks of Utilizing Large Language Models in Scholarly Peer Review}, 
      author={Rui Ye and Xianghe Pang and Jingyi Chai and Jiaao Chen and Zhenfei Yin and Zhen Xiang and Xiaowen Dong and Jing Shao and Siheng Chen},
      year={2024},
      eprint={2412.01708},
      archivePrefix={arXiv},
      primaryClass={cs.CL},
      url={https://arxiv.org/abs/2412.01708}, 
}

@misc{zhang2025reviewingscientificpaperscritical,
      title={Reviewing Scientific Papers for Critical Problems With Reasoning LLMs: Baseline Approaches and Automatic Evaluation}, 
      author={Tianmai M. Zhang and Neil F. Abernethy},
      year={2025},
      eprint={2505.23824},
      archivePrefix={arXiv},
      primaryClass={cs.CL},
      url={https://arxiv.org/abs/2505.23824}, 
}

@misc{zhao2025sciarenaopenevaluationplatform,
      title={SciArena: An Open Evaluation Platform for Foundation Models in Scientific Literature Tasks}, 
      author={Yilun Zhao and Kaiyan Zhang and Tiansheng Hu and Sihong Wu and Ronan Le Bras and Taira Anderson and Jonathan Bragg and Joseph Chee Chang and Jesse Dodge and Matt Latzke and Yixin Liu and Charles McGrady and Xiangru Tang and Zihang Wang and Chen Zhao and Hannaneh Hajishirzi and Doug Downey and Arman Cohan},
      year={2025},
      eprint={2507.01001},
      archivePrefix={arXiv},
      primaryClass={cs.CL},
      url={https://arxiv.org/abs/2507.01001}, 
}

@misc{chiang2024chatbotarenaopenplatform,
      title={Chatbot Arena: An Open Platform for Evaluating LLMs by Human Preference}, 
      author={Wei-Lin Chiang and Lianmin Zheng and Ying Sheng and Anastasios Nikolas Angelopoulos and Tianle Li and Dacheng Li and Hao Zhang and Banghua Zhu and Michael Jordan and Joseph E. Gonzalez and Ion Stoica},
      year={2024},
      eprint={2403.04132},
      archivePrefix={arXiv},
      primaryClass={cs.AI},
      url={https://arxiv.org/abs/2403.04132}, 
}

@misc{tyser2024aidrivenreviewsystemsevaluating,
      title={AI-Driven Review Systems: Evaluating LLMs in Scalable and Bias-Aware Academic Reviews}, 
      author={Keith Tyser and Ben Segev and Gaston Longhitano and Xin-Yu Zhang and Zachary Meeks and Jason Lee and Uday Garg and Nicholas Belsten and Avi Shporer and Madeleine Udell and Dov Te'eni and Iddo Drori},
      year={2024},
      eprint={2408.10365},
      archivePrefix={arXiv},
      primaryClass={cs.AI},
      url={https://arxiv.org/abs/2408.10365}, 
}

@misc{gao2024retrievalaugmentedgenerationlargelanguage,
      title={Retrieval-Augmented Generation for Large Language Models: A Survey}, 
      author={Yunfan Gao and Yun Xiong and Xinyu Gao and Kangxiang Jia and Jinliu Pan and Yuxi Bi and Yi Dai and Jiawei Sun and Meng Wang and Haofen Wang},
      year={2024},
      eprint={2312.10997},
      archivePrefix={arXiv},
      primaryClass={cs.CL},
      url={https://arxiv.org/abs/2312.10997}, 
}

@inproceedings{Wang_2020,
   title={ReviewRobot: Explainable Paper Review Generation based on Knowledge Synthesis},
   url={http://dx.doi.org/10.18653/v1/2020.inlg-1.44},
   DOI={10.18653/v1/2020.inlg-1.44},
   booktitle={Proceedings of the 13th International Conference on Natural Language Generation},
   publisher={Association for Computational Linguistics},
   author={Wang, Qingyun and Zeng, Qi and Huang, Lifu and Knight, Kevin and Ji, Heng and Rajani, Nazneen Fatema},
   year={2020} }

@misc{chitale2025autorevautomaticpeerreview,
      title={AutoRev: Automatic Peer Review System for Academic Research Papers}, 
      author={Maitreya Prafulla Chitale and Ketaki Mangesh Shetye and Harshit Gupta and Manav Chaudhary and Vasudeva Varma},
      year={2025},
      eprint={2505.14376},
      archivePrefix={arXiv},
      primaryClass={cs.CL},
      url={https://arxiv.org/abs/2505.14376}, 
}

@article{kamoi-etal-2024-llms,
    title = "When Can {LLM}s Actually Correct Their Own Mistakes? A Critical Survey of Self-Correction of {LLM}s",
    author = "Kamoi, Ryo  and
      Zhang, Yusen  and
      Zhang, Nan  and
      Han, Jiawei  and
      Zhang, Rui",
    journal = "Transactions of the Association for Computational Linguistics",
    volume = "12",
    year = "2024",
    address = "Cambridge, MA",
    publisher = "MIT Press",
    url = "https://aclanthology.org/2024.tacl-1.78/",
    doi = "10.1162/tacl_a_00713",
    pages = "1417--1440"
}

@misc{gou2024criticlargelanguagemodels,
      title={CRITIC: Large Language Models Can Self-Correct with Tool-Interactive Critiquing}, 
      author={Zhibin Gou and Zhihong Shao and Yeyun Gong and Yelong Shen and Yujiu Yang and Nan Duan and Weizhu Chen},
      year={2024},
      eprint={2305.11738},
      archivePrefix={arXiv},
      primaryClass={cs.CL},
      url={https://arxiv.org/abs/2305.11738}, 
}

@misc{han2024smalllanguagemodelselfcorrect,
      title={Small Language Model Can Self-correct}, 
      author={Haixia Han and Jiaqing Liang and Jie Shi and Qianyu He and Yanghua Xiao},
      year={2024},
      eprint={2401.07301},
      archivePrefix={arXiv},
      primaryClass={cs.CL},
      url={https://arxiv.org/abs/2401.07301}, 
}

@misc{chang2025treereviewdynamictreequestions,
      title={TreeReview: A Dynamic Tree of Questions Framework for Deep and Efficient LLM-based Scientific Peer Review}, 
      author={Yuan Chang and Ziyue Li and Hengyuan Zhang and Yuanbo Kong and Yanru Wu and Zhijiang Guo and Ngai Wong},
      year={2025},
      eprint={2506.07642},
      archivePrefix={arXiv},
      primaryClass={cs.CL},
      url={https://arxiv.org/abs/2506.07642}, 
}

@misc{li2020controllablepersonalizedreviewgeneration,
      title={Towards Controllable and Personalized Review Generation}, 
      author={Pan Li and Alexander Tuzhilin},
      year={2020},
      eprint={1910.03506},
      archivePrefix={arXiv},
      primaryClass={cs.CL},
      url={https://arxiv.org/abs/1910.03506}, 
}

@misc{novikov2025alphaevolvecodingagentscientific,
      title={AlphaEvolve: A coding agent for scientific and algorithmic discovery}, 
      author={Alexander Novikov and Ngân Vũ and Marvin Eisenberger and Emilien Dupont and Po-Sen Huang and Adam Zsolt Wagner and Sergey Shirobokov and Borislav Kozlovskii and Francisco J. R. Ruiz and Abbas Mehrabian and M. Pawan Kumar and Abigail See and Swarat Chaudhuri and George Holland and Alex Davies and Sebastian Nowozin and Pushmeet Kohli and Matej Balog},
      year={2025},
      eprint={2506.13131},
      archivePrefix={arXiv},
      primaryClass={cs.AI},
      url={https://arxiv.org/abs/2506.13131}, 
}

@inproceedings{bleu,
author = {Papineni, Kishore and Roukos, Salim and Ward, Todd and Zhu, Wei-Jing},
title = {BLEU: a method for automatic evaluation of machine translation},
year = {2002},
publisher = {Association for Computational Linguistics},
address = {USA},
url = {https://doi.org/10.3115/1073083.1073135},
doi = {10.3115/1073083.1073135},
booktitle = {Proceedings of the 40th Annual Meeting on Association for Computational Linguistics},
pages = {311–318},
numpages = {8},
location = {Philadelphia, Pennsylvania},
series = {ACL '02}
}

@inproceedings{rouge,
    title = "{ROUGE}: A Package for Automatic Evaluation of Summaries",
    author = "Lin, Chin-Yew",
    booktitle = "Text Summarization Branches Out",
    month = jul,
    year = "2004",
    address = "Barcelona, Spain",
    publisher = "Association for Computational Linguistics",
    url = "https://aclanthology.org/W04-1013/",
    pages = "74--81"
}

@misc{rouge2,
      title={ROUGE 2.0: Updated and Improved Measures for Evaluation of Summarization Tasks}, 
      author={Kavita Ganesan},
      year={2018},
      eprint={1803.01937},
      archivePrefix={arXiv},
      primaryClass={cs.IR},
      url={https://arxiv.org/abs/1803.01937}, 
}

@misc{bertscore,
      title={BERTScore: Evaluating Text Generation with BERT}, 
      author={Tianyi Zhang and Varsha Kishore and Felix Wu and Kilian Q. Weinberger and Yoav Artzi},
      year={2020},
      eprint={1904.09675},
      archivePrefix={arXiv},
      primaryClass={cs.CL},
      url={https://arxiv.org/abs/1904.09675}, 
}

@article{Buyukkaramikli2019TECHVER,
  author    = {Büyükkaramikli, N. C. and Rutten-van Mölken, M. P. M. H. and Severens, J. L. and Al, M.},
  title     = {{TECH-VER}: A Verification Checklist to Reduce Errors in Models and Improve Their Credibility},
  journal   = {Pharmacoeconomics},
  year      = {2019},
  volume    = {37},
  number    = {11},
  pages     = {1391--1408},
  month     = {Nov},
  doi       = {10.1007/s40273-019-00844-y},
  pmid      = {31705406},
  pmcid     = {PMC6860463}
}

@misc{lu2025identifyingaspectspeerreviews,
      title={Identifying Aspects in Peer Reviews}, 
      author={Sheng Lu and Ilia Kuznetsov and Iryna Gurevych},
      year={2025},
      eprint={2504.06910},
      archivePrefix={arXiv},
      primaryClass={cs.CL},
      url={https://arxiv.org/abs/2504.06910}, 
}

@misc{zhang2025adversarialtestingllmsinsights,
      title={Adversarial Testing in LLMs: Insights into Decision-Making Vulnerabilities}, 
      author={Lili Zhang and Haomiaomiao Wang and Long Cheng and Libao Deng and Tomas Ward},
      year={2025},
      eprint={2505.13195},
      archivePrefix={arXiv},
      primaryClass={cs.AI},
      url={https://arxiv.org/abs/2505.13195}, 
}

@misc{liao2025redteamcuarealisticadversarialtesting,
      title={RedTeamCUA: Realistic Adversarial Testing of Computer-Use Agents in Hybrid Web-OS Environments}, 
      author={Zeyi Liao and Jaylen Jones and Linxi Jiang and Eric Fosler-Lussier and Yu Su and Zhiqiang Lin and Huan Sun},
      year={2025},
      eprint={2505.21936},
      archivePrefix={arXiv},
      primaryClass={cs.CL},
      url={https://arxiv.org/abs/2505.21936}, 
}

@misc{allamanis2024unsupervisedevaluationcodellms,
      title={Unsupervised Evaluation of Code LLMs with Round-Trip Correctness}, 
      author={Miltiadis Allamanis and Sheena Panthaplackel and Pengcheng Yin},
      year={2024},
      eprint={2402.08699},
      archivePrefix={arXiv},
      primaryClass={cs.SE},
      url={https://arxiv.org/abs/2402.08699}, 
}

@misc{lin2023llmevalunifiedmultidimensionalautomatic,
      title={LLM-Eval: Unified Multi-Dimensional Automatic Evaluation for Open-Domain Conversations with Large Language Models}, 
      author={Yen-Ting Lin and Yun-Nung Chen},
      year={2023},
      eprint={2305.13711},
      archivePrefix={arXiv},
      primaryClass={cs.CL},
      url={https://arxiv.org/abs/2305.13711}, 
}

@article{Drozdz2024PeerReview,
  author    = {Drozdz, J. A. and Ladomery, M. R.},
  title     = {The Peer Review Process: Past, Present, and Future},
  journal   = {British Journal of Biomedical Science},
  year      = {2024},
  volume    = {81},
  pages     = {12054},
  month     = {Jun 17},
  doi       = {10.3389/bjbs.2024.12054},
  pmid      = {38952614},
  pmcid     = {PMC11215012}
}

@misc{peersum,
      title={PeerSum: A Peer Review Dataset for Abstractive Multi-document Summarization}, 
      author={Miao Li and Jianzhong Qi and Jey Han Lau},
      year={2022},
      eprint={2203.01769},
      archivePrefix={arXiv},
      primaryClass={cs.IR},
      url={https://arxiv.org/abs/2203.01769}, 
}

@misc{hu2021loralowrankadaptationlarge,
      title={LoRA: Low-Rank Adaptation of Large Language Models}, 
      author={Edward J. Hu and Yelong Shen and Phillip Wallis and Zeyuan Allen-Zhu and Yuanzhi Li and Shean Wang and Lu Wang and Weizhu Chen},
      year={2021},
      eprint={2106.09685},
      archivePrefix={arXiv},
      primaryClass={cs.CL},
      url={https://arxiv.org/abs/2106.09685}, 
}

@inproceedings{dycke-etal-2023-overview,
    title = "Overview of {P}rag{T}ag-2023: Low-Resource Multi-Domain Pragmatic Tagging of Peer Reviews",
    author = "Dycke, Nils  and
      Kuznetsov, Ilia  and
      Gurevych, Iryna",
    editor = "Alshomary, Milad  and
      Chen, Chung-Chi  and
      Muresan, Smaranda  and
      Park, Joonsuk  and
      Romberg, Julia",
    booktitle = "Proceedings of the 10th Workshop on Argument Mining",
    month = dec,
    year = "2023",
    address = "Singapore",
    publisher = "Association for Computational Linguistics",
    url = "https://aclanthology.org/2023.argmining-1.21/",
    doi = "10.18653/v1/2023.argmining-1.21",
    pages = "187--196"
}

@inproceedings{teufel-etal-1999-annotation,
    title = "An annotation scheme for discourse-level argumentation in research articles",
    author = "Teufel, Simone  and
      Carletta, Jean  and
      Moens, Marc",
    editor = "Thompson, Henry S.  and
      Lascarides, Alex",
    booktitle = "Ninth Conference of the {E}uropean Chapter of the Association for Computational Linguistics",
    month = jun,
    year = "1999",
    address = "Bergen, Norway",
    publisher = "Association for Computational Linguistics",
    url = "https://aclanthology.org/E99-1015/",
    pages = "110--117"
}

@misc{mistral7b,
  author       = {Mistral AI},
  title        = {Mistral-7B-Instruct-v0.2},
  year         = {2023},
  howpublished = {\url{https://huggingface.co/mistralai/Mistral-7B-Instruct-v0.2}},
  note         = {Accessed: 2025-07-28}
}

@misc{qwen2.5,
    title = {Qwen2.5: A Party of Foundation Models},
    url = {https://qwenlm.github.io/blog/qwen2.5/},
    author = {Qwen Team},
    month = {September},
    year = {2024}
}

@misc{qwen,
      title={Qwen Technical Report}, 
      author={Jinze Bai and Shuai Bai and Yunfei Chu and Zeyu Cui and Kai Dang and Xiaodong Deng and Yang Fan and Wenbin Ge and Yu Han and Fei Huang and Binyuan Hui and Luo Ji and Mei Li and Junyang Lin and Runji Lin and Dayiheng Liu and Gao Liu and Chengqiang Lu and Keming Lu and Jianxin Ma and Rui Men and Xingzhang Ren and Xuancheng Ren and Chuanqi Tan and Sinan Tan and Jianhong Tu and Peng Wang and Shijie Wang and Wei Wang and Shengguang Wu and Benfeng Xu and Jin Xu and An Yang and Hao Yang and Jian Yang and Shusheng Yang and Yang Yao and Bowen Yu and Hongyi Yuan and Zheng Yuan and Jianwei Zhang and Xingxuan Zhang and Yichang Zhang and Zhenru Zhang and Chang Zhou and Jingren Zhou and Xiaohuan Zhou and Tianhang Zhu},
      year={2023},
      eprint={2309.16609},
      archivePrefix={arXiv},
      primaryClass={cs.CL},
      url={https://arxiv.org/abs/2309.16609}, 
}

@misc{mistrallarge,
  author       = {Mistral AI},
  title        = {Mistral Large},
  year         = {2024},
  howpublished = {\url{https://mistral.ai/news/mistral-large/}},
  note         = {Accessed: 2025-07-28}
}

@article{llama3modelcard,

title={Llama 3 Model Card},

author={AI@Meta},

year={2024},

url = {https://github.com/meta-llama/llama3/blob/main/MODEL_CARD.md}

}

@misc{deepseekr1,
  author       = {Daya Guo and Dejian Yang and Haowei Zhang and Junxiao Song and Ruoyu Zhang and Runxin Xu and Qihao Zhu and Shirong Ma and Peiyi Wang and Xiao Bi and others},
  title        = {DeepSeek-R1: Incentivizing Reasoning Capability in LLMs via Reinforcement Learning},
  year         = {2024},
  archivePrefix = {arXiv},
  eprint       = {2501.12948},
  primaryClass = {cs.CL},
  url          = {https://arxiv.org/abs/2501.12948}
}

@misc{GROBID,
    title = {GROBID},
    howpublished = {\url{https://github.com/kermitt2/grobid}},
    publisher = {GitHub},
    year = {2008--2025},
    archivePrefix = {swh},
    eprint = {1:dir:dab86b296e3c3216e2241968f0d63b68e8209d3c}
}

@misc{ScienceParse,
    title = {Science Parse},
    howpublished = {\url{https://github.com/allenai/science-parse}},
    publisher = {GitHub},
    year = {2017--2025},
}

@misc{Marker,
    title = {Marker},
    howpublished = {\url{https://github.com/datalab-to/marker}},
    publisher = {GitHub},
    year = {2023--2025},
}

@misc{2024magic-doc,
    title={Magic-Doc: A Toolkit that Converts Multiple File Types to Markdown},
    author={Magic-Doc Contributors},
    howpublished = {\url{https://github.com/InternLM/magic-doc}},
    year={2024}
}

@misc{bommasani2021opportunities,
      title={On the Opportunities and Risks of Foundation Models}, 
      author={Rishi Bommasani and Drew A. Hudson and Ehsan Adeli and Russ Altman and Simran Arora and Sydney von Arx and Michael S. Bernstein and Jeannette Bohg and Antoine Bosselut and Emma Brunskill and Erik Brynjolfsson and Shyamal Buch and Dallas Card and Rodrigo Castellon and Niladri Chatterji and Annie Chen and Kathleen Creel and Jared Quincy Davis and Dora Demszky and Chris Donahue and Moussa Doumbouya and Esin Durmus and Stefano Ermon and John Etchemendy and Kawin Ethayarajh and Li Fei-Fei and Chelsea Finn and Trevor Gale and Lauren Gillespie and Karan Goel and Noah Goodman and Shelby Grossman and Neel Guha and Tatsunori Hashimoto and Peter Henderson and John Hewitt and Daniel E. Ho and Jenny Hong and Kyle Hsu and Jing Huang and Thomas Icard and Saahil Jain and Dan Jurafsky and Pratyusha Kalluri and Siddharth Karamcheti and Geoff Keeling and Fereshte Khani and Omar Khattab and Pang Wei Koh and Mark Krass and Ranjay Krishna and Rohith Kuditipudi and Ananya Kumar and Faisal Ladhak and Mina Lee and Tony Lee and Jure Leskovec and Isabelle Levent and Xiang Lisa Li and Xuechen Li and Tengyu Ma and Ali Malik and Christopher D. Manning and Suvir Mirchandani and Eric Mitchell and Zanele Munyikwa and Suraj Nair and Avanika Narayan and Deepak Narayanan and Ben Newman and Allen Nie and Juan Carlos Niebles and Hamed Nilforoshan and Julian Nyarko and Giray Ogut and Laurel Orr and Isabel Papadimitriou and Joon Sung Park and Chris Piech and Eva Portelance and Christopher Potts and Aditi Raghunathan and Rob Reich and Hongyu Ren and Frieda Rong and Yusuf Roohani and Camilo Ruiz and Jack Ryan and Christopher Ré and Dorsa Sadigh and Shiori Sagawa and Keshav Santhanam and Andy Shih and Krishnan Srinivasan and Alex Tamkin and Rohan Taori and Armin W. Thomas and Florian Tramèr and Rose E. Wang and William Wang and Bohan Wu and Jiajun Wu and Yuhuai Wu and Sang Michael Xie and Michihiro Yasunaga and Jiaxuan You and Matei Zaharia and Michael Zhang and Tianyi Zhang and Xikun Zhang and Yuhui Zhang and Lucia Zheng and Kaitlyn Zhou and Percy Liang},
      year={2022},
      eprint={2108.07258},
      archivePrefix={arXiv},
      primaryClass={cs.LG},
      url={https://arxiv.org/abs/2108.07258}, 
}

@inproceedings{vaswani2017attention,
  title={Attention is all you need},
  author={Vaswani, Ashish and Shazeer, Noam and Parmar, Niki and Uszkoreit, Jakob and Jones, Llion and Gomez, Aidan N and Kaiser, Łukasz and Polosukhin, Illia},
  booktitle={Advances in Neural Information Processing Systems},
  pages={5998--6008},
  year={2017}
}

@article{Kinney2023TheSS,
  title={The Semantic Scholar Open Data Platform},
  author={Rodney Michael Kinney and Chloe Anastasiades and Russell Authur and Iz Beltagy and Jonathan Bragg and Alexandra Buraczynski and Isabel Cachola and Stefan Candra and Yoganand Chandrasekhar and Arman Cohan and Miles Crawford and Doug Downey and Jason Dunkelberger and Oren Etzioni and Rob Evans and Sergey Feldman and Joseph Gorney and David W. Graham and F.Q. Hu and Regan Huff and Daniel King and Sebastian Kohlmeier and Bailey Kuehl and Michael Langan and Daniel Lin and Haokun Liu and Kyle Lo and Jaron Lochner and Kelsey MacMillan and Tyler C. Murray and Christopher Newell and Smita R Rao and Shaurya Rohatgi and Paul Sayre and Zejiang Shen and Amanpreet Singh and Luca Soldaini and Shivashankar Subramanian and A. Tanaka and Alex D Wade and Linda M. Wagner and Lucy Lu Wang and Christopher Wilhelm and Caroline Wu and Jiangjiang Yang and Angele Zamarron and Madeleine van Zuylen and Daniel S. Weld},
  journal={ArXiv},
  year={2023},
  volume={abs/2301.10140},
  url={https://api.semanticscholar.org/CorpusID:256194545}
}

@misc{radford2019language,
  title={Language Models are Unsupervised Multitask Learners},
  author={Radford, Alec and Wu, Jeffrey and Child, Rewon and Luan, David and Amodei, Dario and Sutskever, Ilya},
  year={2019},
  howpublished={OpenAI Technical Report},
  note={\url{https://cdn.openai.com/better-language-models/language_models_are_unsupervised_multitask_learners.pdf}}
}

@misc{zhao2026survey,
      title={A Survey of Large Language Models}, 
      author={Wayne Xin Zhao and Kun Zhou and Junyi Li and Tianyi Tang and Xiaolei Wang and Yupeng Hou and Yingqian Min and Beichen Zhang and Junjie Zhang and Zican Dong and Yifan Du and Chen Yang and Yushuo Chen and Zhipeng Chen and Jinhao Jiang and Ruiyang Ren and Yifan Li and Xinyu Tang and Zikang Liu and Peiyu Liu and Jian-Yun Nie and Ji-Rong Wen},
      year={2026},
      eprint={2303.18223},
      archivePrefix={arXiv},
      primaryClass={cs.CL},
      url={https://arxiv.org/abs/2303.18223}, 
}

@misc{touvron2023llama,
      title={LLaMA: Open and Efficient Foundation Language Models}, 
      author={Hugo Touvron and Thibaut Lavril and Gautier Izacard and Xavier Martinet and Marie-Anne Lachaux and Timothée Lacroix and Baptiste Rozière and Naman Goyal and Eric Hambro and Faisal Azhar and Aurelien Rodriguez and Armand Joulin and Edouard Grave and Guillaume Lample},
      year={2023},
      eprint={2302.13971},
      archivePrefix={arXiv},
      primaryClass={cs.CL},
      url={https://arxiv.org/abs/2302.13971}, 
}

@misc{christiano2023deepreinforcementlearninghuman,
      title={Deep reinforcement learning from human preferences}, 
      author={Paul Christiano and Jan Leike and Tom B. Brown and Miljan Martic and Shane Legg and Dario Amodei},
      year={2023},
      eprint={1706.03741},
      archivePrefix={arXiv},
      primaryClass={stat.ML},
      url={https://arxiv.org/abs/1706.03741}, 
}

@misc{ziegler2020finetuninglanguagemodelshuman,
      title={Fine-Tuning Language Models from Human Preferences}, 
      author={Daniel M. Ziegler and Nisan Stiennon and Jeffrey Wu and Tom B. Brown and Alec Radford and Dario Amodei and Paul Christiano and Geoffrey Irving},
      year={2020},
      eprint={1909.08593},
      archivePrefix={arXiv},
      primaryClass={cs.CL},
      url={https://arxiv.org/abs/1909.08593}, 
}

@misc{bai2022traininghelpfulharmlessassistant,
      title={Training a Helpful and Harmless Assistant with Reinforcement Learning from Human Feedback}, 
      author={Yuntao Bai and Andy Jones and Kamal Ndousse and Amanda Askell and Anna Chen and Nova DasSarma and Dawn Drain and Stanislav Fort and Deep Ganguli and Tom Henighan and Nicholas Joseph and Saurav Kadavath and Jackson Kernion and Tom Conerly and Sheer El-Showk and Nelson Elhage and Zac Hatfield-Dodds and Danny Hernandez and Tristan Hume and Scott Johnston and Shauna Kravec and Liane Lovitt and Neel Nanda and Catherine Olsson and Dario Amodei and Tom Brown and Jack Clark and Sam McCandlish and Chris Olah and Ben Mann and Jared Kaplan},
      year={2022},
      eprint={2204.05862},
      archivePrefix={arXiv},
      primaryClass={cs.CL},
      url={https://arxiv.org/abs/2204.05862}, 
}

@article{lee2013biaspeerreview,
  title={Bias in peer review},
  author={Lee, Carole J and Sugimoto, Cassidy R and Zhang, Guo and Cronin, Blaise},
  journal={Journal of the American Society for Information Science and Technology},
  volume={64},
  number={1},
  pages={2--17},
  year={2013},
  publisher={Wiley},
  doi={10.1002/asi.22784}
}

@inproceedings{bender2021dangers,
  title={On the dangers of stochastic parrots: Can language models be too big?},
  author={Bender, Emily M and Gebru, Timnit and McMillan-Major, Angelina and Shmitchell, Shmargaret},
  booktitle={Proceedings of the 2021 ACM Conference on Fairness, Accountability, and Transparency},
  pages={610--623},
  year={2021},
  doi={10.1145/3442188.3445922}
}

@misc{swamy2025roadsleadlikelihoodvalue,
      title={All Roads Lead to Likelihood: The Value of Reinforcement Learning in Fine-Tuning}, 
      author={Gokul Swamy and Sanjiban Choudhury and Wen Sun and Zhiwei Steven Wu and J. Andrew Bagnell},
      year={2025},
      eprint={2503.01067},
      archivePrefix={arXiv},
      primaryClass={cs.LG},
      url={https://arxiv.org/abs/2503.01067}, 
}

@inproceedings{asano-etal-2017-reference,
    title = "Reference-based Metrics can be Replaced with Reference-less Metrics in Evaluating Grammatical Error Correction Systems",
    author = "Asano, Hiroki  and
      Mizumoto, Tomoya  and
      Inui, Kentaro",
    editor = "Kondrak, Greg  and
      Watanabe, Taro",
    booktitle = "Proceedings of the Eighth International Joint Conference on Natural Language Processing (Volume 2: Short Papers)",
    month = nov,
    year = "2017",
    address = "Taipei, Taiwan",
    publisher = "Asian Federation of Natural Language Processing",
    url = "https://aclanthology.org/I17-2058/",
    pages = "343--348"
}

@inproceedings{Zero-shot, series={RANLP},
   title={A Practical Survey on Zero-shot Prompt Design for In-context Learning},
   url={http://dx.doi.org/10.26615/978-954-452-092-2_069},
   DOI={10.26615/978-954-452-092-2_069},
   booktitle={Proceedings of the Conference Recent Advances in Natural Language Processing - Large Language Models for Natural Language Processings},
   publisher={INCOMA Ltd., Shoumen, BULGARIA},
   author={Li, Yinheng},
   year={2023},
   pages={641–647},
   collection={RANLP} }

@misc{sivarajkumar2023empiricalevaluationpromptingstrategies,
      title={An Empirical Evaluation of Prompting Strategies for Large Language Models in Zero-Shot Clinical Natural Language Processing}, 
      author={Sonish Sivarajkumar and Mark Kelley and Alyssa Samolyk-Mazzanti and Shyam Visweswaran and Yanshan Wang},
      year={2023},
      eprint={2309.08008},
      archivePrefix={arXiv},
      primaryClass={cs.CL},
      url={https://arxiv.org/abs/2309.08008}, 
}

@misc{rao2020rlcycleganreinforcementlearningaware,
      title={RL-CycleGAN: Reinforcement Learning Aware Simulation-To-Real}, 
      author={Kanishka Rao and Chris Harris and Alex Irpan and Sergey Levine and Julian Ibarz and Mohi Khansari},
      year={2020},
      eprint={2006.09001},
      archivePrefix={arXiv},
      primaryClass={cs.RO},
      url={https://arxiv.org/abs/2006.09001}, 
}

@misc{wei2025aiimperativescalinghighquality,
      title={The AI Imperative: Scaling High-Quality Peer Review in Machine Learning}, 
      author={Qiyao Wei and Samuel Holt and Jing Yang and Markus Wulfmeier and Mihaela van der Schaar},
      year={2025},
      eprint={2506.08134},
      archivePrefix={arXiv},
      primaryClass={cs.AI},
      url={https://arxiv.org/abs/2506.08134}, 
}

@misc{dycke2022yesyesyesproactivedatacollection,
      title={Yes-Yes-Yes: Proactive Data Collection for ACL Rolling Review and Beyond}, 
      author={Nils Dycke and Ilia Kuznetsov and Iryna Gurevych},
      year={2022},
      eprint={2201.11443},
      archivePrefix={arXiv},
      primaryClass={cs.CL},
      url={https://arxiv.org/abs/2201.11443}, 
}

@misc{tang2025airesearcherautonomousscientificinnovation,
      title={AI-Researcher: Autonomous Scientific Innovation}, 
      author={Jiabin Tang and Lianghao Xia and Zhonghang Li and Chao Huang},
      year={2025},
      eprint={2505.18705},
      archivePrefix={arXiv},
      primaryClass={cs.AI},
      url={https://arxiv.org/abs/2505.18705}, 
}

@misc{kovács2025lettucedetecthallucinationdetectionframework,
      title={LettuceDetect: A Hallucination Detection Framework for RAG Applications}, 
      author={Ádám Kovács and Gábor Recski},
      year={2025},
      eprint={2502.17125},
      archivePrefix={arXiv},
      primaryClass={cs.CL},
      url={https://arxiv.org/abs/2502.17125}, 
}

@misc{shuster2021retrievalaugmentationreduceshallucination,
      title={Retrieval Augmentation Reduces Hallucination in Conversation}, 
      author={Kurt Shuster and Spencer Poff and Moya Chen and Douwe Kiela and Jason Weston},
      year={2021},
      eprint={2104.07567},
      archivePrefix={arXiv},
      primaryClass={cs.CL},
      url={https://arxiv.org/abs/2104.07567}, 
}

@misc{lewis2021retrievalaugmentedgenerationknowledgeintensivenlp,
      title={Retrieval-Augmented Generation for Knowledge-Intensive NLP Tasks}, 
      author={Patrick Lewis and Ethan Perez and Aleksandra Piktus and Fabio Petroni and Vladimir Karpukhin and Naman Goyal and Heinrich Küttler and Mike Lewis and Wen-tau Yih and Tim Rocktäschel and Sebastian Riedel and Douwe Kiela},
      year={2021},
      eprint={2005.11401},
      archivePrefix={arXiv},
      primaryClass={cs.CL},
      url={https://arxiv.org/abs/2005.11401}, 
}

@article{Nauta_2023,
   title={From Anecdotal Evidence to Quantitative Evaluation Methods: A Systematic Review on Evaluating Explainable AI},
   volume={55},
   ISSN={1557-7341},
   url={http://dx.doi.org/10.1145/3583558},
   DOI={10.1145/3583558},
   number={13s},
   journal={ACM Computing Surveys},
   publisher={Association for Computing Machinery (ACM)},
   author={Nauta, Meike and Trienes, Jan and Pathak, Shreyasi and Nguyen, Elisa and Peters, Michelle and Schmitt, Yasmin and Schlötterer, Jörg and van Keulen, Maurice and Seifert, Christin},
   year={2023},
   month=jul, pages={1–42} }

@inproceedings{kumar-etal-2024-longform,
    title = "Longform Multimodal Lay Summarization of Scientific Papers: Towards Automatically Generating Science Blogs from Research Articles",
    author = "Kumar, Sandeep  and
      Kohli, Guneet Singh  and
      Ghosal, Tirthankar  and
      Ekbal, Asif",
    editor = "Calzolari, Nicoletta  and
      Kan, Min-Yen  and
      Hoste, Veronique  and
      Lenci, Alessandro  and
      Sakti, Sakriani  and
      Xue, Nianwen",
    booktitle = "Proceedings of the 2024 Joint International Conference on Computational Linguistics, Language Resources and Evaluation (LREC-COLING 2024)",
    month = may,
    year = "2024",
    address = "Torino, Italia",
    publisher = "ELRA and ICCL",
    url = "https://aclanthology.org/2024.lrec-main.942/",
    pages = "10790--10801"
}

@misc{wang2025sciverevaluatingfoundationmodels,
      title={SciVer: Evaluating Foundation Models for Multimodal Scientific Claim Verification}, 
      author={Chengye Wang and Yifei Shen and Zexi Kuang and Arman Cohan and Yilun Zhao},
      year={2025},
      eprint={2506.15569},
      archivePrefix={arXiv},
      primaryClass={cs.CL},
      url={https://arxiv.org/abs/2506.15569}, 
}

@misc{jiang2024bridgingresearchreadersmultimodal,
      title={Bridging Research and Readers: A Multi-Modal Automated Academic Papers Interpretation System}, 
      author={Feng Jiang and Kuang Wang and Haizhou Li},
      year={2024},
      eprint={2401.09150},
      archivePrefix={arXiv},
      primaryClass={cs.CL},
      url={https://arxiv.org/abs/2401.09150}, 
}

@inproceedings{im-etal-2021-self,
    title = "Self-Supervised Multimodal Opinion Summarization",
    author = "Im, Jinbae  and
      Kim, Moonki  and
      Lee, Hoyeop  and
      Cho, Hyunsouk  and
      Chung, Sehee",
    editor = "Zong, Chengqing  and
      Xia, Fei  and
      Li, Wenjie  and
      Navigli, Roberto",
    booktitle = "Proceedings of the 59th Annual Meeting of the Association for Computational Linguistics and the 11th International Joint Conference on Natural Language Processing (Volume 1: Long Papers)",
    month = aug,
    year = "2021",
    address = "Online",
    publisher = "Association for Computational Linguistics",
    url = "https://aclanthology.org/2021.acl-long.33/",
    doi = "10.18653/v1/2021.acl-long.33",
    pages = "388--403"
}

@misc{xu2025chaindraftthinkingfaster,
      title={Chain of Draft: Thinking Faster by Writing Less}, 
      author={Silei Xu and Wenhao Xie and Lingxiao Zhao and Pengcheng He},
      year={2025},
      eprint={2502.18600},
      archivePrefix={arXiv},
      primaryClass={cs.CL},
      url={https://arxiv.org/abs/2502.18600}, 
}

@misc{kennard2021disapere,
  title        = {DISAPERE: A Dataset for Discourse Structure in Peer Review Discussions},
  author       = {Neha Kennard and Tim O'Gorman and Rajarshi Das and Akshay Sharma and Chhandak Bagchi and Matthew Clinton and Pranay Kumar Yelugam and Hamed Zamani and Andrew McCallum},
  year         = {2021},
  eprint       = {2110.08520},
  archivePrefix= {arXiv},
  primaryClass = {cs.CL},
  url          = {https://arxiv.org/abs/2110.08520}
}

@InProceedings{bharti2021peerassist,
author="Bharti, Prabhat Kumar
and Ranjan, Shashi
and Ghosal, Tirthankar
and Agrawal, Mayank
and Ekbal, Asif",
editor="Ke, Hao-Ren
and Lee, Chei Sian
and Sugiyama, Kazunari",
title="PEERAssist: Leveraging on Paper-Review Interactions to Predict Peer Review Decisions",
booktitle="Towards Open and Trustworthy Digital Societies",
year="2021",
publisher="Springer International Publishing",
address="Cham",
pages="421--435",
isbn="978-3-030-91669-5"
}

@misc{shen2021mred,
  title        = {{MReD}: A Meta-Review Dataset for Structure-Controllable Text Generation},
  author       = {Chenhui Shen and Liying Cheng and Ran Zhou and Lidong Bing and Yang You and Luo Si},
  year         = {2021},
  eprint       = {2110.07474},
  archivePrefix= {arXiv},
  primaryClass = {cs.CL},
  url          = {https://arxiv.org/abs/2110.07474}
}

@article{yuan2022asapreview,
  title       = {{Can We Automate Scientific Reviewing?}},
  author      = {Weizhe Yuan and Pengfei Liu and Graham Neubig},
  journal     = {Journal of Artificial Intelligence Research},
  volume      = {75},
  pages       = {171--212},
  year        = {2022},
  doi         = {10.1613/jair.1.12862},
  url         = {https://jair.org/index.php/jair/article/view/12862},
  note        = {Introduces the \textit{ASAP-Review} dataset.}
}

@inproceedings{wu2022prrca,
  title       = {Incorporating Peer Reviews and Rebuttal Counter-Arguments for Meta-Review Generation},
  author      = {Po-Cheng Wu and An-Zi Yen and Hen-Hsen Huang and Hsin-Hsi Chen},
  booktitle   = {Proceedings of the 31st ACM International Conference on Information \& Knowledge Management (CIKM 2022)},
  pages       = {2189--2198},
  year        = {2022},
  publisher   = {Association for Computing Machinery},
  doi         = {10.1145/3511808.3557360},
  url         = {https://doi.org/10.1145/3511808.3557360},
  note        = {Introduces the \textit{PRRCA} dataset}
}

@misc{zhang2022fairness,
  title        = {Investigating Fairness Disparities in Peer Review: A Language Model Enhanced Approach},
  author       = {Jiayao Zhang and Hongming Zhang and Zhun Deng and Dan Roth},
  year         = {2022},
  eprint       = {2211.06398},
  archivePrefix= {arXiv},
  primaryClass = {cs.CL},
  url          = {https://arxiv.org/abs/2211.06398},
  note         = {Introduces the \textit{ICLR-DB} dataset (10k submissions, 36k reviews, 68k responses from ICLR 2017–2022).}
}

@misc{darcy2023aries,
  title        = {{ARIES}: A Corpus of Scientific Paper Edits Made in Response to Peer Reviews},
  author       = {Mike D'Arcy and Alexis Ross and Erin Bransom and Bailey Kuehl and Jonathan Bragg and Tom Hope and Doug Downey},
  year         = {2023},
  eprint       = {2306.12587},
  archivePrefix= {arXiv},
  primaryClass = {cs.CL},
  url          = {https://arxiv.org/abs/2306.12587}
}

@misc{jin2024agentreview,
  title        = {{AgentReview}: Exploring Peer Review Dynamics with {LLM} Agents},
  author       = {Yiqiao Jin and Qinlin Zhao and Yiyang Wang and Hao Chen and Kaijie Zhu and Yijia Xiao and Jindong Wang},
  year         = {2024},
  eprint       = {2406.12708},
  archivePrefix= {arXiv},
  primaryClass = {cs.CL},
  url          = {https://arxiv.org/abs/2406.12708}
}

@misc{gao2024reviewer2,
  title        = {Reviewer2: Optimizing Review Generation Through Prompt Generation},
  author       = {Zhaolin Gao and Kianté Brantley and Thorsten Joachims},
  year         = {2024},
  eprint       = {2402.10886},
  archivePrefix= {arXiv},
  primaryClass = {cs.CL},
  url          = {https://arxiv.org/abs/2402.10886}
}

@inproceedings{zhou2024rrmcq,
  title       = {Is {LLM} a Reliable Reviewer? A Comprehensive Evaluation of {LLM} on Automatic Paper Reviewing Tasks},
  author      = {Ruiyang Zhou and Lu Chen and Kai Yu},
  booktitle   = {Proceedings of the 2024 Joint International Conference on Computational Linguistics, Language Resources and Evaluation (LREC-COLING 2024)},
  year        = {2024},
  url         = {https://aclanthology.org/2024.lrec-main.816},
  note        = {Introduces the \textit{RR-MCQ} dataset of 197 review–revision multiple-choice questions.}
}

@inproceedings{zhu2025deepreview,
    title = "{D}eep{R}eview: Improving {LLM}-based Paper Review with Human-like Deep Thinking Process",
    author = "Zhu, Minjun  and
      Weng, Yixuan  and
      Yang, Linyi  and
      Zhang, Yue",
    editor = "Che, Wanxiang  and
      Nabende, Joyce  and
      Shutova, Ekaterina  and
      Pilehvar, Mohammad Taher",
    booktitle = "Proceedings of the 63rd Annual Meeting of the Association for Computational Linguistics (Volume 1: Long Papers)",
    month = jul,
    year = "2025",
    address = "Vienna, Austria",
    publisher = "Association for Computational Linguistics",
    url = "https://aclanthology.org/2025.acl-long.1420/",
    doi = "10.18653/v1/2025.acl-long.1420",
    pages = "29330--29355",
    ISBN = "979-8-89176-251-0"
}

@misc{zhang2025re2,
  title        = {Re$^{2}$: A Consistency-Ensured Dataset for Full-Stage Peer Review and Multi-Turn Rebuttal Discussions},
  author       = {Daoze Zhang and Zhijian Bao and Sihang Du and Zhiyi Zhao and Kuangling Zhang and Dezheng Bao and Yang Yang},
  year         = {2025},
  eprint       = {2505.07920},
  archivePrefix= {arXiv},
  primaryClass = {cs.CL},
  url          = {https://arxiv.org/abs/2505.07920}
}

@misc{choudhary2022react,
  title        = {ReAct: A Review Comment Dataset for Actionability (and more)},
  author       = {Gautam Choudhary and Natwar Modani and Nitish Maurya},
  year         = {2022},
  eprint       = {2210.00443},
  archivePrefix= {arXiv},
  primaryClass = {cs.CL},
  url          = {https://arxiv.org/abs/2210.00443}
}

@misc{meng2024simposimplepreferenceoptimization,
      title={SimPO: Simple Preference Optimization with a Reference-Free Reward}, 
      author={Yu Meng and Mengzhou Xia and Danqi Chen},
      year={2024},
      eprint={2405.14734},
      archivePrefix={arXiv},
      primaryClass={cs.CL},
      url={https://arxiv.org/abs/2405.14734}, 
}

@misc{purkayastha2023exploringjiujitsuargumentationwriting,
      title={Exploring Jiu-Jitsu Argumentation for Writing Peer Review Rebuttals}, 
      author={Sukannya Purkayastha and Anne Lauscher and Iryna Gurevych},
      year={2023},
      eprint={2311.03998},
      archivePrefix={arXiv},
      primaryClass={cs.CL},
      url={https://arxiv.org/abs/2311.03998}, 
}

@inproceedings{gao-etal-2019-rebuttal,
    title = "Does My Rebuttal Matter? Insights from a Major {NLP} Conference",
    author = "Gao, Yang  and
      Eger, Steffen  and
      Kuznetsov, Ilia  and
      Gurevych, Iryna  and
      Miyao, Yusuke",
    editor = "Burstein, Jill  and
      Doran, Christy  and
      Solorio, Thamar",
    booktitle = "Proceedings of the 2019 Conference of the North {A}merican Chapter of the Association for Computational Linguistics: Human Language Technologies, Volume 1 (Long and Short Papers)",
    month = jun,
    year = "2019",
    address = "Minneapolis, Minnesota",
    publisher = "Association for Computational Linguistics",
    url = "https://aclanthology.org/N19-1129/",
    doi = "10.18653/v1/N19-1129",
    pages = "1274--1290"
}

@article{Huang_2023,
   title={What makes a successful rebuttal in computer science conferences?: A perspective on social interaction},
   volume={17},
   ISSN={1751-1577},
   url={http://dx.doi.org/10.1016/j.joi.2023.101427},
   DOI={10.1016/j.joi.2023.101427},
   number={3},
   journal={Journal of Informetrics},
   publisher={Elsevier BV},
   author={Huang, Junjie and Huang, Win-bin and Bu, Yi and Cao, Qi and Shen, Huawei and Cheng, Xueqi},
   year={2023},
   month=aug, pages={101427} }

@inproceedings{Bhatia,
author = {Bhatia, Chaitanya and Pradhan, Tribikram and Pal, Sukomal},
title = {MetaGen: An academic Meta-review Generation system},
year = {2020},
isbn = {9781450380164},
publisher = {Association for Computing Machinery},
address = {New York, NY, USA},
url = {https://doi.org/10.1145/3397271.3401190},
doi = {10.1145/3397271.3401190},
booktitle = {Proceedings of the 43rd International ACM SIGIR Conference on Research and Development in Information Retrieval},
pages = {1653–1656},
numpages = {4},
location = {Virtual Event, China},
series = {SIGIR '20}
}

@inproceedings{Kumar,
author = {Kumar, Asheesh and Ghosal, Tirthankar and Ekbal, Asif},
title = {A Deep Neural Architecture for Decision-Aware Meta-Review Generation},
year = {2024},
isbn = {9781665417709},
publisher = {IEEE Press},
url = {https://doi.org/10.1109/JCDL52503.2021.00064},
doi = {10.1109/JCDL52503.2021.00064},
booktitle = {Proceedings of the 2021 ACM/IEEE Joint Conference on Digital Libraries},
pages = {222–225},
numpages = {4},
location = {Virtual Event},
series = {JCDL '21}
}

@misc{zeng2024scientificopinionsummarizationpaper,
      title={Scientific Opinion Summarization: Paper Meta-review Generation Dataset, Methods, and Evaluation}, 
      author={Qi Zeng and Mankeerat Sidhu and Ansel Blume and Hou Pong Chan and Lu Wang and Heng Ji},
      year={2024},
      eprint={2305.14647},
      archivePrefix={arXiv},
      primaryClass={cs.CL},
      url={https://arxiv.org/abs/2305.14647}, 
}

@inproceedings{li-etal-2024-sentiment,
    title = "A Sentiment Consolidation Framework for Meta-Review Generation",
    author = "Li, Miao  and
      Lau, Jey Han  and
      Hovy, Eduard",
    editor = "Ku, Lun-Wei  and
      Martins, Andre  and
      Srikumar, Vivek",
    booktitle = "Proceedings of the 62nd Annual Meeting of the Association for Computational Linguistics (Volume 1: Long Papers)",
    month = aug,
    year = "2024",
    address = "Bangkok, Thailand",
    publisher = "Association for Computational Linguistics",
    url = "https://aclanthology.org/2024.acl-long.547/",
    doi = "10.18653/v1/2024.acl-long.547",
    pages = "10158--10177"
}

@misc{jiang2022arxiveditsunderstandinghumanrevision,
      title={arXivEdits: Understanding the Human Revision Process in Scientific Writing}, 
      author={Chao Jiang and Wei Xu and Samuel Stevens},
      year={2022},
      eprint={2210.15067},
      archivePrefix={arXiv},
      primaryClass={cs.CL},
      url={https://arxiv.org/abs/2210.15067}, 
}

@misc{jourdan2024casimircorpusscientificarticles,
      title={CASIMIR: A Corpus of Scientific Articles enhanced with Multiple Author-Integrated Revisions}, 
      author={Leane Jourdan and Florian Boudin and Nicolas Hernandez and Richard Dufour},
      year={2024},
      eprint={2403.00241},
      archivePrefix={arXiv},
      primaryClass={cs.CL},
      url={https://arxiv.org/abs/2403.00241}, 
}

@inproceedings{dycke-etal-2025-stricta,
    title = "{STRICTA}: Structured Reasoning in Critical Text Assessment for Peer Review and Beyond",
    author = "Dycke, Nils  and
      Ze{\v{c}}evi{\'c}, Matej  and
      Kuznetsov, Ilia  and
      Suess, Beatrix  and
      Kersting, Kristian  and
      Gurevych, Iryna",
    editor = "Che, Wanxiang  and
      Nabende, Joyce  and
      Shutova, Ekaterina  and
      Pilehvar, Mohammad Taher",
    booktitle = "Proceedings of the 63rd Annual Meeting of the Association for Computational Linguistics (Volume 1: Long Papers)",
    month = jul,
    year = "2025",
    address = "Vienna, Austria",
    publisher = "Association for Computational Linguistics",
    url = "https://aclanthology.org/2025.acl-long.1107/",
    doi = "10.18653/v1/2025.acl-long.1107",
    pages = "22687--22727",
    ISBN = "979-8-89176-251-0"
}

@misc{lin2024evaluatingenhancinglargelanguage,
      title={Evaluating and Enhancing Large Language Models for Novelty Assessment in Scholarly Publications}, 
      author={Ethan Lin and Zhiyuan Peng and Yi Fang},
      year={2024},
      eprint={2409.16605},
      archivePrefix={arXiv},
      primaryClass={cs.CL},
      url={https://arxiv.org/abs/2409.16605}, 
}

@misc{afzal2025notnovelenoughenriching,
      title={Beyond "Not Novel Enough": Enriching Scholarly Critique with LLM-Assisted Feedback}, 
      author={Osama Mohammed Afzal and Preslav Nakov and Tom Hope and Iryna Gurevych},
      year={2025},
      eprint={2508.10795},
      archivePrefix={arXiv},
      primaryClass={cs.CL},
      url={https://arxiv.org/abs/2508.10795}, 
}

@misc{rubaiat2025mappingevolutionresearchcontributions,
      title={Mapping the Evolution of Research Contributions using KnoVo}, 
      author={Sajratul Y. Rubaiat and Syed N. Sakib and Hasan M. Jamil},
      year={2025},
      eprint={2506.17508},
      archivePrefix={arXiv},
      primaryClass={cs.DL},
      url={https://arxiv.org/abs/2506.17508}, 
}

@article{wedel2024affiliation,
  author    = {von Wedel, D. and Schmitt, R. A. and Thiele, M. and Leuner, R. and Shay, D. and Redaelli, S. and Schaefer, M. S.},
  title     = {Affiliation Bias in Peer Review of Abstracts by a Large Language Model},
  journal   = {JAMA},
  year      = {2024},
  volume    = {331},
  number    = {3},
  pages     = {252--253},
  doi       = {10.1001/jama.2023.24641},
  pmid      = {38150261},
  pmcid     = {PMC10753437},
  month     = jan
}

@misc{liang2024monitoringaimodifiedcontentscale,
      title={Monitoring AI-Modified Content at Scale: A Case Study on the Impact of ChatGPT on AI Conference Peer Reviews}, 
      author={Weixin Liang and Zachary Izzo and Yaohui Zhang and Haley Lepp and Hancheng Cao and Xuandong Zhao and Lingjiao Chen and Haotian Ye and Sheng Liu and Zhi Huang and Daniel A. McFarland and James Y. Zou},
      year={2024},
      eprint={2403.07183},
      archivePrefix={arXiv},
      primaryClass={cs.CL},
      url={https://arxiv.org/abs/2403.07183}, 
}

@misc{thelwall2024evaluatingpredictivecapacitychatgpt,
      title={Evaluating the Predictive Capacity of ChatGPT for Academic Peer Review Outcomes Across Multiple Platforms}, 
      author={Mike Thelwall and Abdullah Yaghi},
      year={2024},
      eprint={2411.09763},
      archivePrefix={arXiv},
      primaryClass={cs.DL},
      url={https://arxiv.org/abs/2411.09763}, 
}

@article{Laureano2025Predicting,
	author = {Laureano, Mayte H and Calvo, Hiram and Alcantara, Tania and Garc\i{}a Vazquez, Omar and Cardoso Moreno, Marco A},
	journal = {Journal of Scientometric Research},
	doi = {10.5530/jscires.20251456},
	issn = {2321-6654},
	number = {1},
	year = {2025},
	month = {mar 27},
	pages = {331--341},
	publisher = {Manuscript Technomedia LLP},
	title = {Predicting {Reviewers}\textquoteright{} {Decisions} in {Scientific} {Submissions} through {Linguistic} {Analysis}},
	url = {http://dx.doi.org/10.5530/jscires.20251456},
	volume = {14},
}

@article{kitchenham2007guidelines,
  title={Guidelines for performing systematic literature reviews in software engineering technical report},
  author={Kitchenham, Barbara and Charters, Stuart},
  journal={Software Engineering Group, EBSE Technical Report, Keele University and Department of Computer Science University of Durham},
  volume={2},
  year={2007}
}

@article {Pagen71,
	author = {Page, Matthew J and McKenzie, Joanne E and Bossuyt, Patrick M and Boutron, Isabelle and Hoffmann, Tammy C and Mulrow, Cynthia D and Shamseer, Larissa and Tetzlaff, Jennifer M and Akl, Elie A and Brennan, Sue E and Chou, Roger and Glanville, Julie and Grimshaw, Jeremy M and Hr{\'o}bjartsson, Asbj{\o}rn and Lalu, Manoj M and Li, Tianjing and Loder, Elizabeth W and Mayo-Wilson, Evan and McDonald, Steve and McGuinness, Luke A and Stewart, Lesley A and Thomas, James and Tricco, Andrea C and Welch, Vivian A and Whiting, Penny and Moher, David},
	title = {The PRISMA 2020 statement: an updated guideline for reporting systematic reviews},
	volume = {372},
	elocation-id = {n71},
	year = {2021},
	doi = {10.1136/bmj.n71},
	publisher = {BMJ Publishing Group Ltd},
	URL = {https://www.bmj.com/content/372/bmj.n71},
	eprint = {https://www.bmj.com/content/372/bmj.n71.full.pdf},
	journal = {BMJ}
}

@misc{mineru,
      title={MinerU: An Open-Source Solution for Precise Document Content Extraction}, 
      author={Bin Wang and Chao Xu and Xiaomeng Zhao and Linke Ouyang and Fan Wu and Zhiyuan Zhao and Rui Xu and Kaiwen Liu and Yuan Qu and Fukai Shang and Bo Zhang and Liqun Wei and Zhihao Sui and Wei Li and Botian Shi and Yu Qiao and Dahua Lin and Conghui He},
      year={2024},
      eprint={2409.18839},
      archivePrefix={arXiv},
      primaryClass={cs.CV},
      url={https://arxiv.org/abs/2409.18839}, 
}

@misc{zou2026diagpaperdiagnosingvalidspecific,
      title={DIAGPaper: Diagnosing Valid and Specific Weaknesses in Scientific Papers via Multi-Agent Reasoning}, 
      author={Zhuoyang Zou and Abolfazl Ansari and Delvin Ce Zhang and Dongwon Lee and Wenpeng Yin},
      year={2026},
      eprint={2601.07611},
      archivePrefix={arXiv},
      primaryClass={cs.AI},
      url={https://arxiv.org/abs/2601.07611}, 
}

@misc{zeng2025reviewrlautomatedscientificreview,
      title={ReviewRL: Towards Automated Scientific Review with RL}, 
      author={Sihang Zeng and Kai Tian and Kaiyan Zhang and Yuru wang and Junqi Gao and Runze Liu and Sa Yang and Jingxuan Li and Xinwei Long and Jiaheng Ma and Biqing Qi and Bowen Zhou},
      year={2025},
      eprint={2508.10308},
      archivePrefix={arXiv},
      primaryClass={cs.CL},
      url={https://arxiv.org/abs/2508.10308}, 
}

@misc{han2026drpgdecomposeretrieveplan,
      title={DRPG (Decompose, Retrieve, Plan, Generate): An Agentic Framework for Academic Rebuttal}, 
      author={Peixuan Han and Yingjie Yu and Jingjun Xu and Jiaxuan You},
      year={2026},
      eprint={2601.18081},
      archivePrefix={arXiv},
      primaryClass={cs.LG},
      url={https://arxiv.org/abs/2601.18081}, 
}

@misc{ma2026paper2rebuttalmultiagentframeworktransparent,
      title={Paper2Rebuttal: A Multi-Agent Framework for Transparent Author Response Assistance}, 
      author={Qianli Ma and Chang Guo and Zhiheng Tian and Siyu Wang and Jipeng Xiao and Yuanhao Yue and Zhipeng Zhang},
      year={2026},
      eprint={2601.14171},
      archivePrefix={arXiv},
      primaryClass={cs.AI},
      url={https://arxiv.org/abs/2601.14171}, 
}

@misc{ruan2026authorintheloopresponsegenerationevaluation,
      title={Author-in-the-Loop Response Generation and Evaluation: Integrating Author Expertise and Intent in Responses to Peer Review}, 
      author={Qian Ruan and Iryna Gurevych},
      year={2026},
      eprint={2602.11173},
      archivePrefix={arXiv},
      primaryClass={cs.CL},
      url={https://arxiv.org/abs/2602.11173}, 
}

@misc{khatri2026defendautomatedrebuttalspeer,
      title={Defend: Automated Rebuttals for Peer Review with Minimal Author Guidance}, 
      author={Jyotsana Khatri and Manasi Patwardhan},
      year={2026},
      eprint={2603.27360},
      archivePrefix={arXiv},
      primaryClass={cs.AI},
      url={https://arxiv.org/abs/2603.27360}, 
}

@misc{purkayastha2026decisionmakingdeliberationmetareviewingdocumentgrounded,
      title={Decision-Making with Deliberation: Meta-reviewing as a Document-grounded Dialogue}, 
      author={Sukannya Purkayastha and Nils Dycke and Anne Lauscher and Iryna Gurevych},
      year={2026},
      eprint={2508.05283},
      archivePrefix={arXiv},
      primaryClass={cs.CL},
      url={https://arxiv.org/abs/2508.05283}, 
}

@misc{taechoyotin2026remctxautomatedpeerreview,
      title={REM-CTX: Automated Peer Review via Reinforcement Learning with Auxiliary Context}, 
      author={Pawin Taechoyotin and Daniel E. Acuna},
      year={2026},
      eprint={2604.00248},
      archivePrefix={arXiv},
      primaryClass={cs.CL},
      url={https://arxiv.org/abs/2604.00248}, 
}

@misc{wu2026novbenchevaluatinglargelanguage,
      title={NovBench: Evaluating Large Language Models on Academic Paper Novelty Assessment}, 
      author={Wenqing Wu and Yi Zhao and Yuzhuo Wang and Siyou Li and Juexi Shao and Yunfei Long and Chengzhi Zhang},
      year={2026},
      eprint={2604.11543},
      archivePrefix={arXiv},
      primaryClass={cs.CL},
      url={https://arxiv.org/abs/2604.11543}, 
}

@misc{sadallah2025goodbadconstructiveautomatically,
      title={The Good, the Bad and the Constructive: Automatically Measuring Peer Review's Utility for Authors}, 
      author={Abdelrahman Sadallah and Tim Baumgärtner and Iryna Gurevych and Ted Briscoe},
      year={2025},
      eprint={2509.04484},
      archivePrefix={arXiv},
      primaryClass={cs.CL},
      url={https://arxiv.org/abs/2509.04484}, 
}

@misc{wu2026rbtactrebuttalsupervisionactionable,
      title={RbtAct: Rebuttal as Supervision for Actionable Review Feedback Generation}, 
      author={Sihong Wu and Yiling Ma and Yilun Zhao and Tiansheng Hu and Owen Jiang and Manasi Patwardhan and Arman Cohan},
      year={2026},
      eprint={2603.09723},
      archivePrefix={arXiv},
      primaryClass={cs.CL},
      url={https://arxiv.org/abs/2603.09723}, 
}

@misc{mun2026goodpointlearningconstructivescientific,
      title={GoodPoint: Learning Constructive Scientific Paper Feedback from Author Responses}, 
      author={Jimin Mun and Chani Jung and Xuhui Zhou and Hyunwoo Kim and Maarten Sap},
      year={2026},
      eprint={2604.11924},
      archivePrefix={arXiv},
      primaryClass={cs.AI},
      url={https://arxiv.org/abs/2604.11924}, 
}

@misc{actreview,
  title = {ActReview: A Review-to-Revision Framework for Actionable Peer Review Generation},
  author = {Yiling Ma and Yilun Zhao and Sihong Wu and Ziyu Chen and Manasi Patwardhan and Arman Cohan},
  year = {2026},
}

@inproceedings{shen-etal-2022-mred,
    title = "{MR}e{D}: A Meta-Review Dataset for Structure-Controllable Text Generation",
    author = "Shen, Chenhui  and
      Cheng, Liying  and
      Zhou, Ran  and
      Bing, Lidong  and
      You, Yang  and
      Si, Luo",
    editor = "Muresan, Smaranda  and
      Nakov, Preslav  and
      Villavicencio, Aline",
    booktitle = "Findings of the Association for Computational Linguistics: ACL 2022",
    month = may,
    year = "2022",
    address = "Dublin, Ireland",
    publisher = "Association for Computational Linguistics",
    url = "https://aclanthology.org/2022.findings-acl.198/",
    doi = "10.18653/v1/2022.findings-acl.198",
    pages = "2521--2535"
}

@misc{garg2024revieweval,
      title={ReviewEval: An Evaluation Framework for AI-Generated Reviews}, 
      author={Madhav Krishan Garg and Tejash Prasad and Tanmay Singhal and Chhavi Kirtani and Murari Mandal and Dhruv Kumar},
      year={2025},
      eprint={2502.11736},
      archivePrefix={arXiv},
      primaryClass={cs.CL},
      url={https://arxiv.org/abs/2502.11736}, 
}
